\documentclass[sigconf,nonacm]{acmart}

\AtBeginDocument{%
  }

\usepackage{booktabs}
\usepackage{multirow}
\usepackage{graphicx}
\usepackage{amsmath}

\usepackage{array}

\usepackage{listings}

\usepackage{pifont}
\usepackage{xcolor}

\usepackage{hyperref}
\usepackage{url}
\usepackage{booktabs}
\usepackage{multirow}
\usepackage{graphicx}
\usepackage{tabularx,amsmath}

\usepackage{array}

\usepackage{listings}
\usepackage{xcolor}

\usepackage[table]{xcolor}

\definecolor{ablationgain}{HTML}{2166AC}
\definecolor{ablationloss}{HTML}{B2182B}

\usepackage[table]{xcolor}
\usepackage{pifont}

\usepackage{booktabs}
\usepackage{longtable}
\usepackage{array}

\usepackage{multirow}
\usepackage{graphicx}
\usepackage{amsmath}

\definecolor{coverageyes}{HTML}{4F8A61}
\definecolor{coverageno}{HTML}{C96B6B}

\newcommand{\yesmark}{\textcolor{coverageyes}{\ding{51}}}
\newcommand{\nomark}{\textcolor{coverageno}{\ding{55}}}

\begin{document}


\title{Spec2Game: Can LLMs Generate Complete Playable Games from Detailed Specifications?}


\author{Yixue Cai}
\affiliation{%
  \institution{The Chinese University of Hong Kong}
  \city{Hong Kong}
  \country{China}
}
\email{1155267536@link.cuhk.edu.hk}

\author{Yuzhe Zhao}
\affiliation{%
  \institution{Nankai University}
  \city{Tianjin}
  \country{China}
}
\email{zhaoyuzhe@mail.nankai.edu.cn}

\author{Hanxiang Chao}
\affiliation{%
  \institution{Wuhan University}
  \city{Wuhan}
  \country{China}
}
\email{chx\_whu@whu.edu.cn}

\author{Qingsen Ma}
\affiliation{%
  \institution{The Chinese University of Hong Kong}
  \city{Hong Kong}
  \country{China}
}
\email{1155281709@link.cuhk.edu.hk}

\author{Ziheng Xiong}
\affiliation{%
  \institution{The Chinese University of Hong Kong}
  \city{Hong Kong}
  \country{China}
}
\email{1155186155@link.cuhk.edu.hk}

\author{Jinhu Qi}
\affiliation{%
  \institution{The Chinese University of Hong Kong}
  \city{Hong Kong}
  \country{China}
}
\email{1155251284@link.cuhk.edu.hk}

\author{Irwin King}
\affiliation{%
  \institution{The Chinese University of Hong Kong}
  \city{Hong Kong}
  \country{China}
}
\email{king@cse.cuhk.edu.hk}

\renewcommand{\shortauthors}{Cai et al.}


\begin{abstract}

Generating an executable program does not necessarily mean that it correctly implements the behavioral requirements specified in natural language. To evaluate large language models' ability to realize detailed specifications as complete interactive programs, we introduce \textbf{Spec2Game}, a benchmark that requires models to generate complete Pygame projects from detailed natural-language game specifications. Spec2Game comprises 15 game families and 150 task instances, with one canonical task and nine controlled rule variants per family, spanning three levels of implementation complexity. Using source-code, runtime, and visual evidence, we evaluate generated projects along four dimensions---\textbf{Executability}, \textbf{Specification Realization}, \textbf{Code Quality}, and \textbf{User-Facing Quality}. Across 14 LLMs and 3,330 generated projects, we find that high executability does not imply faithful specification realization. Component-level analysis further shows that models perform substantially better on Game Element Modeling than on Rule and Mechanism Modeling or Goal and Termination Modeling, indicating that faithfully implementing game rules and termination logic remains a major challenge.

\end{abstract}





\maketitle

\section{Introduction}
\label{sec:introduction}

Generating complete software from natural-language specifications requires more than producing locally correct code. A model must coordinate multiple components, maintain consistent program state, and implement requirements that interact over time. For interactive software, successful execution does not necessarily mean that the program implements the specified behaviors or provides appropriate feedback. Evaluating this capability therefore requires examining how faithfully a program realizes specific requirements.

Games provide a suitable setting for studying this problem by combining objects, user inputs, rules, state transitions, feedback, and termination conditions within a single interactive system. These elements allow concrete functional requirements to be specified and their implementation to be examined through controlled execution. Functional behavior and visual presentation are also distinct: a game may appear correct while violating its specified rules, or implement the rules correctly while failing to communicate its current state clearly.

\begin{figure*}[htbp]
    \centering
    \includegraphics[width=\textwidth]{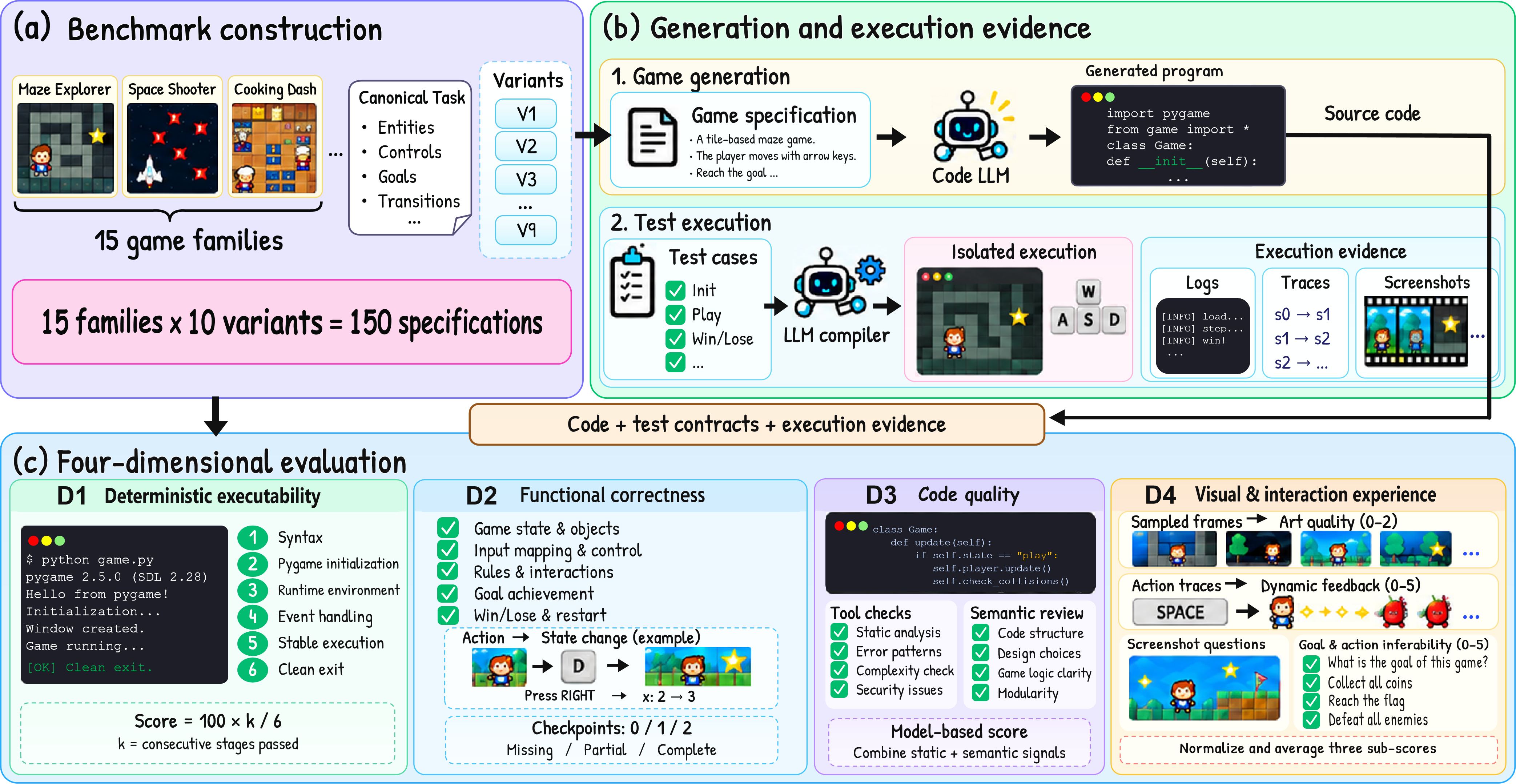}
    \caption{Overview of Spec2Game.
    (a) The benchmark comprises 15 game families, each with one canonical task and nine controlled variants, forming Core-15 (15 canonical tasks) and Full-150 (150 task instances).
    (b) A fixed staged generation protocol produces complete Pygame projects, followed by isolated execution and evaluation evidence collection.
    (c) Generated projects are evaluated along four complementary dimensions (D1--D4).
    Code snippets, traces, and screenshots are schematic.}
    \label{fig:spec2game-overview}
\end{figure*}

Existing research has extended code-generation evaluation to interactive applications and complete game generation~\citep{zhang2026webgamebench,luo2026gamecraftbench,zhang2026miniappbench}. Building on this work, we introduce \textbf{Spec2Game} (Figure~\ref{fig:spec2game-overview}), a benchmark for evaluating LLMs' ability to generate complete interactive Pygame projects from detailed natural-language specifications, with an emphasis on the realization of specific requirements. The benchmark contains 15 game families and 150 task instances. Canonical tasks and their controlled rule variants provide different rule configurations, while a mechanism-grounded complexity scoring framework organizes the tasks to support evaluation of specification realization across implementation-complexity levels.

Spec2Game evaluates generated projects along four complementary dimensions: \textbf{Executability}, \textbf{Specification Realization}, \textbf{Code Quality}, and \textbf{User-Facing Quality}. We introduce checkpoints to assess the realization of specific requirements and user-visible dynamic feedback. The framework supports comparisons across models, tasks, and complexity levels, while checkpoint-level results identify specific implementation failures. We additionally conduct two ablation studies and a human assessment.

Our contributions are threefold:
\begin{itemize}
    \item We construct a benchmark of complete game-generation tasks driven by detailed natural-language specifications, comprising canonical tasks and controlled rule variants. We also establish a mechanism-grounded complexity scoring framework that provides a consistent basis for task organization and future extension.

    \item We establish a standardized benchmark pipeline comprising a unified staged generation protocol and a four-dimensional evaluation framework. We further develop checkpoint-level evaluation methods to assess the realization of specific requirements and user-visible dynamic feedback.

    \item We systematically evaluate 14 LLMs on 3,330 generated projects and identify key capability limitations in complete game generation: even when executability is nearly saturated, specification realization remains substantially weaker. Component-level analysis further shows that models perform well on Game Element Modeling but substantially worse on Rule and Mechanism Modeling and Goal and Termination Modeling.
\end{itemize}

\section{Related Work}
\label{sec:related_work}

\paragraph{Code generation and development.}
HumanEval, MBPP, EvalPlus, and LiveCodeBench evaluate self-contained programming problems~\citep{chen2021evaluatinglargelanguagemodels,austin2021programsynthesislargelanguage,liu2023codegeneratedchatgpt,jain2024livecodebench}, while SWE-bench addresses repository-level issues~\citep{jimenez2024swebench}. ChatDev and MetaGPT explore structured multi-agent development~\citep{qian2024chatdev,hong2024metagpt}. Spec2Game instead evaluates generated games under a fixed four-stage protocol without execution-guided semantic repair.

\paragraph{Interactive artifacts and game testbeds.}
WebGen-Bench, WebCoderBench, and MiniAppBench study interactive application generation~\citep{lu2025webgenbench,liu2026webcoderbench,zhang2026miniappbench}; PlayEval examines GUI behavior~\citep{peng2026playcoder}; and ArtifactsBench incorporates code, rendered outputs, and temporal visual evidence~\citep{zhang2025artifactsbench}. Other game testbeds assess game-playing agents or procedural content generation~\citep{Bellemare_2013,MarioAI,sudhakaran2023mariogpt,khalifa2025pcg}.

\paragraph{Game generation and verification.}
V-GameGym evaluates repository-derived Pygame tasks using code, image, and video evidence~\citep{zhang2026vgamegym}. WebGameBench and OpenGame-Bench assess browser-game generation~\citep{zhang2026webgamebench,jiang2026opengame}, while GameCraft-Bench evaluates complete Godot games and GameDevBench targets existing game projects~\citep{luo2026gamecraftbench,chi2026gamedevbench}. GameGen-Verifier tests specification keypoints through runtime state injection and bounded interactions~\citep{jia2026gamegenverifier}. Spec2Game complements these settings with curated Pygame specifications, controlled rule variants within each game family, and separate evaluation of executability, specification realization, code quality, and user-facing quality.

\section{Spec2Game}
\label{sec:spec2game}

Spec2Game measures specification realization: how faithfully
generated projects implement specified states, interactions, rules,
feedback, and termination conditions.
We use Pygame to reduce dependence on engine-specific workflows
while retaining stateful logic, interaction, and visual feedback.

\subsection{Game Family Selection}
\label{sec:game_family_selection}

We selected 30 candidate games, primarily classic small-scale games,
based on implementation feasibility in Pygame, evaluation feasibility, and diversity of gameplay mechanics.
Using the framework in Section~\ref{sec:implementation_complexity},
we classified them as Easy, Medium, or Hard.
Two games per level formed a six-task pilot, with one project
generated per task by a single model.
The pilot assessed end-to-end pipeline feasibility, not model
performance or judge validity; its outputs were not reused in the
formal experiments
(Appendix~\ref{app:pilot}).

We retained one pilot game per complexity level and selected
12 additional families from the candidate pool to cover diverse
gameplay mechanics.
The final benchmark contains 15 game families, with five per
complexity level and three per game type
(Table~\ref{tab:15_game_families}).
Game type and implementation complexity serve as complementary
coverage criteria rather than a fully crossed design.
Each family contains one canonical task and nine controlled
variants.
The 15 canonical tasks constitute Core-15, while all 150 task
instances form Full-150.

\begin{table*}[htbp]
    \centering
    \small
    \setlength{\tabcolsep}{2pt}
    \renewcommand{\arraystretch}{1.20}

    \begin{tabular}{
        >{\centering\arraybackslash}m{0.27\textwidth}
        >{\centering\arraybackslash}m{0.13\textwidth}
        >{\centering\arraybackslash}m{0.25\textwidth}
        >{\centering\arraybackslash}m{0.32\textwidth}
    }
        \toprule
        \textbf{Game Type}
        & \textbf{Easy}
        & \textbf{Medium}
        & \textbf{Hard} \\
        \midrule

        Real-Time Reaction
        & \shortstack{Snake \\ Dodge Blocks \\ Whack-a-Mole}
        & --
        & -- \\[3pt]

        Shooting and Combat Action
        & Asteroids
        & Space Invaders
        & Top-down Arena Shooter \\[3pt]

        Spatial Exploration and Platform Adventure
        & Maze Runner
        & Simple Platformer
        & Mini Zelda-like Room Adventure \\[3pt]

        Logic and Puzzle
        & --
        & \shortstack{Tetris \\ Match-3}
        & Pipe Connection Puzzle \\[3pt]

        Strategy and Resource Management
        & --
        & Simple Shop Management
        & \shortstack{Carrot Defense-like \\ Farming-lite} \\

        \bottomrule
    \end{tabular}

    \caption{The 15 game families by type and implementation complexity.}
    \label{tab:15_game_families}
\end{table*}

\subsection{Implementation Complexity}
\label{sec:implementation_complexity}

Implementation complexity characterizes the demands of translating a game specification into a complete program, rather than the challenge faced by a player. Since code length alone cannot capture these demands, we draw on game-design research~\citep{jarvinen2008games,schell2008art,sicart2008defining} to define five dimensions: Operation Complexity, Object and Rule Complexity, Spatial and Scene Complexity, Adversarial Behavior Complexity, and Interface and Information Complexity.

The framework comprises 13 sub-dimensions, each scored from 1 to 3 according to predefined criteria. The overall complexity score is
\[
S=\sum_{i=1}^{13}s_i,\qquad s_i\in\{1,2,3\}.
\]
Based on the score distribution of the 30 candidate games, we define three implementation-complexity levels: Easy ($S\leq18$), Medium ($19\leq S\leq27$), and Hard ($S\geq28$). Each game family's nine variants inherit the complexity label of its canonical task. The full scoring rubric and game-level scores are provided in Appendix~\ref{app:complexity_rubric}.

\subsection{Standardized Game Specifications}
\label{sec:game_specifications}

We describe each game using a standardized seven-part natural-language specification: game overview, scene and interface, objects and states, player interaction, core rules and mechanics, game flow and progression, and goal, feedback, and termination.

For classic small-scale games, we introduce task-specific rule variations while preserving core gameplay to discourage models from simply reproducing conventional implementations. Each game family includes nine controlled variants derived from its canonical task, each introducing one primary change to a rule or design factor, with affected specification content updated accordingly. All variants are provided as standalone tasks. We extract checkable functional checkpoints from the specifications and provide the generation model with checkpoint-derived public behavioral requirements, while withholding evaluator-only metadata, detection methods, and scoring information. The complete template, variant-construction details, and examples are provided in Appendix~\ref{app:game_specifications}.

\subsection{Generation Protocol}
\label{sec:generation_protocol}

Spec2Game uses a fixed staged generation protocol to translate each task specification into a complete Pygame project. The benchmark automatically creates \texttt{\_\_init\_\_.py}, while the model generates the remaining four modules in a fixed order:
\[
\texttt{state.py} \rightarrow \texttt{game.py} \rightarrow \texttt{rendering.py} \rightarrow \texttt{main.py}.
\]
Each stage uses a shared generation contract and stage-specific instructions, and is issued as a separate request without shared chat history. Once generated, each module is locked and can only be used as fixed context by subsequent stages.

The four stages respectively handle state modeling, interaction and rule implementation, state visualization, and program integration, receiving the relevant specification content. The \texttt{rendering.py} stage also follows fixed task-independent visual-quality requirements. The \texttt{main.py} stage receives only public API summaries of the preceding modules rather than their full source code. Checkpoint-derived public behavioral requirements are assigned only to the responsible stages: D2.1 requirements to \texttt{state.py} and D2.2--D2.5 requirements to \texttt{game.py}. Evaluator-only checkpoint information and scoring rules are not exposed to the generation model. Detailed stage-specific inputs and prior context allocation are provided in Appendix~\ref{app:generation_protocol}.

Candidate programs are not executed during generation, and no runtime or evaluator feedback is returned to the model. The protocol does not perform semantic repair. A single technical retry using the same prompt and no additional error feedback is permitted only for request failures or outputs deemed incomplete.

\section{Evaluation Framework}
\label{sec:evaluation}

\subsection{Evaluation Dimensions}
\label{sec:evaluation_dimensions}

Evaluating complete interactive games requires considering not only execution and functional realization, but also engineering quality and user-facing presentation. Execution success or functional correctness alone therefore cannot capture the different quality aspects of generated projects. Existing benchmarks for interactive applications and game generation draw on evidence from static code, runtime behavior, and user-visible outputs~\citep{peng2026playcoder,zhang2026webgamebench,luo2026gamecraftbench,zhang2026miniappbench}. Drawing on these evaluation perspectives and the distinctions among software product quality attributes in ISO/IEC 25010:2023~\citep{iso25010}, Spec2Game evaluates generated game projects along four complementary dimensions: \textbf{D1 Executability}, \textbf{D2 Specification Realization}, \textbf{D3 Code Quality}, and \textbf{D4 User-Facing Quality}.

D1 assesses basic program executability; D2 examines the realization of specified objects, interactions, rules, dynamics, and goals; D3 evaluates the engineering quality of the source code; and D4 assesses visual presentation, dynamic feedback fidelity, and whether visible game states provide sufficient information for players to select appropriate actions. D4 does not assess gameplay fun, novelty, or difficulty. Table~\ref{tab:evaluation_dimensions} summarizes the four dimensions and their components.

\begin{table*}[htbp]
\centering
\caption{Evaluation dimensions and components in Spec2Game.}
\label{tab:evaluation_dimensions}
\small
\setlength{\tabcolsep}{5pt}
\renewcommand{\arraystretch}{1.08}
\begin{tabular}{p{0.27\linewidth} p{0.68\linewidth}}
\toprule
\textbf{Dimension} & \textbf{Components} \\
\midrule

\textbf{D1 Executability}
&
Syntax Validity; Dependency Readiness; Window Creation; Event Handling; Short-Horizon Runtime Stability; Controlled Termination
\\
\addlinespace[3pt]

\textbf{D2 Specification Realization}
&
Game Element Modeling; Interaction Modeling; Rule and Mechanism Modeling; Dynamic Evolution; Goal and Termination Modeling
\\
\addlinespace[3pt]

\textbf{D3 Code Quality}
&
Maintainability; Reliability; Security; Efficiency; Conformance
\\
\addlinespace[3pt]

\textbf{D4 User-Facing Quality}
&
Visual Presentation; Dynamic Feedback Fidelity; Visual Actionability
\\

\bottomrule
\end{tabular}
\end{table*}

\subsection{Evaluation Procedure}
\label{sec:evaluation_procedure}

Spec2Game uses dimension-specific evidence and evaluation procedures. 

\paragraph{D1: Executability.}
D1 uses a deterministic local evaluator without an LLM judge. Six component checks follow a fixed order: static checks, an approximately five-second runtime stability probe in a separate subprocess, and a \texttt{QUIT} event to verify normal termination. Scoring is sequentially gated: the first \(k\) consecutive passes yield \(100k/6\), and later passes cannot compensate for earlier failures.

\paragraph{D2: Specification Realization.}
D2 uses an evaluator-side LLM judge to assess all task-specific checkpoints across the five components, based on checkpoint-specific functional evidence obtained from controlled execution and, when needed, the generated source code. Checkpoints with explicit ordinal rubrics are scored as 0, 1, or 2, while condition-based checkpoints aggregate binary checks of their constituent conditions into a checkpoint score. Checkpoint scores are normalized within their respective components. A separate specification-deviation audit identifies substantive extra or incompatible mechanisms without penalizing harmless implementation differences, and assigns each component a deviation level \(p_j\in\{0,1,2,3\}\). Let \(r_j\in[0,1]\) denote the normalized requirement score for component \(j\):
\[
C_j=100\max\left(0,r_j-\frac{p_j}{3}\right),
\qquad
D2=\frac{1}{5}\sum_{j=1}^{5}C_j.
\]

\paragraph{D3: Code Quality.}
D3 combines deterministic code analysis with semantic source-code judging. The former uses static-analysis tools, custom AST checks, and an import probe, while the latter uses an LLM judge to assess source-code quality attributes not fully captured by deterministic rules. The two methods contribute 70 and 30 points, respectively, to the final D3 score, following predefined allocations across the five components. Severe security violations cap the final D3 score at 50.

\paragraph{D4: User-Facing Quality.}
D4 uses runtime visual evidence and a multimodal judge to evaluate three components. \textbf{Visual Presentation} is scored using sampled execution screenshots and five fixed visual criteria. \textbf{Dynamic Feedback Fidelity} evaluates the faithfulness, completeness, and timeliness of visual feedback across all dynamic checkpoints. \textbf{Visual Actionability} is scored through two binary next-action selection questions based on execution screenshots, given the task objective and the set of legal actions. Let \(d_1\), \(d_2\), and \(d_3\) denote the three component scores, with ranges 0--2, 0--5, and 0--5, respectively:
\[
D4=
\frac{100}{3}
\left(
\frac{d_1}{2}
+
\frac{d_2}{5}
+
\frac{d_3}{5}
\right).
\]

\paragraph{Overall score aggregation.}
The four dimensions are combined into the Overall score as
\[
g(D1)=\operatorname{clip}\left(\frac{D1-40}{20},0,1\right),
\]
\[
\mathrm{Overall}
=
0.20D1
+
g(D1)\left(
0.50D2
+
0.15D3
+
0.15D4
\right).
\]
D2 receives the largest weight to emphasize Specification Realization, the primary capability evaluated by Spec2Game, while D1, D3, and D4 provide complementary signals. The executability gate reduces the contribution of D2--D4 when basic execution is insufficient.

Detailed scoring rubrics, evidence construction procedures, and evaluator implementations for D1--D4 are provided in Appendix~\ref{app:evaluation}. D2 and D4 evaluation results are additionally compared with human judgments on a sampled subset of outputs, with the human-assessment protocol described in Section~\ref{sec:human_assessment}.

\section{Experiments}
\label{sec:experiments}

\subsection{Experimental Setup}
\label{sec:experimental_setup}

We evaluate 14 candidate models spanning multiple providers and model families (Table~\ref{tab:candidate_models}). All 14 models are evaluated on Core-15, with 6 models additionally evaluated on Full-150, which includes rule variants. For each model--track combination, we independently generate three candidates per task. Even when a model participates in both tracks, its Core-15 candidates are generated independently rather than reused from the corresponding canonical tasks in Full-150. This design comprises 2,700 Full-150 projects and 630 Core-15 projects, totaling 3,330 formal candidate projects.

\begin{table}[t]
    \centering
    \caption{Candidate models, access routes, and evaluation tracks.}
    \label{tab:candidate_models}
    \small
    \setlength{\tabcolsep}{3pt}
    \renewcommand{\arraystretch}{1.05}

    \resizebox{\columnwidth}{!}{%
    \begin{tabular}{llcc}
        \toprule
        \textbf{Candidate Model}
        & \textbf{Access Route}
        & \textbf{Full-150}
        & \textbf{Core-15} \\
        \midrule

        GPT-5.6 Sol
        & OpenRouter
        & \yesmark
        & \yesmark \\

        Grok 4.3
        & OpenRouter
        & \yesmark
        & \yesmark \\

        GLM-5.2
        & Zhipu official
        & \yesmark
        & \yesmark \\

        Kimi K2.6
        & Moonshot official
        & \yesmark
        & \yesmark \\

        Qwen3.8 Max-0902
        & Alibaba official
        & \yesmark
        & \yesmark \\

        DeepSeek V4 Pro-0813
        & DeepSeek official
        & \yesmark
        & \yesmark \\

        \midrule

        GPT-5.6 Terra
        & OpenRouter
        & \nomark
        & \yesmark \\

        GPT-5.6 Luna
        & OpenRouter
        & \nomark
        & \yesmark \\

        Claude Sonnet 5
        & OpenRouter
        & \nomark
        & \yesmark \\

        Claude Haiku 4.5
        & OpenRouter
        & \nomark
        & \yesmark \\

        Mistral Small 2603
        & OpenRouter
        & \nomark
        & \yesmark \\

        Qwen3.7 Plus
        & Alibaba official
        & \nomark
        & \yesmark \\

        Qwen3.8 Flash
        & Alibaba official
        & \nomark
        & \yesmark \\

        DeepSeek V4.1 Flash
        & DeepSeek official
        & \nomark
        & \yesmark \\

        \bottomrule
    \end{tabular}%
    }
\end{table}

\paragraph{Candidate generation.}
All formal candidates follow the staged generation protocol described in Section~\ref{sec:generation_protocol}. Detailed generation settings are provided in Appendix~\ref{app:main_benchmark_config}.

\paragraph{Evaluation models.}

We select text and vision judges based on their suitability for the evaluation tasks and the computational demands of large-scale evaluation, and access both through AWS. The text judge, Qwen3-235B-A22B-2507, performs action-program compilation, D2 scoring, D3 semantic source-code assessment, and evaluation-contract generation; the vision judge, Qwen3-VL-235B-A22B-Instruct, performs D4 scoring and generates and answers the Visual Actionability questions. Both judges use a temperature of 0 with reasoning disabled. The Qwen-family evaluators share a model family with three candidate models. Ablation II varies the text-scoring judge while keeping action-program compilation, evaluation-contract generation, evidence collection, and the vision judge fixed.

\subsection{Benchmark Results and Analysis}

\paragraph{Overall performance.}

\begin{figure*}[htbp]
    \centering
    \includegraphics[width=0.9\textwidth]{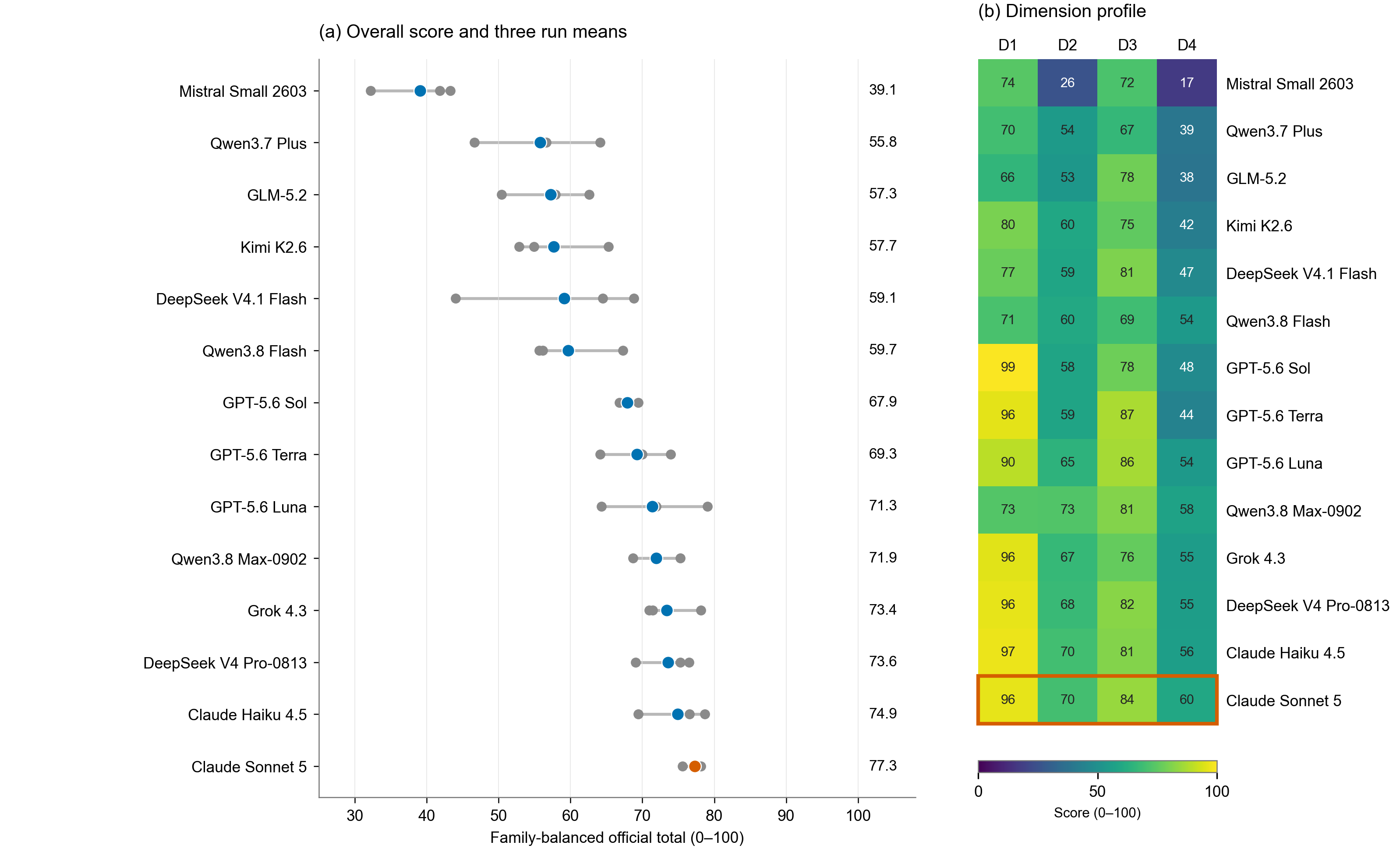}
    \caption{Core-15 comparison of Overall performance and dimension-wise scores. Overall scores in panel (a) are computed first and then aggregated; dimension scores in panel (b) are aggregated separately.}
    \label{fig:core15_main_comparison}
\end{figure*}

Core-15 provides a common basis for cross-model comparison. As shown in Fig.~\ref{fig:core15_main_comparison}, the family-balanced Overall score ranges from 39.11 to 77.27, with Claude Sonnet 5 achieving the highest score and Mistral Small 2603 the lowest. Results are first averaged across three runs within each game family and then averaged equally across the 15 families. Detailed per-model results are reported in Appendix~\ref{app:core15_detailed}.

\paragraph{Executability and specification realization.}



Overall scores can obscure differences across evaluation dimensions. GPT-5.6 Sol scores 99.26, 58.43, 77.81, and 47.87 on D1--D4, respectively, with an Overall score of 67.92. In comparison, Qwen3.8 Max-0902 scores 73.33, 72.85, 81.30, and 57.73 on D1--D4, respectively, with an Overall score of 71.95. Although their Overall scores are close, they differ substantially on D1 and D2, showing that Overall score alone cannot fully characterize model performance across the four evaluation dimensions.

Passing all executability checks does not guarantee a high D2 score. Among 630 candidate projects, 387 pass all six D1 checks, yet 87 of these still obtain a D2 score below 60. Additional D1 diagnostics are provided in Appendix~\ref{app:d1_diagnostics}. Component-level results further reveal substantial variation across capability components. Within D2, the average score for Game Element Modeling is 80.83, compared with 45.65 for Rule and Mechanism Modeling and 44.05 for Goal and Termination Modeling. Within D4, Visual Presentation has an average score of 70.38, while Dynamic Feedback Fidelity and Visual Actionability score 32.10 and 40.71, respectively. The D2 results indicate that models are substantially stronger at modeling game elements than at realizing rules, mechanisms, and termination conditions. Detailed component-level results for Core-15 and Full-150 are reported in Appendix~\ref{app:component_results}.

\paragraph{Performance across implementation complexity.}

Model performance decreases overall as the assigned implementation complexity increases. On Core-15, the family-balanced Overall scores for Easy, Medium, and Hard are 71.19, 65.00, and 58.42, respectively; on Full-150, the corresponding averages are 68.24, 59.61, and 51.23. Both independent evaluation tracks exhibit the same downward trend, indicating that the implementation complexity levels defined in Spec2Game distinguish model performance across tasks with different implementation demands. Family-level results and within-level variation are provided in Appendix~\ref{app:complexity_results}.

\paragraph{Sensitivity to controlled rule variants.}

For each model-family pair, we compare the canonical task with the nine controlled rule variants from the same family. Among the 90 pairs, 47 obtain lower scores on the variants and 43 obtain higher scores; with the difference defined as variant mean minus canonical score, the median difference is $-2.27$. These results show that controlled rule variants do not produce a consistent overall increase or decrease in model performance. Full-150 extends the rule space of Core-15, thereby broadening the rule coverage for comparing model performance under different rule configurations. Model- and family-level canonical--variant comparisons are provided in Appendix~\ref{app:variant_results}.

\subsection{Ablation Study I: Specification-Input Conditions}
\label{sec:ablations}

\paragraph{Design.}
We examine how specification detail and additional behavioral guidance
affect generated game quality.
We compare three input conditions on Core-15:
\textbf{M} provides a minimal game description;
\textbf{D} provides detailed specifications allocated to the relevant
generation stages; and
\textbf{B} augments D with a public behavioral checklist derived from
checkpoints.
The $D-M$ comparison evaluates the benefit of supplying detailed target
information, while $B-D$ evaluates the additional benefit of the checklist
once detailed specifications are available.
All conditions are evaluated against the same full target specification.
We evaluate three models across 15 games, retaining one candidate per
model--game--condition cell, for 135 candidates in total.
Generation follows the four-stage file order.
Detailed experimental settings are provided in
Appendix~\ref{app:ablation_1_configuration}.

\begin{table}[htbp]
    \centering
    \caption{Mean Core-15 overall scores under three specification-input
    conditions.}
    \label{tab:ablation_1_summary}
    \small
    \setlength{\tabcolsep}{3pt}
    \renewcommand{\arraystretch}{1.05}
    \begin{tabular}{@{}lrrrrr@{}}
        \toprule
        Model & M & D & B & $D-M$ & $B-D$ \\
        \midrule
        GPT-5.6 Terra
        & 55.18 & 74.28 & \textbf{75.40}
        & $+19.10$ & $+1.11$ \\
        Qwen3.7 Plus
        & 45.79 & \textbf{64.22} & 59.84
        & $+18.43$ & $-4.38$ \\
        Grok 4.3
        & 50.79 & 67.35 & \textbf{70.02}
        & $+16.56$ & $+2.66$ \\
        \midrule
        Overall
        & 50.59 & \textbf{68.62} & 68.42
        & $+18.03$ & $-0.20$ \\
        \bottomrule
    \end{tabular}
\end{table}

\paragraph{Results.}
Detailed specifications raise the mean Overall score from 50.59 to
68.62, with gains of 16.56--19.10 points across all three models
(Table~\ref{tab:ablation_1_summary}).
Adding the behavioral checklist yields smaller, model-dependent
changes, leaving the mean nearly unchanged at 68.42.
Thus, detailed target information provides the main average gain in
this setting, while checklist guidance offers no consistent
additional benefit.
Per-game results and additional analyses are provided in
Appendix~\ref{app:ablation_1_details}.

\subsection{Ablation Study II: Sensitivity to the Text-Scoring judge}
\label{sec:ablation_judge_rank}

\paragraph{Design.}
We compare Qwen3-235B-A22B-2507 and GPT-5.6 Sol as alternative text-scoring judges on 27 matched candidate pairs from nine games and three generation runs. Each pair contains one Qwen3.8 Max-0902 candidate and one GPT-5.6 Terra candidate. We replace the text-scoring roles for D2 functionality, specification deviation, and D3 code quality, while holding the candidates, action compiler, evaluation contracts, evidence collection, and vision judge fixed. All 54 candidates have recorded $D1=100$.

\begin{table}[t]
    \centering
    \caption{Pairwise rankings of Qwen- and Terra-generated candidates under two text judges.}
    \label{tab:judge_rank_sensitivity}
    \small
    \setlength{\tabcolsep}{3pt}
    \renewcommand{\arraystretch}{1.05}

    \begin{tabular}{@{}lrrrrr@{}}
        \toprule
        Judge condition
        & Qwen wins
        & Terra wins
        & Ties
        & $p$-value
        & \shortstack{Qwen\\win rate} \\
        \midrule

        Qwen Judge
        & 13 & 14 & 0 & 1.000 & 48.1\% \\

        GPT-5.6 Sol Judge
        & 15 & 12 & 0 & 0.701 & 55.6\% \\

        \bottomrule
    \end{tabular}
\end{table}

\paragraph{Results.}
The Qwen-generated candidate ranks higher in 13/27 pairs under the Qwen Judge and 15/27 under the Sol Judge. The two conditions agree on 21/27 ranking directions (77.8\%), with six reversals. The exact McNemar test yields $p=0.688$; this does not establish equivalence between the judges or rule out family-related preference. The results characterize scoring-stage sensitivity on this fully executable, two-model subset. Candidate coverage and additional statistics are reported in Appendix~\ref{app:ablation_study_ii}.

\subsection{Human Assessment}
\label{sec:human_assessment}

\subsubsection{Human Assessment Protocol}
\label{sec:human_assessment_protocol}

To examine the correspondence between automatic evaluation and human judgments of specification realization and user-facing quality, we conduct a human assessment of 585 generated projects selected from the Core-15 experiments, covering all 15 game families and three implementation complexity levels. Five annotators assess their individually assigned projects, with a shared subset used for inter-annotator agreement analysis. For D2, the human assessment covers selected checkpoints using the same 0--2 scoring rubrics as the automatic evaluator. For D4, the human assessment provides complementary judgments of overall visual quality, dynamic feedback, and action selection, with scoring criteria and coverage that do not fully match those of the automatic D4 evaluation. Details of sample selection, task assignment, and scoring protocols are provided in Appendix~\ref{app:human_assessment}.

\subsubsection{Human Assessment Results}
\label{sec:human_assessment_results}

On the primary assignments, we compare candidate-level questionnaire aggregates with automatic scores for the same generated projects. Spearman correlations are $\rho=0.779$ for D2 ($N=243$) and $\rho=0.564$ for D4 ($N=295$), indicating positive correspondence in candidate rankings. Because the questionnaire assesses one visual action-choice question whereas official D4.3 aggregates two, we additionally align the automatic D4.3 contribution with the questionnaire's corresponding question. This sensitivity comparison yields $\rho=0.692$ for D4 on the same 295 candidates, highlighting the role of question coverage.


\section{Limitations and Future Work}
\label{sec:limitations}

Spec2Game mainly consists of classic small-scale games. Although controlled variants broaden coverage of rule configurations, the benchmark still does not cover the broader space of game families and game mechanics. Future work can further expand the task scale and mechanic diversity.

This work evaluates open-loop generation from detailed specifications to complete Pygame projects, without runtime feedback, semantic repair, or agentic coding workflows. Future work can study execution-guided debugging and iterative repair as separate settings.

D4 does not cover gameplay fun, novelty, or difficulty. In addition, parts of D2--D4 rely on text or vision judges and may still be affected by evaluator-specific bias and limited evidence coverage. Future work can further validate automatic evaluation with additional evaluator families and broader execution-grounded verification.

\section{Conclusion}
\label{sec:conclusion}

We introduce Spec2Game, a benchmark for generating complete interactive Pygame projects from detailed natural-language specifications, with canonical tasks, controlled rule variants, implementation-complexity levels, and four-dimensional evaluation. Across 14 LLMs and 3,330 generated projects, high executability does not imply faithful specification realization: several models nearly saturate D1 while scoring substantially lower on D2. Component-level analysis shows stronger performance on Game Element Modeling than on Rule and Mechanism Modeling or Goal and Termination Modeling. Thus, implementing game rules and termination logic remains a major challenge even for runnable projects. Evaluating complete interactive program generation therefore requires assessing how faithfully programs realize specified requirements alongside whether they execute successfully.

\subsection*{AI Use Statement}

We used generative AI tools to assist with research brainstorming, identifying relevant literature, translating game specifications into English, and drafting, revising, and polishing parts of the manuscript. AI tools also provided feedback and suggestions on game rules, evaluation criteria, and experimental design, as well as assistance with code implementation and refinement. All literature identified with AI assistance was independently reviewed by the authors. The benchmark framework, evaluation criteria, and experimental methodology were designed and finalized by the authors. All AI-assisted manuscript content and code were reviewed by the authors, who take full responsibility for the final content of this work.

\subsection*{Reproducibility Statement}

We release the benchmark specifications, evaluation code, and experiment scripts required to reproduce the main results of \textsc{Spec2Game} in a public repository.\footnote{\url{https://github.com/Aurora1094/Spec2Game-Bench}} The repository provides the benchmark instances, generation and evaluation pipelines, prompts, model configurations, and instructions for reproducing the reported experiments. Additional details on benchmark construction, evaluation protocols, and the human assessment are provided in the main text and appendix.

\begin{acks}
We thank Haoyan Zhang for assistance with running the model experiments, organizing the experimental data, and verifying the results.
\end{acks}

\bibliographystyle{ACM-Reference-Format}
\bibliography{software}

\appendix



\section{Implementation Complexity Rubric}
\label{app:complexity_rubric}

\subsection{Scoring Dimensions}

This section presents the complete scoring rubric for implementation complexity in Spec2Game. Scores are assigned solely based on the mechanisms explicitly required by the task specification and do not account for additional functionality introduced by the model. The rubric contains 13 sub-dimensions, each scored from 1 to 3, organized under five dimensions: Operation Complexity, Object and Rule Complexity, Spatial and Scene Complexity, Adversarial Behavior Complexity, and Interface and Information Complexity. Table~\ref{tab:implementation_complexity_appendix} provides an overview of the five dimensions and their corresponding sub-dimensions.

\begin{table*}[htbp]
\centering
\small
\setlength{\tabcolsep}{5pt}
\renewcommand{\arraystretch}{1.12}
\begin{tabular}{
    >{\raggedright\arraybackslash}p{0.22\textwidth}
    >{\centering\arraybackslash}p{0.15\textwidth}
    >{\raggedright\arraybackslash}p{0.53\textwidth}
}
\toprule
\textbf{Dimension} & \textbf{Sub-dimensions} & \textbf{Captures} \\
\midrule
Operation Complexity
& A1--A2
& Input actions, target localization, and operation flow \\
\addlinespace[5pt]

Object and Rule Complexity
& B1--B6
& State representation, rule updates, constraint checking, object interactions, progression, and termination conditions \\
\addlinespace[5pt]

Spatial and Scene Complexity
& C1--C2
& Spatial structure and dependencies involving layout and reachability \\
\addlinespace[5pt]

Adversarial Behavior Complexity
& D1
& Independent adversarial entities, their behavior patterns, and state dependence \\
\addlinespace[5pt]

Interface and Information Complexity
& E1--E2
& Organization of interface information and visual state representation \\
\bottomrule
\end{tabular}
\caption{Overview of the implementation complexity dimensions and sub-dimensions in Spec2Game.}
\label{tab:implementation_complexity_appendix}
\end{table*}

\paragraph{A. Operation Complexity.}

\leavevmode\par

\textbf{A1. Basic Input Action Complexity.}
This sub-dimension measures the number of basic action types that a player can directly issue. We count action types rather than individual keys; multiple keys implementing the same type of action are treated as one action type. For example, the four directional inputs in Snake all correspond to directional movement, whereas Space Invaders includes both movement and shooting.

\begin{itemize}
    \item \textbf{Score 1:} The game contains only one core basic action type, such as directional movement, clicking, or jumping.
    \item \textbf{Score 2:} The game contains multiple basic action types, but they mainly consist of simple and immediate actions, such as movement and shooting, or movement, rotation, and rapid dropping.
    \item \textbf{Score 3:} The action set is substantially richer and contains multiple distinct groups of actions, tools, or commands, such as movement, attacking, collecting, item use, skills, or turn-based commands.
\end{itemize}

\textbf{A2. Input Targeting and Interaction Flow Complexity.}
This sub-dimension measures whether, after receiving an input, the program must further determine the target location, target object, or current stage of an interaction. A2 does not consider the number of action types or the state updates triggered by the input.

\begin{itemize}
    \item \textbf{Score 1:} Inputs map directly to actions without requiring further identification of a target, location, or interaction stage.
    \item \textbf{Score 2:} Inputs require single-step targeting, where one input is sufficient to determine a specific object, grid cell, or adjacent target.
    \item \textbf{Score 3:} Inputs involve a multi-stage interaction flow, requiring the program to maintain intermediate states such as the current selection, interaction mode, valid targets, or the next interaction stage.
\end{itemize}

A1 measures how many types of action commands the player can issue, whereas A2 measures how the program must resolve the object, location, or interaction stage to which those inputs apply.

\paragraph{B. Object and Rule Complexity.}
\leavevmode\par

\textbf{B1. Object and State Representation Complexity.}
This sub-dimension measures the complexity of the object types and associated state representations that must be maintained internally by the game.

\begin{itemize}
    \item \textbf{Score 1:} Only a small number of simple objects and scalar states need to be maintained, without explicit object collections or structured state representations.
    \item \textbf{Score 2:} Multiple objects or object collections must be maintained using lists, grids, or multiple independent state variables.
    \item \textbf{Score 3:} Multiple object collections must be maintained together with clearly layered, persistent, or interdependent state structures, such as multiple entity types, resource states, map states, or phase states.
\end{itemize}

\textbf{B2. Rule Update Complexity.}
This sub-dimension measures the complexity of rule computation and state updates that must be performed after a player action or a game-loop update.

\begin{itemize}
    \item \textbf{Score 1:} State updates are simple, local, and largely linear, typically involving only a small number of direct changes.
    \item \textbf{Score 2:} An update requires multiple sequential steps or explicit conditional branches, but the overall update process remains relatively direct.
    \item \textbf{Score 3:} A single action may trigger complex algorithms, iterative processing, cascading updates, or multi-stage resolution, requiring multiple intermediate states to be processed.
\end{itemize}

\textbf{B3. Validity and Constraint Checking Complexity.}
This sub-dimension measures the number and complexity of conditions that must be checked before executing movement, placement, attacks, swaps, or other actions.

\begin{itemize}
    \item \textbf{Score 1:} Only simple boundary, collision, or basic validity checks are required.
    \item \textbf{Score 2:} Multiple explicit but relatively independent conditions must be checked, and these conditions can usually be evaluated through local and direct checks.
    \item \textbf{Score 3:} The validity of an action depends on multiple interrelated conditions and requires jointly considering information such as position, resources, current state, target state, or phase conditions.
\end{itemize}

B3 concerns whether an individual action is valid. Constraints involving overall path connectivity, access order, or global spatial reachability are captured by C2.

\textbf{B4. Object Interaction and Dependency Complexity.}
This sub-dimension measures interactions among objects, grid cells, resources, or system variables, and whether a single event can trigger state changes across multiple objects or systems. B2 concerns how rules update state, whereas B4 concerns whether those updates propagate across multiple objects or systems.

\begin{itemize}
    \item \textbf{Score 1:} Objects are largely independent, with only simple and local direct interactions.
    \item \textbf{Score 2:} Multiple objects or systems exhibit explicit dependencies, but the affected scope is usually local and the propagation pattern remains direct.
    \item \textbf{Score 3:} A single event may propagate across multiple objects, resources, or system variables, producing cascading or system-wide state changes.
\end{itemize}

\textbf{B5. Flow and Phase Progression Complexity.}
This sub-dimension measures whether the game must maintain multiple core gameplay phases and whether those phases have distinct states and transition rules. Start menus or Game Over screens do not increase this score by themselves, and the number of levels is not directly used as a scoring criterion.

\begin{itemize}
    \item \textbf{Score 1:} Core gameplay is primarily organized around a single loop and does not require meaningful phase-state management.
    \item \textbf{Score 2:} The game contains a limited number of fixed phase transitions, with relatively small rule differences between phases and a largely direct progression flow.
    \item \textbf{Score 3:} The game contains multiple core phases with distinct rules, permitted actions, or state-update procedures, and requires persistent maintenance of phase state.
\end{itemize}

\textbf{B6. Goal and Termination Determination Complexity.}
This sub-dimension measures how the program determines goal completion, victory, failure, or other termination conditions from the current game state. It evaluates only the decision logic and not the visual presentation of outcomes.

\begin{itemize}
    \item \textbf{Score 1:} The game has a single, direct goal or termination condition.
    \item \textbf{Score 2:} The game has multiple explicit goal, victory, or failure conditions, but each condition can generally be evaluated directly.
    \item \textbf{Score 3:} Termination requires jointly considering multiple global states, phase dependencies, condition priorities, or special cases.
\end{itemize}

\paragraph{C. Spatial and Scene Complexity.}
\leavevmode\par

\textbf{C1. Spatial Structure Complexity.}
This sub-dimension measures the organizational and representational structure of the game space itself.

\begin{itemize}
    \item \textbf{Score 1:} The game uses a simple continuous coordinate space without explicit structured spatial organization.
    \item \textbf{Score 2:} The game uses explicitly structured space, such as a fixed grid, platform structure, fixed path, or clearly partitioned scene.
    \item \textbf{Score 3:} The game uses multiple rooms, regions, or spatial structures with complex connectivity, requiring explicit topological or hierarchical relationships to be maintained.
\end{itemize}

\textbf{C2. Spatial Layout Dependency and Reachability Complexity.}
This sub-dimension measures whether spatial elements must satisfy adjacency, path, reachability, or access-order constraints.

\begin{itemize}
    \item \textbf{Score 1:} Spatial elements are largely independent, with no reachability or access-order constraints that materially affect core gameplay.
    \item \textbf{Score 2:} The game contains local adjacency, placement, or simple path constraints that can generally be evaluated through local checks.
    \item \textbf{Score 3:} The game contains global reachability, path-connectivity, access-order, or cross-region dependencies that require maintaining valid relationships among multiple spatial elements.
\end{itemize}

B3 concerns the validity of an individual action, whereas C2 concerns global layout, path, and reachability relationships among multiple spatial elements.

\paragraph{D. Adversarial Behavior Complexity.}
\leavevmode\par

\textbf{D1. Adversarial Behavior Complexity.}
This sub-dimension evaluates whether the game contains independent adversarial entities and the complexity of the behavior that these entities must implement. Ordinary obstacles, traps, or environmental hazards are not treated as adversarial entities unless they exhibit independent behavior logic.

\begin{itemize}
    \item \textbf{Score 1:} The game contains no independent adversarial entities; environmental obstacles or fixed hazards do not increase this score.
    \item \textbf{Score 2:} The game contains explicit enemies or adversarial entities, but their behavior is primarily governed by fixed paths, fixed rules, or simple scripts.
    \item \textbf{Score 3:} Adversarial entities select behavior based on the game state, for example by tracking the player's position, switching behavioral states, selecting targets, or choosing movement and attack behavior.
\end{itemize}

This sub-dimension measures adversarial behavior logic rather than the number of enemies, and it does not assess the amount of strategic reasoning required from the player.

\paragraph{E. Interface and Information Complexity.}
\leavevmode\par

\textbf{E1. Interface Information Organization Complexity.}
This sub-dimension measures how many types of game-state information must be organized and displayed simultaneously, and whether this information requires a structured interface layout.

\begin{itemize}
    \item \textbf{Score 1:} Only one or a small number of simple state indicators need to be displayed, such as a standalone score, health value, or timer.
    \item \textbf{Score 2:} Multiple types of game-state information must be displayed simultaneously and require basic layout organization, such as health, resources, phase information, or score.
    \item \textbf{Score 3:} The interface must organize multiple information regions, panels, menus, or selection states while simultaneously presenting resources, phases, goals, or other continuously changing information.
\end{itemize}

\textbf{E2. Visual Presentation Complexity.}
This sub-dimension measures the complexity of visual states and visual feedback explicitly required by the specification, rather than the aesthetic quality or subjective visual appeal of the final output.

\begin{itemize}
    \item \textbf{Score 1:} The game primarily uses simple geometric elements and a small number of fixed visual states, without requiring persistent visual representation of changing states.
    \item \textbf{Score 2:} Different objects or states must be distinguished using color, shape, orientation, simple animation, or other forms of visual encoding.
    \item \textbf{Score 3:} The game requires multiple dynamic, mechanism-related visual states and feedback signals that must remain synchronized with the internal game state, such as attacking, taking damage, selection, freezing, growth, or phase transitions.
\end{itemize}

\subsection{Game Family Complexity Scores}

Table~\ref{tab:family_complexity_scores} reports the sub-dimension scores and total implementation complexity score for each game family.

\begin{table*}[t]
\centering
\scriptsize
\setlength{\tabcolsep}{3.2pt}
\renewcommand{\arraystretch}{1.12}

\begin{tabular}{
    >{\raggedright\arraybackslash}p{0.20\textwidth}
    *{13}{>{\centering\arraybackslash}p{0.036\textwidth}}
    >{\centering\arraybackslash}p{0.055\textwidth}
}
\toprule
\textbf{Game} &
\textbf{A1} & \textbf{A2} &
\textbf{B1} & \textbf{B2} & \textbf{B3} & \textbf{B4} & \textbf{B5} & \textbf{B6} &
\textbf{C1} & \textbf{C2} &
\textbf{D1} &
\textbf{E1} & \textbf{E2} &
\textbf{Total} \\
\midrule

Snake
& 1 & 1 & 2 & 2 & 2 & 2 & 1 & 1 & 2 & 1 & 1 & 1 & 1 & 18 \\

Dodge Blocks
& 1 & 1 & 2 & 1 & 1 & 1 & 1 & 1 & 1 & 1 & 1 & 1 & 1 & 14 \\

Whack-a-Mole
& 1 & 2 & 2 & 2 & 1 & 1 & 1 & 1 & 2 & 1 & 1 & 1 & 1 & 17 \\

Asteroids
& 2 & 1 & 2 & 2 & 1 & 2 & 1 & 2 & 1 & 1 & 1 & 1 & 1 & 18 \\

Maze Runner
& 1 & 1 & 2 & 2 & 1 & 1 & 1 & 2 & 2 & 2 & 1 & 1 & 1 & 18 \\

\midrule

Space Invaders
& 2 & 1 & 2 & 2 & 1 & 2 & 1 & 2 & 1 & 1 & 2 & 1 & 1 & 19 \\

Simple Platformer
& 2 & 1 & 2 & 2 & 2 & 2 & 1 & 2 & 2 & 1 & 1 & 1 & 1 & 20 \\

Tetris
& 2 & 1 & 2 & 3 & 2 & 2 & 1 & 1 & 2 & 1 & 1 & 1 & 1 & 20 \\

Match-3
& 1 & 2 & 2 & 3 & 2 & 3 & 1 & 3 & 2 & 1 & 1 & 1 & 2 & 24 \\

Simple Shop Management
& 1 & 2 & 3 & 2 & 2 & 3 & 1 & 1 & 1 & 1 & 1 & 3 & 2 & 23 \\

\midrule

Carrot Defense-like
& 2 & 3 & 3 & 3 & 3 & 3 & 3 & 3 & 2 & 2 & 2 & 3 & 2 & 34 \\

Farming-lite
& 3 & 3 & 3 & 3 & 3 & 3 & 3 & 2 & 2 & 1 & 1 & 3 & 2 & 32 \\

Pipe Connection Puzzle
& 2 & 2 & 3 & 3 & 3 & 3 & 2 & 3 & 2 & 3 & 1 & 2 & 2 & 31 \\

Top-down Arena Shooter
& 3 & 2 & 3 & 3 & 3 & 3 & 2 & 3 & 2 & 3 & 3 & 2 & 2 & 34 \\

Mini Zelda-like Room Adventure
& 2 & 1 & 3 & 3 & 3 & 3 & 3 & 3 & 3 & 3 & 3 & 2 & 2 & 34 \\

\bottomrule
\end{tabular}

\caption{Implementation complexity scores for the 15 game families.}
\label{tab:family_complexity_scores}
\end{table*}

\section{Game Specifications}
\label{app:game_specifications}

This section provides additional details on the structure and construction of game specifications in Spec2Game. All benchmark instances follow a unified seven-part specification template and present task requirements as complete, standalone natural-language descriptions.

\subsection{Specification Template}

Each specification is organized into the following seven sections:

\begin{enumerate}
    \item \textbf{Game Overview.}
    Summarizes the core gameplay, including the player-controlled entity, primary actions, core gameplay loop, game objective, and basic termination conditions.

    \item \textbf{Game Scene \& Interface.}
    Describes the game window, primary gameplay area, information display regions, initial screen, and basic spatial layout of the interface.

    \item \textbf{Game Objects \& States.}
    Defines the main game objects, their roles, key attributes, initial states, and possible state changes.

    \item \textbf{Player Interaction.}
    Describes the available player inputs and how different inputs affect game objects or modify the game state.

    \item \textbf{Core Rules \& Mechanics.}
    Specifies the core game rules and mechanics, including movement, collision, spawning, removal, attacks, collection, resource changes, and other game-specific state transition rules.

    \item \textbf{Game Flow \& Progression.}
    Describes the runtime loop and how the game state progresses over time, player input, phases, or other progression factors.

    \item \textbf{Goal, Feedback \& Termination.}
    Specifies player goals, game feedback, constraints, victory and failure conditions, terminal states, and restart behavior.
\end{enumerate}

The unified template standardizes how information is organized, while the specific objects, rules, states, and interactions remain defined by each individual game.

\subsection{Variant Construction}

For each game family, we first construct a base task and write its corresponding base specification. We then construct nine variants from this base task. Each variant introduces one primary change to a rule or design factor, such as speed, map layout, object parameters, resource conditions, or termination conditions. When this change affects other parts of the specification, the relevant content is updated accordingly to maintain internal consistency; these resulting updates are not treated as additional independent variation factors.

All variants preserve the core gameplay of their game family and are rewritten as complete standalone specifications rather than being expressed as incremental modifications to the base task. As a result, each variant specification can be understood independently without referring to the corresponding base specification.

\subsection{Representative Specifications}

We provide two representative specifications to illustrate how the unified seven-part template is instantiated in benchmark tasks with different levels of implementation complexity. The complete specifications for all 150 benchmark instances are released with the benchmark artifacts.

\noindent\textbf{Example 1: Snake (Easy).}
\par\smallskip

\begin{lstlisting}[
    basicstyle=\ttfamily\footnotesize,
    breaklines=true,
    columns=fullflexible,
    keepspaces=true,
    showstringspaces=false,
    frame=single,
    backgroundcolor=\color{gray!8},
    rulecolor=\color{gray!40},
    framesep=6pt
]

# Game Specification

## Game Name
Snake

---



# 1. Game Overview
Snake is a single-player grid-based game. The player uses W, A, S, and D to control the snake's direction while it continuously moves through the main gameplay area.
The objective is to collect randomly spawned food while avoiding self-collision and the lethal left and right boundaries.

Each time the snake collects food, its length increases by 1, the score increases by 1, and a new food item is randomly spawned on an unoccupied grid cell.
The game has no fixed victory score, completion condition, or victory state. The player's objective is to survive for as long as possible and achieve the highest possible score.

Unlike conventional Snake games, this game uses asymmetric boundary behavior: crossing the top or bottom boundary wraps the snake to the opposite side, while crossing the left or right boundary immediately ends the game.

---



# 2. Game Scene & Interface
The game window is 600 × 800 px and uses a vertical layout. From top to bottom, the window is divided into three areas:
1. Top information area: 600 × 100 px
2. Main gameplay area: 600 × 600 px
3. Bottom instruction area: 600 × 100 px

## 2.1 Top Information Area
The top information area displays the current score in the fixed format `Score: X`, where X is the current score.
At the start of the game, it displays `Score: 0`.

The score text uses a font size of 40 px and should be clearly visible. It may be horizontally centered or left-aligned within the top information area.

## 2.2 Main Gameplay Area
The main gameplay area is located directly below the top information area and has a size of 600 × 600 px. It is divided into a 30 × 30 grid, with each grid cell measuring 20 × 20 px.
Both the snake and food must remain aligned with the grid.

The main gameplay area uses a light-colored background with moderately visible grid lines so that individual cells can be clearly distinguished.
The snake is drawn using orange blocks, and the food is drawn using a brown block.

At the start of the game:
- Snake length: 3
- Initial movement direction: right
- Food: exactly one item randomly spawned on a grid cell not occupied by the snake

## 2.3 Bottom Instruction Area
While the game is in the `playing` state, the bottom instruction area displays:
`Move: WASD`
`Top/bottom wrap, left/right walls kill`

After the game enters the `game_over` state, it displays:
`Game Over`
`Press R to Restart`

Both lines use a font size of 28 px and should be centered within the bottom instruction area.
The top information area and bottom instruction area should be visually distinct from the main gameplay area and may use light or neutral background colors. All text should use a dark or otherwise high-contrast color for readability.

---



# 3. Game Objects & States
The main game objects and state variables include the snake, food, gameplay grid, score, and game state.

## 3.1 Snake
The snake is the primary player-controlled object and consists of contiguous grid cells.
Its initial properties are:
- Length: 3
- Movement direction: right

The snake state includes the head position, ordered body-cell positions, current movement direction, and alive/dead state.
During each movement update, the head moves exactly one grid cell in the current direction and the body positions are updated accordingly.

If food is collected, the snake's length increases by 1.
If the snake crosses the left or right boundary or collides with a body cell that remains occupied after the current movement update, it dies and the game enters `game_over`.

## 3.2 Food
Each food item occupies exactly one grid cell, and only one food item may exist at any time.
Food is spawned once during game initialization and immediately after the previous food item is collected.

Every newly spawned food item must:
- Be inside the 30 × 30 gameplay grid.
- Occupy a grid cell not occupied by the snake after the current movement update.

The game does not include special food, timed food, bonus food, or other food types.

## 3.3 Gameplay Grid
The gameplay grid contains 30 columns × 30 rows, with each cell measuring 20 × 20 px.
Both the snake and food must always remain aligned with grid cells, and each movement step advances the snake head exactly one cell.

## 3.4 Score
Score records the number of food items collected.
The initial score is 0. Each valid food collection increases the score by 1 and immediately updates the top information display to `Score: X`.

## 3.5 Game State
The game contains two states:
- `playing`
- `game_over`

The initial state is `playing`.
The game enters `game_over` if the snake crosses the left or right boundary or collides with its own body according to the self-collision rule in Section 5.4.

After entering `game_over`, the snake stops moving, movement input no longer affects gameplay, and only R is accepted to restart.

---



# 4. Player Interaction
The player uses W, A, S, D, and R.

## 4.1 Direction Control
The movement controls are:
- W: up
- S: down
- A: left
- D: right

The snake continuously moves in its current direction, so the player does not need to press a movement key for every step.
After a valid directional input is received, the new direction takes effect on the next movement update.

## 4.2 Opposite-Direction Input
The snake cannot directly reverse its movement direction.
An input opposite to the current movement direction is ignored:
- While moving right, A is ignored.
- While moving left, D is ignored.
- While moving up, S is ignored.
- While moving down, W is ignored.

When an opposite-direction input is ignored, the snake continues moving in its current direction.

## 4.3 Restart
After the game enters `game_over`, pressing R restarts the game.
Restart fully resets the snake, food, score, interface state, and game state to their initial conditions.

---



# 5. Core Rules & Mechanics

## 5.1 Movement
The snake moves at a fixed speed of 8 steps per second. Each movement step advances the snake head exactly one grid cell.
The movement speed remains constant throughout the game and does not change according to score, elapsed time, or snake length.

If the snake does not collect food during a movement step, the final tail cell is removed so that the snake's length remains unchanged.
If the snake collects food, the tail is retained, causing the snake's length to increase by 1.

## 5.2 Food Collection
Food is collected when the snake head enters the grid cell occupied by the food.
When food is collected:
- Score increases by 1.
- Snake length increases by 1.
- The collected food is removed.
- One new food item is randomly spawned on a valid unoccupied grid cell.

Only one normal food item may exist at any time.

## 5.3 Boundary Rules
The game uses asymmetric boundary behavior:
- Top boundary: the snake re-enters from the bottom boundary in the same column.
- Bottom boundary: the snake re-enters from the top boundary in the same column.
- Left boundary: crossing it immediately causes `game_over`.
- Right boundary: crossing it immediately causes `game_over`.

Crossing the top or bottom boundary never causes Game Over.

## 5.4 Self-Collision
Self-collision is checked against the body cells that remain occupied after the current movement update.

If the current movement does not collect food, the final tail cell is removed during the same update. That cell is therefore excluded from self-collision detection, and the snake head may move into it without causing Game Over.

If the current movement collects food, the tail is retained. The current tail cell therefore remains occupied and is included in self-collision detection.

A self-collision occurs when the new head position overlaps any body cell that remains occupied after the movement update.

## 5.5 Movement-Step Evaluation Order
Each movement step follows this logical order:
1. Calculate the new head position from the current movement direction.
2. Apply top/bottom wrapping if necessary.
3. Check whether the head has crossed the left or right boundary. If so, immediately enter `game_over`.
4. Determine whether the new head position reaches the food.
5. Determine which body cells remain occupied after the movement:
   - If food is collected, retain the tail.
   - Otherwise, treat the final tail cell as vacated.
6. Check the new head position against the body cells that remain occupied. If a self-collision occurs, enter `game_over`.
7. Update the snake body by adding the new head and retaining or removing the tail as determined above.
8. If food was collected, increase the score by 1, remove the collected food, and spawn one new valid food item.

If a terminal collision occurs during Step 3 or Step 6, no food collection, score increase, growth, or new food spawning is processed for that movement step.

---



# 6. Game Flow & Progression
At initialization, the game creates the window, interface areas, gameplay grid, snake, food, score, and game state.
The initial state is:
- Snake length: 3
- Snake direction: right
- Score: 0
- Food: one randomly spawned valid item
- Game state: `playing`

While `playing`, the program continuously processes directional input and updates the snake at a fixed rate of 8 movement steps per second.
Each movement update follows the boundary, food-collection, self-collision, and body-update rules defined in Section 5.

When the snake collects food, the score and snake length each increase by 1 and a new food item is spawned.
When the snake crosses the top or bottom boundary, it wraps to the corresponding position on the opposite side.
When the snake crosses the left or right boundary or collides with a body cell that remains occupied after the movement update, the game enters `game_over`.

After entering `game_over`:
- The snake stops moving.
- The final score is preserved.
- Movement inputs no longer affect gameplay.
- The bottom instruction area displays `Game Over` and `Press R to Restart`.
- R remains available to restart.

The movement speed remains fixed throughout the game. The game does not include levels, increasing difficulty, acceleration, or progression stages.
Progression is represented only by increasing score and snake length as food is collected.

---



# 7. Goal, Feedback & Termination
The player's goal is to collect as much food as possible before the game ends and achieve the highest possible score.
The game has no fixed completion score, victory threshold, or victory state.

The top information area displays the current score in real time using `Score: X`, starting from `Score: 0`.
Each food collection immediately increases X by 1.

During `playing`, the bottom instruction area displays:
`Move: WASD`
`Top/bottom wrap, left/right walls kill`

After entering `game_over`, it displays:
`Game Over`
`Press R to Restart`

The game ends when:
- The snake head crosses the left or right boundary.
- The snake head moves into a body cell that remains occupied after the current movement update.

Moving into the current final tail cell does not cause Game Over when that tail cell is removed during the same update.
Crossing the top or bottom boundary does not end the game and instead wraps the snake to the corresponding position on the opposite side.

After `game_over`, pressing R restores the snake to its initial length and configuration, resets the direction to right and score to 0, generates one new valid food item, restores the normal gameplay instructions, and returns the game state to `playing`.

---





\end{lstlisting}

\noindent\textbf{Example 2: Mini Zelda-like Room Adventure (Hard).}
\par\smallskip

\begin{lstlisting}[
    basicstyle=\ttfamily\footnotesize,
    breaklines=true,
    columns=fullflexible,
    keepspaces=true,
    showstringspaces=false,
    frame=single,
    backgroundcolor=\color{gray!8},
    rulecolor=\color{gray!40},
    framesep=6pt
]
# Game Specification

## Game Name
Mini Zelda-like Room Adventure

---



# 1. Game Overview
Mini Zelda-like Room Adventure is a top-down spatial exploration and action-adventure game. The player explores a fixed map consisting of 5 rooms, moves between rooms through doors, defeats enemies, obtains Keys, unlocks new areas, and ultimately reaches the final room to obtain the treasure.

The fixed room roles are:
- Room A: starting room
- Room C: central connecting room
- Room B: Key 1 combat room
- Room D: Key 2 combat room
- Room E: final treasure room

The intended progression is:
`Room A → Room C → Room B → obtain Key 1 → Room C → unlock Room D → obtain Key 2 → Room C → unlock Room E → open the treasure chest → Victory`

The player uses WASD for movement and Space for a short-range melee attack in the current facing direction.

The game contains two enemy types:
- Slime: continuously tracks the player.
- Guard: initially patrols along a fixed vertical route and permanently enters chase once the player comes within range.

Room state is persistent. Defeated enemies do not respawn when the player leaves and re-enters a room.
Spawned or collected Keys, unlocked doors, surviving enemy HP and positions, Guard behavior states and patrol directions, and the treasure chest state are preserved across room transitions.

A custom rule is the sequential dependency between Key 1 and Key 2. The player must first clear Room B to obtain Key 1 and unlock Room D, then clear Room D to obtain Key 2 and unlock Room E.

The game does not include random maps, random enemies, random drops, multiple floors, camera scrolling, a minimap, fog of war, ranged attacks, magic, bombs, blocking, rolling, dashing, stamina, Mana, equipment, weapon switching, Inventory, Gold, shops, XP, upgrades, NPCs, dialogue, Bosses, saving, Checkpoints, healing items, automatic healing, a time limit, Score, Lives, enemy respawning, knockback, or complex pathfinding.

---



# 2. Game Scene & Interface
The game window is 800 × 700 px. From top to bottom, it contains:
1. Top information area: 800 × 80 px
2. Main room area: x = 50–750, y = 100–600
3. Bottom instruction area: 800 × 80 px

Every room uses the same 700 × 500 px main room area.
Room transitions occur instantly, without scrolling or camera animation.

## 2.1 Top Information Area
The top area displays:

| Item | Position | Initial Display |
|---|---|---|
| HP | `(30, 28)` | `HP: 5/5` |
| Keys | `(200, 28)` | `Keys: 0` |
| Room | `(360, 28)` | `Room: A` |
| Status | `(560, 28)` | `Status: Playing` |

All text uses a font size of 28 px.

Room displays:
- A
- B
- C
- D
- E

Status displays:
- `Playing`
- `Victory`
- `Game Over`

## 2.2 Bottom Instruction Area
During `playing`, the bottom area displays:
`WASD: Move    Space: Attack`
`Explore rooms, find keys, and reach the treasure`

Both lines use a font size of 22 px and should be centered.

During `victory`:
`Treasure Found!`
`Press R to Restart`

During `game_over`:
`Game Over`
`Press R to Restart`

The result line uses a font size of 36 px.
`Press R to Restart` uses a font size of 26 px.

## 2.3 Visual Objects
Player:
- Blue circle
- Radius: 18 px
- Approximate 20 px facing-direction line

Slime:
- Green circle
- Radius: 16 px

Guard:
- Visual size: 32 × 32 px purple square
- Logical collision shape: circle with radius 16 px

Each living enemy displays a 32 × 5 px HP bar.

Keys use clearly visible gold key graphics.
Fixed obstacles use dark-gray rectangles.

Normal doors are displayed as open entrances.
Locked doors are visually distinct closed barriers with a lock symbol.

The final treasure chest uses a brown or gold appearance and has two visually distinct states:
- `closed`
- `opened`

---



# 3. Game Objects & States
The main game objects and state variables include the player, Slimes, Guards, obstacles, doors, Key 1, Key 2, treasure chest, HP, Keys, attack state, damage invulnerability, room state, and game state.

## 3.1 Player
The player has:
- Collision shape: circle
- Radius: 18 px
- Movement speed: 200 px/s
- Maximum HP: 5
- Initial HP: 5
- Initial facing: `right`
- Initial room: Room A
- Initial center: `(180, 350)`

## 3.2 Room Connection Structure
The fixed room structure is:

```text
          Room B
            |
Room A — Room C — Room D
            |
          Room E
```

Door relationships:
- Room A ↔ Room C: normal
- Room B ↔ Room C: normal
- Room C ↔ Room D: initially locked, requires Key 1
- Room C ↔ Room E: initially locked, requires Key 2

## 3.3 Room A
Room A is the starting room.

Right-side door to Room C:
- Boundary: x = 750
- Opening: y = 320–380

Returning from Room C places the player at:
`(705, 350)`

Obstacles:

| Obstacle | Top-Left Coordinate | Size |
|---|---|---|
| A1 | `(290, 220)` | `100 × 60 px` |
| A2 | `(420, 420)` | `120 × 60 px` |

Room A contains no enemies, Keys, locked doors, or treasure chest.

## 3.4 Room B
Room B is the Key 1 combat room.

Bottom door to Room C:
- Boundary: y = 600
- Opening: x = 370–430

Entering from Room C places the player at:
`(400, 555)`

Initial enemies:
- Slime B1: `(220, 250)`
- Slime B2: `(580, 250)`

Obstacles:

| Obstacle | Top-Left Coordinate | Size |
|---|---|---|
| B1 | `(300, 300)` | `80 × 120 px` |
| B2 | `(420, 300)` | `80 × 120 px` |

A 40 px passage remains between the two obstacles.

After both Slimes are defeated, Key 1 spawns at:
`(400, 350)`

## 3.5 Room C
Room C is the central connecting room and contains no enemies.

Doors:
- Left to Room A: x = 50, y = 320–380
- Top to Room B: y = 100, x = 370–430
- Right to Room D: x = 750, y = 320–380, initially locked by Key 1
- Bottom to Room E: y = 600, x = 370–430, initially locked by Key 2

Obstacles:

| Obstacle | Top-Left Coordinate | Size |
|---|---|---|
| C1 | `(250, 220)` | `80 × 60 px` |
| C2 | `(470, 220)` | `80 × 60 px` |
| C3 | `(250, 420)` | `80 × 60 px` |
| C4 | `(470, 420)` | `80 × 60 px` |

Spawn positions when entering Room C:
- From Room A: `(95, 350)`
- From Room B: `(400, 145)`
- From Room D: `(705, 350)`
- From Room E: `(400, 555)`

## 3.6 Room D
Room D is the Key 2 combat room.

Left door to Room C:
- Boundary: x = 50
- Opening: y = 320–380

Entering from Room C places the player at:
`(95, 350)`

Initial enemies:
- Slime D1: `(600, 450)`
- Guard D1: `(320, 220)`
- Guard D2: `(320, 480)`

Obstacles:

| Obstacle | Top-Left Coordinate | Size |
|---|---|---|
| D1 | `(410, 190)` | `70 × 110 px` |
| D2 | `(410, 400)` | `70 × 110 px` |
| D3 | `(535, 310)` | `90 × 80 px` |

After all 3 enemies are defeated, Key 2 spawns at:
`(600, 240)`

## 3.7 Room E
Room E is the final room and contains no enemies.

Top door to Room C:
- Boundary: y = 100
- Opening: x = 370–430

Entering from Room C places the player at:
`(400, 145)`

Obstacles:

| Obstacle | Top-Left Coordinate | Size |
|---|---|---|
| E1 | `(220, 260)` | `100 × 60 px` |
| E2 | `(480, 260)` | `100 × 60 px` |

Treasure chest:
- Center: `(400, 480)`
- Size: 60 × 45 px
- Initial state: `closed`

## 3.8 Game State
The game contains:
- `playing`
- `victory`
- `game_over`

Both `victory` and `game_over` are terminal states.

---



# 4. Player Interaction
The player uses W, A, S, D, Space, and R.

## 4.1 Movement
Controls:
- W: up
- S: down
- A: left
- D: right

Opposite inputs on the same axis cancel.

If one valid horizontal and one valid vertical direction are active simultaneously, normalize the movement vector so that diagonal speed remains:
`200 px/s`

The player physically collides with:
- Room outer walls
- Fixed obstacles
- Closed locked doors
- Closed treasure chest

Movement collision uses the player's complete circle.

Axis order:
1. Process x movement.
2. Cancel x displacement if blocked.
3. Process y movement.
4. Cancel y displacement if blocked.

## 4.2 Facing
The player has four facing directions:
- `up`
- `down`
- `left`
- `right`

Initial facing:
`right`

Facing updates according to the most recently pressed valid directional key and remains unchanged when all movement keys are released.

For diagonal input, the most recently pressed valid direction determines facing.

If one horizontal and one vertical direction are first pressed simultaneously in the same frame and order cannot be determined, vertical input has priority:
- W + A → up
- W + D → up
- S + A → down
- S + D → down

## 4.3 Melee Attack
Space triggers an attack only on a new key-down event.
Holding Space does not repeatedly trigger attacks.

Attack duration:
`0.20 seconds`

Minimum interval between attack starts:
`0.45 seconds`

When an attack begins:
- Lock the current facing direction.
- Keep that direction for the entire attack.

The player may continue moving during the attack.
The attack area follows the player's current center position while keeping the locked direction.

---



# 5. Core Rules & Mechanics

## 5.1 Sword Attack Area
The sword uses a rectangular logical attack area.

Dimensions:
- Length along attack direction: 42 px
- Width perpendicular to attack direction: 34 px

The rectangle begins at the edge of the player's circular collision shape and extends forward.

Each frame during an active attack:
- Recalculate the attack rectangle from the player's current position.
- Keep the attack direction locked to the direction recorded at attack start.

The visual effect may use a rectangle, arc, or sword graphic, but logical collision uses the defined rectangle.

## 5.2 Sword Damage
Each attack can damage each enemy at most once during its full 0.20-second duration.

A single attack:
- May damage multiple different enemies.
- Deals exactly 1 HP to each successfully hit enemy.
- Does not damage the player.
- Does not damage obstacles.
- Does not affect doors.
- Does not affect Keys.
- Does not open the treasure chest.

Enemies do not block the sword from reaching other enemies.

## 5.3 Sword Blocking
Room walls and fixed obstacles block sword attacks.

For the locked attack direction, the effective attack area is clipped at the nearest blocking wall or fixed obstacle boundary.

An enemy behind a blocking obstacle cannot be damaged even if it overlaps the unclipped attack rectangle.

## 5.4 Slimes
Each Slime has:
- Radius: 16 px
- HP: 2
- Movement speed: 90 px/s
- Contact damage: 1 HP

Each frame, a Slime calculates the normalized direction toward the player's current center and moves at 90 px/s.

Movement uses x-first, y-second collision against room walls and fixed obstacles.
If one axis is blocked, the other may continue.

Slimes do not use A* or other complex pathfinding.

## 5.5 Guards
Each Guard has:
- Visual size: 32 × 32 px
- Logical radius: 16 px
- HP: 3
- Patrol speed: 70 px/s
- Chase speed: 110 px/s
- Contact damage: 1 HP

Guard behavior states:
- `patrol`
- `chase`

Guard D1:
- Initial center: `(320, 220)`
- Patrol x: 320
- Patrol y range: 170–300
- Initial direction: down

Guard D2:
- Initial center: `(320, 480)`
- Patrol x: 320
- Patrol y range: 400–530
- Initial direction: up

During `patrol`, the Guard moves vertically along its fixed route and reverses immediately at the endpoints.

If:
`distance(guard_center, player_center) ≤ 180 px`

the Guard immediately enters `chase`.

Once a Guard enters `chase`, it never returns to `patrol`.

During `chase`, it moves toward the player at:
`110 px/s`

Chase movement uses x-first, y-second collision.
If blocked, the Guard does not search for an alternative route.

## 5.6 Enemy Cross-Room & Physical Collision Rules
Enemies remain permanently assigned to their original rooms.

They:
- Cannot use doors.
- Cannot transition between rooms.
- Remain constrained by the full room boundary.

The player and enemies do not physically block one another.
Enemies also do not physically block one another.

Therefore:
- Slime-Slime overlap is allowed.
- Slime-Guard overlap is allowed.
- Guard-Guard overlap is allowed.
- Player-enemy overlap is allowed.

Character-character overlap is used only for contact damage.

## 5.7 Player Damage & Invulnerability
The player starts with:
`HP = 5`

Slime or Guard contact deals:
`1 HP`

After valid damage:
- HP decreases by 1.
- HP is clamped to a minimum of 0.
- Start 0.8 seconds of damage invulnerability.

During invulnerability:
- Additional enemy contact cannot reduce HP.
- The player flashes between normal and semi-transparent display every approximately 0.1 seconds.

The player can lose at most 1 HP per frame.

If invulnerability ends while the player is still touching an enemy, contact may deal damage again.

## 5.8 Enemy Hits & Death
Sword hits deal:
`1 HP`

Enemy HP:
- Slime: 2
- Guard: 3

After a valid hit, the enemy displays a brief hit highlight for approximately 0.1 seconds.

When enemy HP reaches 0:
- Remove the enemy immediately from active objects.
- Stop movement.
- Stop contact damage.
- Stop collision participation.
- Preserve its dead state permanently until Restart.

If one attack kills multiple enemies, each death is processed independently.

## 5.9 Key State
Key 1 and Key 2 independently track:
- Whether they have spawned.
- Whether they have ever been collected.
- Whether they are currently held.

The persistent state distinguishes at least:
- `key1_collected`
- `key1_held`
- `door_d_unlocked`
- `key2_collected`
- `key2_held`
- `door_e_unlocked`

`collected` prevents a Key from spawning again.
`held` indicates current possession.

When a Key is used:
- `held = false`
- `collected` remains true

The top `Keys` value counts only currently held, unconsumed Keys.

## 5.10 Key Spawning
Key 1:
- Spawn condition: both Room B Slimes are dead
- Position: `(400, 350)`

Key 2:
- Spawn condition: Room D Slime and both Guards are dead
- Position: `(600, 240)`

Each Key:
- Collision rectangle: 28 × 28 px
- Centered at its fixed spawn coordinate
- Remains stationary after spawning

If the player already overlaps the Key collision area in the same frame the Key spawns, it may be collected immediately.

## 5.11 Key Collection
A Key is collected automatically when the player's circle overlaps its 28 × 28 px rectangle.

Collecting Key 1:
- Remove Key 1 from Room B.
- `key1_collected = true`
- `key1_held = true`
- Keys += 1

Collecting Key 2:
- Remove Key 2 from Room D.
- `key2_collected = true`
- `key2_held = true`
- Keys += 1

Keys do not block the player, enemies, or sword attacks.

## 5.12 Locked Doors
The C-D door requires Key 1.
The C-E door requires Key 2.

When the player attempts to pass through a locked door, the corresponding `held` state is checked before physical blocking.

If the required Key is not held:
- Keep the door locked.
- Treat the door as a physical barrier.
- Do not allow transition.
- Keep Keys unchanged.
- Display `Locked`.

If the required Key is held:
- Consume the Key.
- Set the corresponding `held = false`.
- Decrease Keys by 1.
- Set the corresponding door unlocked.
- Permit transition through that door during the same frame.

The actual room transition is still performed during the unified room-transition stage.

After unlocking:
- The door remains permanently open.
- It is passable in both directions.
- The Key is never checked or consumed again.

Keys cannot become negative.

## 5.13 Locked Feedback
`Locked` uses a font size of 24 px.

It appears approximately 40 px inward from the blocked doorway:
- Right-side door → message appears to the left
- Bottom-side door → message appears above

The message remains for approximately 1 second.
Continuously pushing against the same locked door does not need to restart the timer every frame.

## 5.14 Terminal Priority
Victory is marked when the treasure chest is successfully opened.
Game Over occurs when:
`HP = 0`

Terminal priority is:
1. Game Over
2. Victory
3. Continue

Therefore, if HP reaches 0 and the final treasure is obtained during the same frame, the game enters `game_over`.

---



# 6. Game Flow & Progression

## 6.1 Room Transitions
Only the player can transition between rooms.

All doorway openings are 60 px wide.

Left/right doors:
- Opening: y = 320–380
- Safe player-center range: y = 338–362

Top/bottom doors:
- Opening: x = 370–430
- Safe player-center range: x = 388–412

A transition occurs only if:
1. The player's complete circle fits within the doorway.
2. The player's outer edge reaches or attempts to cross the room boundary.
3. Valid movement input points outward through the door.
4. The door is passable.

Merely standing inside the doorway does not trigger transition.

Spawn mapping:

| Transition | Target Spawn |
|---|---|
| Room A → Room C | `(95, 350)` |
| Room C → Room A | `(705, 350)` |
| Room C → Room B | `(400, 555)` |
| Room B → Room C | `(400, 145)` |
| Room C → Room D | `(95, 350)` |
| Room D → Room C | `(705, 350)` |
| Room C → Room E | `(400, 145)` |
| Room E → Room C | `(400, 555)` |

After a successful transition:
- Set the player to the target spawn point.
- Cancel remaining player movement for that frame.
- Prevent a second transition during the same frame.

If the player is attacking:
- End the current attack immediately.
- Preserve the existing 0.45-second attack cooldown.

## 6.2 Persistent Room State
Across transitions, preserve:
- Player HP
- Keys
- Key collected/held states
- Door unlocked states
- Damage invulnerability and remaining duration
- Attack cooldown
- Enemy deaths
- Surviving enemy HP
- Surviving enemy positions
- Guard patrol/chase state
- Guard patrol direction
- Spawned but uncollected Keys
- Treasure chest state

Enemy hit highlights are purely visual and are cleared when leaving the room.

## 6.3 Non-Current Room Freeze
Only the current room updates in real time.

When the player leaves a room:
- Surviving enemies stop moving.
- Enemy HP remains unchanged.
- Enemy positions remain unchanged.
- Guard state remains unchanged.
- Guard patrol direction remains unchanged.
- No background AI continues.

When the player returns, those objects resume from their stored state.

After a successful room transition, the previous room receives no further updates during that frame.
Enemies in the previous room cannot deal contact damage after the transition.

## 6.4 Treasure Chest & Victory Trigger
The final treasure chest in Room E:
- Center: `(400, 480)`
- Size: 60 × 45 px
- Initial state: `closed`

While closed, the chest is a physical obstacle.

If the player's proposed movement first touches or overlaps the closed chest rectangle:
- Block movement into the chest.
- Record `treasure_contact` for the current frame.

The player's final position does not need to overlap the chest.

Later in the same frame, if `treasure_contact` is recorded:
- Set chest state to `opened`.
- Grant the final treasure.
- Mark Victory as a candidate terminal result.

Sword attacks cannot open the chest.
The opened chest remains visible.

## 6.5 Fixed Frame Update Order
While `playing`, each frame follows this fixed order:

1. Process directional input, movement direction, and facing.
2. Process a new Space key-down event and begin an attack if the attack cooldown allows it.
3. Process player x/y movement and collision:
   - Room outer walls
   - Fixed obstacles
   - Locked doors
   - Closed treasure chest
   - For locked doors, check required Key before applying physical blocking.
   - For the chest, block entry and record `treasure_contact`.
4. Check and execute room transition:
   - Normal doors
   - Previously unlocked doors
   - Doors unlocked through Key consumption during this frame
   - If transition succeeds, cancel remaining movement and terminate the current attack while preserving its cooldown.
5. If the attack remains active, update the attack area and process enemy hits, HP reduction, and deaths.
6. Check combat-room completion and spawn the corresponding Key if required.
7. Check player-Key overlap and collect Keys.
8. Update AI and movement for surviving enemies in the current room.
9. Process enemy contact damage using the 0.8-second invulnerability rule, with at most 1 HP lost during the frame.
10. If `treasure_contact` was recorded, open the chest and mark Victory.
11. Update attack cooldown, damage invulnerability, player flashing, `Locked` feedback, and other visual timers.
12. Determine the terminal result using Game-Over-before-Victory priority.

An enemy killed in Step 5 does not participate in Step 8 AI or Step 9 contact damage.

Terminal resolution in Step 12:
1. If HP = 0, enter `game_over`.
2. Otherwise, if Victory was marked in Step 10, enter `victory`.
3. Otherwise, remain `playing`.

## 6.6 Normal Progression
The game begins in Room A.

Expected progression:
1. Move from Room A to Room C.
2. Enter Room B.
3. Defeat Slime B1 and Slime B2.
4. Spawn and collect Key 1.
5. Return to Room C.
6. Use Key 1 to permanently unlock Room D.
7. Enter Room D.
8. Defeat Slime D1, Guard D1, and Guard D2.
9. Spawn and collect Key 2.
10. Return to Room C.
11. Use Key 2 to permanently unlock Room E.
12. Enter Room E.
13. Contact the final treasure chest.
14. Open the chest and obtain the treasure.
15. Enter Victory if HP remains above 0.

---



# 7. Goal, Feedback & Termination
The player's final goal is to obtain the treasure in Room E.

Required progression:
- Explore Room B.
- Defeat both Slimes.
- Obtain Key 1.
- Unlock Room D.
- Defeat the Room D Slime and both Guards.
- Obtain Key 2.
- Unlock Room E.
- Contact and open the final treasure chest.

Enemy hit feedback:
- Enemy HP decreases immediately.
- HP bar shortens.
- Brief approximately 0.1-second hit highlight appears.

Enemy death:
- Enemy is removed immediately.
- Dead state persists across room transitions.

Player damage:
- HP decreases by 1.
- Top HP display updates.
- 0.8-second damage invulnerability begins.
- Player flashes.

Key feedback:
- Spawned Keys use gold graphics.
- Collection removes the Key immediately.
- Keys updates immediately.
- Keys represents only currently held, unconsumed Keys.

Locked-door feedback:
- Missing required Key → displays `Locked`.
- Successful unlock → changes the door to an open entrance.
- Once unlocked, the door remains permanently passable.

Victory candidate:
- The player contacts the final closed treasure chest.
- The chest becomes `opened`.
- The treasure is obtained.

Game Over:
`HP = 0`

If HP reaches 0 in the same frame in which the final treasure is obtained, Game Over takes priority.

After `victory`:
- Freeze the player.
- Freeze current-room objects.
- Keep the chest opened.
- Stop attacks.
- Stop damage.
- Stop room transitions.
- Stop Key collection.
- Set Status to `Victory`.
- Display `Treasure Found!`.
- Display `Press R to Restart`.
- Accept only R.

After `game_over`:
- End the current attack.
- Freeze the player and current-room enemies.
- Stop damage.
- Stop room transitions.
- Stop Key spawning and collection.
- Stop door unlocking.
- Prevent chest opening from creating Victory.
- Preserve the world state at the moment of death.
- Set Status to `Game Over`.
- Display `Game Over`.
- Display `Press R to Restart`.
- Accept only R.

R is effective only in `victory` and `game_over`.

Restart fully resets the five-room world:
- Current room → Room A
- Player center → `(180, 350)`
- HP → 5
- Keys → 0
- Facing → `right`
- Attack state and cooldown cleared
- Damage invulnerability cleared
- Locked feedback cleared
- Room B Slimes restored at initial positions with 2 HP
- Room D Slime restored at `(600, 450)` with 2 HP
- Guard D1 restored to `(320, 220)`, 3 HP, `patrol`, moving down
- Guard D2 restored to `(320, 480)`, 3 HP, `patrol`, moving up
- Key 1 returned to unspawned, uncollected, unheld
- Key 2 returned to unspawned, uncollected, unheld
- `door_d_unlocked = false`
- `door_e_unlocked = false`
- C-D and C-E doors restored to locked
- Treasure chest restored to `closed`
- Room display → A
- Status → `Playing`
- Game state → `playing`



\end{lstlisting}

\subsection{Full-150 Task Catalog}
\label{app:variant_catalog}


Table~\ref{tab:full150_variants} summarizes the rule changes across all 135 variants. Within each game family, V1 denotes the canonical specification, and V2--V10 denote variants. Only changes relative to V1 are listed.

\section{Generation Protocol Details}
\label{app:generation_protocol}

\subsection{Stage-wise Generation}

Table~\ref{tab:generation_stage_inputs} specifies the task-specific inputs and prior context provided to each generation stage. Checkpoint-derived requirements are public behavioral requirements compiled from checkpoint semantics; evaluator-only checkpoint metadata, identifiers, detection methods, scores, weights, and scoring rubrics are not exposed to the generation model.

\begin{table}[htbp]
\centering
\caption{Stage-specific task information and prior context used in Spec2Game.}
\label{tab:generation_stage_inputs}
\small
\setlength{\tabcolsep}{5pt}
\renewcommand{\arraystretch}{1.12}

\begin{tabular}{
    p{0.15\linewidth}
    p{0.48\linewidth}
    p{0.27\linewidth}
}
\toprule
\textbf{Stage}
&
\textbf{Task-Specific Inputs}
&
\textbf{Prior Context}
\\
\midrule

\texttt{state.py}
&
Game Overview; Game Objects \& States; Game Flow \& Progression.
D2.1 checkpoint-derived requirements related to state modeling.
&
None
\\
\addlinespace[5pt]

\texttt{game.py}
&
Game Overview; Game Objects \& States; Player Interaction;
Core Rules \& Mechanics; Game Flow \& Progression;
Goal, Feedback \& Termination.
D2.2--D2.5 checkpoint-derived requirements related to interaction,
rules, dynamic evolution, and goal/termination behavior.
&
Complete locked \texttt{state.py}
\\
\addlinespace[5pt]

\texttt{rendering.py}
&
Game Overview; Game Scene \& Interface;
Goal, Feedback \& Termination.
Fixed task-independent visual-quality requirements.
No checkpoint-derived functional requirements.
&
Complete locked \texttt{state.py}
\\
\addlinespace[5pt]

\texttt{main.py}
&
Player Interaction.
No checkpoint-derived requirements.
&
Public API summaries of \texttt{state.py},
\texttt{game.py}, and \texttt{rendering.py}
\\

\bottomrule
\end{tabular}
\end{table}

The public API summaries provided to \texttt{main.py} are constructed from the preceding generated modules and contain only the public API information needed for final integration.

\subsection{Output Processing and Retry Policy}

At each stage, the model is instructed to return a single Python code block containing only the complete code for the target file. The response undergoes a conservative output-completeness check to determine whether a technical retry is needed. This check detects provider-reported truncation or length limits, empty responses or extracted code, unclosed code fences, and EOF-like syntax truncation. Formatting noncompliance alone does not trigger regeneration when nonempty source code can still be extracted.

After nonempty candidate source code is extracted and written to the candidate project, static AST, interface, and dependency-direction diagnostics are performed. These diagnostics are non-corrective: they record properties of the generated candidate but do not trigger automatic formatting, import modification, code completion, or semantic repair. Interface diagnostics check the presence of required program elements rather than full API correctness or runtime behavior.

Candidate programs are not executed during generation, and runtime errors, executability results, checkpoint scores, or other evaluator feedback are not returned to the model. Technical retries are reserved for API/network failures or conservatively detected incomplete outputs; ordinary syntax, interface, dependency-direction, functional, or specification errors in otherwise complete outputs do not trigger retries.

Within each generation cycle, a stage is allowed at most two technical attempts: one initial request and at most one retry. The retry uses the same prompt as the initial request and includes no additional error feedback. Invoking \texttt{--resume-generation} preserves all previously locked stages and starts a new two-attempt generation cycle for the failed stage; this explicit cycle is not counted as an automatic retry within the preceding cycle. After the failed stage succeeds, generation continues with any subsequent stages that have not yet been completed.

\clearpage
\onecolumn

\begingroup
\small
\setlength{\tabcolsep}{5pt}
\renewcommand{\arraystretch}{1.08}

\begin{longtable}{
    @{}
    >{\raggedright\arraybackslash}p{0.09\textwidth}
    >{\raggedright\arraybackslash}p{0.87\textwidth}
    @{}
}
\caption{Rule changes across the 135 Full-150 variants.}
\label{tab:full150_variants}\\

\toprule
\textbf{Version} & \textbf{Changes relative to V1} \\
\midrule
\endfirsthead

\multicolumn{2}{@{}l}{\textit{Table~\thetable\ (continued)}}\\
\toprule
\textbf{Version} & \textbf{Changes relative to V1} \\
\midrule
\endhead

\bottomrule
\endfoot

\multicolumn{2}{@{}l}{\textbf{1. Snake}}\\
V2 & Increase initial snake length from 3 to 5 and change its starting position.\\
V3 & Increase movement speed from 8 to 10 steps/s.\\
V4 & Add fixed obstacles; obstacle collisions end the game, and food cannot spawn on obstacles.\\
V5 & Maintain two food items simultaneously; replace each item individually when consumed.\\
V6 & Increase snake length by 2 per food item while retaining a score increment of 1.\\
V7 & Enable wraparound at all four boundaries; self-collision remains terminal.\\
V8 & Introduce a more complex fixed obstacle layout consisting of short walls.\\
V9 & Combine fixed obstacles with two simultaneous food items.\\
V10 & Combine fixed obstacles, two simultaneous food items, and a movement speed of 10 steps/s.\\

\addlinespace[5pt]
\multicolumn{2}{@{}l}{\textbf{2. Dodge Blocks}}\\
V2 & Increase horizontal player speed from 300 to 360 px/s.\\
V3 & Increase obstacle falling speed from 200 to 260 px/s.\\
V4 & Reduce the obstacle spawn interval from 1.0 to 0.75 s.\\
V5 & Increase dash distance from 80 to 120 px; retain the 2 s cooldown.\\
V6 & Reduce dash cooldown from 2 to 1 s; retain the 80 px dash distance.\\
V7 & Spawn two obstacles per event at independently randomized, initially non-overlapping horizontal positions.\\
V8 & Randomize obstacle size between 40 $\times$ 40 and 60 $\times$ 40 px.\\
V9 & Introduce visually distinct normal and fast obstacles, falling at 200 and 300 px/s, respectively.\\
V10 & Combine two obstacles per spawn event with a 120 px dash distance.\\

\addlinespace[5pt]
\multicolumn{2}{@{}l}{\textbf{3. Whack-a-Mole}}\\
V2 & Extend game duration from 30 to 45 s.\\
V3 & Set normal-mole visibility to 0.8 s and golden-mole visibility to 0.5 s.\\
V4 & Reduce the waiting interval between mole appearances from 0.3 to 0.15 s.\\
V5 & Set spawn probabilities to 65\% for normal moles and 35\% for golden moles.\\
V6 & Expand the hole grid from 3 $\times$ 3 to 4 $\times$ 4; retain the one-mole limit.\\
V7 & Allow up to two moles in distinct holes, with independent types, visibility durations, and hit detection.\\
V8 & Deduct 1 point for clicking an empty location; end the game if the score falls below zero.\\
V9 & Prevent a new mole from appearing in the hole occupied by the preceding mole.\\
V10 & Combine a two-mole limit with a 0.15 s waiting interval between appearances.\\

\addlinespace[5pt]
\multicolumn{2}{@{}l}{\textbf{4. Asteroids}}\\
V2 & Increase ship movement speed from 220 to 280 px/s.\\
V3 & Increase ship rotation speed from 180 to 240 degrees/s.\\
V4 & Reduce the firing interval from 0.25 to 0.15 s.\\
V5 & Limit active bullets to three; firing at the limit creates no bullet and does not reset the cooldown.\\
V6 & Increase asteroid speed from approximately 80 to 120 px/s; adjust initial velocity vectors accordingly.\\
V7 & Increase the initial asteroid count from 3 to 4 and the maximum concurrent count from 4 to 5.\\
V8 & Increase asteroid radius from 25 to 35 px; adjust collision thresholds and spawn positions accordingly.\\
V9 & Introduce equally probable normal and large asteroids with radii of 25 and 35 px, distinct appearances, and size-appropriate collision thresholds.\\
V10 & Combine a 0.15 s firing interval with an asteroid speed of approximately 120 px/s.\\

\addlinespace[5pt]
\multicolumn{2}{@{}l}{\textbf{5. Maze Runner}}\\
V2 & Extend the time limit from 30 to 45 s.\\
V3 & Reduce the held-key movement repeat interval from 0.15 to 0.10 s.\\
V4 & Relocate the starting position to another fixed, reachable pathway cell.\\
V5 & Relocate the exit to another fixed, reachable pathway cell.\\
V6 & Replace the maze with a different fixed 15 $\times$ 15 layout.\\
V7 & Introduce a 50-step limit; exceeding it before reaching the exit causes failure.\\
V8 & Deduct 1 s for each invalid movement attempt without moving the player or increasing the step count.\\
V9 & Add a second exit; reaching either exit before time expires wins the game.\\
V10 & Combine the alternative fixed 15 $\times$ 15 maze with a 45 s time limit.\\

\addlinespace[5pt]
\multicolumn{2}{@{}l}{\textbf{6. Space Invaders}}\\
V2 & Increase horizontal player speed from 300 to 380 px/s.\\
V3 & Reduce the player firing interval from 0.20 to 0.15 s.\\
V4 & Increase initial lives from 3 to 5.\\
V5 & Increase enemy-bullet speed from 250 to 320 px/s.\\
V6 & Extend post-hit invulnerability from 1.0 to 1.5 s.\\
V7 & Increase enemy counts across the three waves while retaining their fixed formations.\\
V8 & Make two distinct surviving enemies fire simultaneously at each firing event; if only one survives, it fires alone.\\
V9 & Introduce armored enemies requiring two player-bullet hits; normal enemies still require one.\\
V10 & Increase enemy-bullet speed to 320 px/s and reduce the firing intervals of Waves 1--3 to 1.3, 1.1, and 0.9 s, respectively.\\

\addlinespace[5pt]
\multicolumn{2}{@{}l}{\textbf{7. Simple Platformer}}\\
V2 & Increase horizontal player speed from 220 to 280 px/s.\\
V3 & Increase jump strength by changing initial vertical velocity from $-420$ to $-500$ px/s.\\
V4 & Increase gravity from 900 to 1,100 px/s$^2$; retain the 600 px/s maximum falling speed.\\
V5 & Increase Moving Platform 1 speed from 80 to 120 px/s.\\
V6 & Extend Moving Platform 1's horizontal range from $x=260$--430 to $x=220$--470.\\
V7 & Add two 40 $\times$ 20 px spikes, increasing the total spike count from 3 to 5.\\
V8 & Add a second moving platform with horizontal range $x=460$--620 and speed 70 px/s; both platforms can carry the player.\\
V9 & Limit each run to eight successful jumps; attempted jumps after the limit have no effect.\\
V10 & Combine an initial jump velocity of $-500$ px/s with Moving Platform 1 speed of 120 px/s.\\

\addlinespace[5pt]
\multicolumn{2}{@{}l}{\textbf{8. Tetris}}\\
V2 & Reduce the automatic falling interval from 0.60 to 0.45 s.\\
V3 & Reduce the held-key repeat interval for left, right, and down inputs from 0.12 to 0.08 s.\\
V4 & Shift every tetromino's spawn column from 3 to 2.\\
V5 & Make each successful soft-drop action move the piece down by up to two cells; fall back to one cell if two-cell movement is blocked.\\
V6 & Hide the next-piece preview while retaining internal next-piece generation and promotion rules.\\
V7 & Change rewards for clearing 1, 2, 3, and 4 lines to 100, 250, 450, and 700 points, respectively.\\
V8 & Display a two-piece preview queue; advance the queue and generate a new second preview after each piece locks.\\
V9 & Limit each active piece to four successful rotations; reset the count for each new piece.\\
V10 & Combine a 0.45 s automatic falling interval with a 0.08 s held-key repeat interval.\\

\addlinespace[5pt]
\multicolumn{2}{@{}l}{\textbf{9. Match-3}}\\
V2 & Reduce the board size from 8 $\times$ 8 to 7 $\times$ 7, adjusting its layout within the unchanged window.\\
V3 & Increase gem types from five to six by adding a visually distinct orange gem.\\
V4 & Increase the reward per distinct cleared gem from 10 to 15 points, including during automatic chains.\\
V5 & Count diagonal runs of at least three identical gems as matches for swap validation and automatic chains.\\
V6 & Increase the horizontal and vertical match threshold from three to four consecutive identical gems.\\
V7 & Change the purple gem's display color to orange while retaining its logical type and shape.\\
V8 & Change selected gem shapes without altering their colors or matching behavior.\\
V9 & Reject swaps between adjacent gems of the same type without changing the board, score, or remaining moves.\\
V10 & Combine six gem types with a horizontal and vertical match threshold of four.\\

\addlinespace[5pt]
\multicolumn{2}{@{}l}{\textbf{10. Simple Shop Management}}\\
V2 & Add Orange as a fourth product: initial stock 3, maximum stock 9, restock amount 3, restock cost 8, and selling price 5; distribute customer requests equally across four products.\\
V3 & Replace Milk with Orange while retaining Milk's original stock, restock, cost, and selling-price parameters.\\
V4 & Increase Apple, Bread, and Milk selling prices from 4/6/8 to 5/8/10.\\
V5 & Reduce Apple, Bread, and Milk restock costs from 6/9/12 to 5/8/11.\\
V6 & Change initial stock to 5/3/1 for Apple/Bread/Milk.\\
V7 & Increase maximum stock for all three products from 9 to 12; retain a restock amount of 3.\\
V8 & Set restock amounts to 2/3/4 for Apple/Bread/Milk, with product-specific restock eligibility.\\
V9 & Set initial stock, maximum stock, and restock amount to 2/8/2 for Apple, 4/10/3 for Bread, and 6/12/4 for Milk.\\
V10 & Combine initial stock of 5/3/1 with selling prices of 5/8/10 for Apple/Bread/Milk.\\

\addlinespace[5pt]
\multicolumn{2}{@{}l}{\textbf{11. Carrot Defense-like}}\\
V2 & Increase initial gold from 120 to 150; increase Basic Tower cost from 50 to 60 and Freeze Tower cost from 70 to 80.\\
V3 & Increase Fast Enemy speed from 90 to 110 px/s and Tank HP from 100 to 130.\\
V4 & Extend the game from four to five waves, adding a final wave dominated by Fast Enemies and Tanks.\\
V5 & Move the sole enemy entrance to the upper-left edge and redefine its fixed path to the unchanged carrot.\\
V6 & Move the carrot to the lower-right edge while retaining the original entrance; redefine the enemy path accordingly.\\
V7 & Introduce two enemy entrances with separate paths that merge before reaching the same carrot.\\
V8 & Split the route from one entrance into two fixed paths that rejoin near the carrot; assign enemies to predefined routes.\\
V9 & Change Freeze Tower's slow effect from 40\% for 2 s to 50\% for 1.5 s; increase Basic Tower range from 120 to 140 px.\\
V10 & Increase the perfect-wave reward from 30 to 50 gold and Tank damage to the carrot from 2 to 3.\\

\addlinespace[5pt]
\multicolumn{2}{@{}l}{\textbf{12. Farming-lite}}\\
V2 & Increase initial gold from 80 to 100; increase Carrot and Pumpkin seed costs from 8/18 to 10/22.\\
V3 & Increase Carrot and Pumpkin harvest rewards from 18/50 to 22/60.\\
V4 & Reduce day length from 30 to 25 s while retaining the eight-day game duration.\\
V5 & Extend the game from 8 to 10 days, with final settlement on Day 10.\\
V6 & Increase required growth events from 2 to 3 for Carrots and from 4 to 5 for Pumpkins.\\
V7 & Make unwatered, immature crops lose one growth stage at day end, down to a minimum of Stage 0.\\
V8 & Charge 1 gold per plot per day for watering; repeated watering has no additional effect or cost.\\
V9 & Replace Pumpkin with Corn: seed cost 12 gold, three required growth events, and harvest reward 32 gold.\\
V10 & Remove automatic harvesting at final settlement; only crops harvested manually before the countdown ends contribute to gold and harvest count.\\

\addlinespace[5pt]
\multicolumn{2}{@{}l}{\textbf{13. Pipe Connection Puzzle}}\\
V2 & Reduce the maximum number of valid rotations from 30 to 24.\\
V3 & Reverse pipe rotation from 90 degrees clockwise to 90 degrees counterclockwise per click.\\
V4 & Allow editing and resubmission after a failed flow check; preserve existing rotations and move count, and recompute flow feedback on every submission.\\
V5 & Relocate one sink to another fixed boundary cell and reconstruct the pipe layout to preserve a valid connection to both sinks.\\
V6 & Relocate the leak cell and adjust the fixed map while preserving a solvable pipe network.\\
V7 & Rearrange fixed block cells and affected pipes while preserving a solvable connection to both sinks.\\
V8 & Relocate the source to another boundary cell, redefine its fixed outlet, and reconstruct the pipe layout.\\
V9 & Replace selected pipe types and reconstruct the layout to preserve a valid connection to both sinks.\\
V10 & Allow victory when both sinks are reached with at most one leak, instead of requiring zero leaks.\\

\addlinespace[5pt]
\multicolumn{2}{@{}l}{\textbf{14. Top-down Arena Shooter}}\\
V2 & Increase player HP from 5 to 7.\\
V3 & Increase magazine capacity from 6 to 8 rounds.\\
V4 & Reduce firing cooldown from 0.25 to 0.18 s.\\
V5 & Change enemy composition from three Chasers and three Shooters to four Chasers and two Shooters.\\
V6 & Increase Chaser speed from 120 to 150 px/s.\\
V7 & Change Shooter distance thresholds from 300/180 to 340/220 px.\\
V8 & Replace the original obstacle arrangement with a new three-wall layout.\\
V9 & Reduce post-hit invulnerability from 0.8 to 0.5 s.\\
V10 & Prioritize victory over game over when the player dies and all enemies are defeated in the same update.\\

\addlinespace[5pt]
\multicolumn{2}{@{}l}{\textbf{15. Mini Zelda-like Room Adventure}}\\
V2 & Increase player HP from 5 to 7.\\
V3 & Reduce attack cooldown from 0.45 to 0.30 s.\\
V4 & Increase sword attack length from 42 to 54 px while retaining its 34 px width.\\
V5 & Change Room B's enemy composition from two Slimes to one Slime and one Guard.\\
V6 & Increase the Guard chase-trigger distance from 180 to 240 px.\\
V7 & Make Key 1 reusable: unlocking Room D no longer consumes it.\\
V8 & Replace Room D's obstacle layout while preserving valid spawns, door access, and traversable paths.\\
V9 & Spawn Key 2 after both Guards in Room D are defeated, regardless of whether Slime D1 remains alive.\\
V10 & Require the player to strike the treasure chest with the sword to open it and achieve victory.\\

\end{longtable}
\endgroup

\clearpage
\twocolumn

\begin{table}[htbp]
\centering
\small
\caption{Pilot tasks and their inclusion in the final benchmark.}
\label{tab:pilot_tasks}
\begin{tabular}{lll}
\toprule
Game & Complexity & Included in Final Benchmark \\
\midrule
Snake & Easy & Yes \\
Flappy Bird & Easy & No \\
2048 & Medium & No \\
Space Invaders & Medium & Yes \\
Carrot Defense-like & Hard & Yes \\
Lode Runner-like & Hard & No \\
\bottomrule
\end{tabular}
\end{table}

\section{Pilot Feasibility Study Details}
\label{app:pilot}

\paragraph{Pilot Tasks.}
The pilot included two tasks at each implementation-complexity level. Table~\ref{tab:pilot_tasks} lists the 6 tasks and indicates which game families were retained in the final benchmark.

\paragraph{Generation Setup.}
The pilot used Alibaba Qwen3.7 Plus (\texttt{qwen3.7-plus}), with one generation per task, a temperature of 0.2, thinking enabled, and no specified generation seed. Candidate projects were generated using the staged generation procedure described in Section~\ref{sec:generation_protocol}.

\paragraph{Evaluation Scope and Protocol Boundary.}
The pilot used the feasibility-oriented evaluation configuration available at the time, which was not identical to the finalized evaluation protocol used for the formal benchmark. All formal results were obtained from subsequent independent evaluation runs using the finalized protocol; pilot-generated candidates were excluded.

\section{Detailed Evaluation Rubrics and Procedures}
\label{app:evaluation}

This section provides the operational definitions, evidence construction procedures, scoring rules, and aggregation procedures for the four evaluation dimensions introduced in Section~\ref{sec:evaluation}. The protocol configuration, score transformations, and evaluator-side selection rules are fixed before formal evaluation and applied consistently across candidate projects. Candidate-specific runtime evidence and the D4.3 question pack are then constructed under those frozen rules.

\subsection{D1: Executability}
\label{app:d1}

D1 measures whether a generated project can satisfy the basic execution requirements of a complete Pygame program. We define six deterministic checks with a fixed scoring order:
\textit{Syntax Validity} $\rightarrow$
\textit{Dependency Readiness} $\rightarrow$
\textit{Window Creation} $\rightarrow$
\textit{Event Handling} $\rightarrow$
\textit{Short-Horizon Runtime Stability} $\rightarrow$
\textit{Controlled Termination}.
Each check that is actually executed produces a pass/fail outcome together with the corresponding static or runtime evidence. Some runtime probes have prerequisite conditions, so the scoring order does not exactly match the actual execution dependencies. The operational pass conditions are summarized in Table~\ref{tab:d1-rubric}.

\begin{table*}[t]
\centering
\small
\caption{Operational definitions for D1 Executability.}
\label{tab:d1-rubric}
\begin{tabular}{p{0.18\linewidth} p{0.54\linewidth} p{0.20\linewidth}}
\toprule
\textbf{Check} & \textbf{Pass condition} & \textbf{Recorded evidence} \\
\midrule

Syntax Validity &
All four generated files, \texttt{state.py}, \texttt{game.py}, \texttt{rendering.py}, and
\texttt{main.py}, are present and can be successfully parsed by \texttt{ast.parse}. &
Project-file validation results and per-file syntax errors. \\

Dependency Readiness &
Syntax Validity has passed; the source code imports Pygame; the evaluation environment can successfully execute
\texttt{import pygame} in an isolated subprocess; and the generated files pass the frozen dependency-direction validation. &
Pygame availability, import results, dependency-direction validation results, and related project-validation information. \\

Window Creation &
Dependency Readiness has passed; during the runtime stability probe, execution of the entry module triggers
\texttt{pygame.display.set\_mode(...)}, and the returned object is a
\texttt{pygame.Surface}. &
Probe return code, stdout/stderr, and window-creation and surface-return status. \\

Event Handling &
Dependency Readiness has passed; either
\texttt{pygame.event.get()} or \texttt{pygame.event.poll()}
is observed at runtime, or one of these event-fetching calls is detected in the AST. &
Runtime event-fetch evidence and AST inspection results. \\

Short-Horizon Runtime Stability &
The entry module does not terminate abnormally during the 5-second runtime stability probe and remains running until the evaluator reaches the predefined probe timeout. &
Return code, stdout/stderr, timeout status, and runtime status. \\

Controlled Termination &
Short-Horizon Runtime Stability has passed; the termination probe begins repeatedly injecting
\texttt{pygame.QUIT} events 1 second after program launch, and at least one QUIT event is successfully posted. The candidate program must then terminate on its own with return code 0 within the probe's 5-second timeout. &
Evidence of successful QUIT posting, return code, stdout/stderr, and timeout status. \\

\bottomrule
\end{tabular}
\end{table*}

D1 uses \emph{prefix-gated} aggregation rather than an unconditional sum over the six checks. Let $k$ denote the number of consecutively passed checks, starting from Syntax Validity and following the fixed scoring order above. Then
\begin{equation}
    D1 = 100 \times \frac{k}{6}.
\end{equation}
Checks after the first failed item therefore cannot compensate for an earlier failure, even if they could otherwise satisfy their individual pass conditions. For example, if the first three checks pass consecutively and the fourth fails, then $k=3$ and $D1=50$.

During execution, several checks have explicit prerequisites: if Syntax Validity fails, Dependency Readiness is not performed; if Dependency Readiness fails, runtime probes are not launched; and if Short-Horizon Runtime Stability fails, the Controlled Termination probe is not executed. A project passes the all-six-check D1 executability indicator only when all six checks pass. This indicator is distinct from the continuous D1 gate factor used in the official total; the latter is defined in Section~\ref{app:overall-score}.

\subsection{D2: Specification Realization}
\label{app:d2}

D2 measures the extent to which a generated project realizes the functional requirements specified in natural language. All requirement-level checks are represented as preconstructed checkpoints and assigned to five components:
\textit{Game Element Modeling},
\textit{Interaction Modeling},
\textit{Rule and Mechanism Modeling},
\textit{Dynamic Evolution}, and
\textit{Goal and Termination Modeling}.
For Core-15, the formal evaluation contains 537 checkpoints. Each checkpoint defines its own natural-language requirement, detection method, evidence requirements, and 0/1/2 scoring rubric. Evaluation evidence includes the evaluator-compiled sampling program, the corresponding state/action traces, and source-code evidence when required by the checkpoint's detection method.

\paragraph{Checkpoint scoring.}
Each checkpoint receives a score $s_i\in\{0,1,2\}$. The judge must score the provided evidence according to the checkpoint-specific rubric rather than a global condition-counting rule. In general, a score of 0 indicates that the target functionality is absent, the evidence contradicts the requirement, the sampling process fails to expose the target behavior, or the evidence is unavailable; a score of 1 indicates partial realization, with substantial conditions still missing or unresolved; and a score of 2 indicates that the complete checkpoint requirement is directly supported by the evidence. These descriptions serve only as general guidance, while the exact decision boundaries are defined by each checkpoint's own 0/1/2 rubric. For checkpoints containing multiple conditions, the absence of a required condition may therefore result in partial credit or may directly result in a score of 0, depending on its specific rubric.

\paragraph{Component aggregation.}
For each component $d$, we first normalize the scores over all checkpoints assigned to that component:
\begin{equation}
    R_d =
    \frac{\sum_{i\in d} s_i}
         {\sum_{i\in d} m_i},
\end{equation}
where $m_i$ denotes the maximum possible score for checkpoint $i$. All formal checkpoints currently use the 0/1/2 scale, so $m_i=2$, and $R_d$ is equivalent to the mean checkpoint score within the component divided by 2.

\paragraph{Specification-deviation penalty.}
In addition to requirement satisfaction, D2 separately checks whether a candidate implements mechanisms that go beyond or conflict with the specification. This penalty applies only to implemented extra or incompatible mechanisms. It does not penalize missing required functionality, purely visual decoration, menus needed to make the game operable, or code-quality issues. Each deviation is assigned to exactly one most relevant D2 component to avoid double counting. For each component $d$, the deviation severity is denoted by $P_d\in\{0,1,2,3\}$, with the operational anchors defined in Table~\ref{tab:d2-deviation}.

\begin{table}[t]
\centering
\small
\caption{Severity anchors for specification-deviation penalties in D2.}
\label{tab:d2-deviation}
\begin{tabular}{c p{0.82\linewidth}}
\toprule
\textbf{Penalty} & \textbf{Operational anchor} \\
\midrule
0 &
No substantive extra mechanism or mechanism incompatible with the specification is present; implementation details, decorative presentation, and operational menus are not treated as deviations. \\

1 &
A single localized and minor extra behavior is present, with limited impact on the requested core system; removing it would not change the main gameplay or task identity. \\

2 &
A complete extra system or a clear mechanism-level deviation is present; it materially changes a rule, interaction, dynamic evolution process, or termination logic, but does not redefine the task as a whole. \\

3 &
Multiple extra systems are present, or a single major deviation is sufficient to change the task identity, main gameplay, or objective. \\
\bottomrule
\end{tabular}
\end{table}

The requirement-realization score for each component is then adjusted by its deviation severity:
\begin{equation}
    A_d =
    \max\left(0,\,
    R_d-\frac{P_d}{3}\right).
\end{equation}
When $P_d=3$, the adjusted component score is necessarily 0 because $R_d\in[0,1]$. The judge also records the corresponding source-code evidence, the unique component assignment, and the rationale for each deviation to make the penalty decision traceable.

The five components are equally weighted in D2. The final D2 score is mapped to the 0--100 scale:
\begin{equation}
    D2 =
    100\times\frac{1}{5}
    \sum_{d=1}^{5} A_d.
\end{equation}

\subsection{D3: Code Quality}
\label{app:d3}

D3 measures the source code quality of generated projects by combining deterministic code analysis with a semantic source-code judge. The evaluation contains five components:
\textit{Maintainability},
\textit{Reliability},
\textit{Security},
\textit{Efficiency}, and
\textit{Conformance}.
Deterministic analysis uses static-analysis tools and custom AST checks to measure code properties that can be evaluated programmatically, while the semantic judge assesses semantic code qualities that are not fully captured by fixed rules. The two parts contribute 70 and 30 points, respectively, to the final D3 score. This split is a global point budget rather than a fixed 70/30 ratio within each component. The allocation is shown in Table~\ref{tab:d3-allocation}.

\begin{table}[t]
\centering
\small
\caption{Point allocation for D3 Code Quality.}
\label{tab:d3-allocation}
\begin{tabular}{lccc}
\toprule
\textbf{Component} & \textbf{Deterministic} & \textbf{judge} & \textbf{Total} \\
\midrule
Maintainability & 20 & 10 & 30 \\
Reliability     & 15 & 10 & 25 \\
Security        & 12 &  3 & 15 \\
Efficiency      & 10 &  5 & 15 \\
Conformance     & 13 &  2 & 15 \\
\midrule
\textbf{Total}  & \textbf{70} & \textbf{30} & \textbf{100} \\
\bottomrule
\end{tabular}
\end{table}

\paragraph{Deterministic scoring.}
Deterministic analysis combines Ruff, Radon, Bandit, and custom AST checks. Each component first receives a tool raw score with a component-specific maximum, which is then linearly mapped to its deterministic point budget. Let $r_k$ and $r_k^{\max}$ denote the raw score and raw maximum for component $k$, respectively, and let $c_k^{\mathrm{tool}}$ denote its deterministic point cap. The resulting tool contribution is
\begin{equation}
    T_k =
    \frac{r_k}{r_k^{\max}}
    c_k^{\mathrm{tool}}.
\end{equation}
The component-specific deterministic rules are summarized in Table~\ref{tab:d3-deterministic}.

\begin{table*}[t]
\centering
\small
\caption{Deterministic scoring rules for D3. All deduction-based scores are floored at zero.}
\label{tab:d3-deterministic}
\begin{tabular}{p{0.14\linewidth} p{0.05\linewidth} p{0.71\linewidth}}
\toprule
\textbf{Component} & \textbf{Raw max} & \textbf{Scoring rule} \\
\midrule

Maintainability &
30 &
This component consists of four parts.
\textit{Complexity} (12 points) uses Radon complexity ranks, mapping A/B/C/D/E/F to
$1/.8/.5/.25/0/0$, and aggregates the resulting factors as
$12\times(0.1\,\mathrm{mean}+0.9\,\mathrm{min})$;
\textit{Structural lint} (8 points) starts from 8 and deducts 1.5 points for each
\texttt{C901} or
\texttt{PLR0904/0911/0912/0913/0914/0915/0916/0917}
finding;
\textit{Duplicate structure} (5 points) deducts 1 point for each additional normalized AST duplicate window consisting of six statements;
\textit{Testability} (5 points) awards 2 points for a main guard, 2 points for a successful clean import, and 1 point when there are at most three executable top-level statements. \\

Reliability &
20 &
This component starts from 20 points.
Each \texttt{F821/F822/F823/B012} finding deducts 5 points;
each \texttt{B006/B008/B023/BLE001/E722/F632/F811/TRY201/TRY203} finding deducts 2 points;
and each other matching \texttt{B*}, \texttt{BLE*}, \texttt{PLE*}, \texttt{PLW*}, or \texttt{TRY*} finding deducts 1 point.
AST analysis additionally deducts 1 point for each bare or broad exception handler; if its sole body is \texttt{pass} or \texttt{continue}, a further 2 points are deducted. The cumulative AST deduction is capped at 4 points. \\

Security &
15 &
This component starts from 15 points. For each Bandit finding, the deduction is the product of a severity weight and a confidence weight, where High/Medium/Low severity correspond to $6/3/1$ and High/Medium/Low confidence correspond to $1/.75/.5$.
If at least one forbidden call is detected, an aggregate deduction of 15 points is applied; if at least one external import other than Pygame or a Python standard-library package is detected, another aggregate deduction of 15 points is applied. \\

Efficiency &
10 &
This component starts from 10 points. Each Ruff \texttt{PERF*} finding deducts 1 point, with this deduction capped at 4 points.
Each hotspot call inside a \texttt{for}/\texttt{while} loop deducts 2 points, with this deduction capped at 6 points. Hotspots include
\texttt{open}, Pygame font creation, image/sound loading,
\texttt{pygame.Surface}, rotation, and scaling. \\

Conformance &
10 &
The score is based on a weighted finding density computed from Ruff findings in
\texttt{E*}, \texttt{W*}, \texttt{I*}, \texttt{N*}, \texttt{UP*}, \texttt{SIM*}, \texttt{F401}, and \texttt{F841}.
Findings in \texttt{E1--E3/W1--W3} receive weight $0.1$;
\texttt{E501/I*/UP*/SIM*} receive weight $0.5$;
and all other included findings receive weight $1$.
Let
$\rho=100\times\text{weighted findings}/\max(1,\text{logical LOC})$.
The raw score is then
$10\max(0,1-\rho/10)$. \\

\bottomrule
\end{tabular}
\end{table*}

\paragraph{Semantic judge.}
The semantic judge receives only the anonymized full source code and does not observe candidate identity, D1/D2/D4 outcomes, deterministic tool findings, or tool scores. It does not reassess executability, gameplay completeness, or visual quality, nor does it repeat deterministic checks. Instead, it directly outputs one score in $[0,100]$ for each of the five components, without using discrete anchors followed by a mapping to the 0--100 scale. The assessment scope of each component is shown in Table~\ref{tab:d3-judge}.

\begin{table}[t]
\centering
\small
\caption{Assessment scope of the semantic judge in D3.}
\label{tab:d3-judge}
\begin{tabular}{p{0.15\linewidth} p{0.75\linewidth}}
\toprule
\textbf{Component} & \textbf{Assessment scope} \\
\midrule
Maintainability &
Responsibility boundaries, cohesion/coupling, change locality, abstraction discipline, and state architecture. \\

Reliability &
State consistency, initialization/reset/termination, update ordering, edge cases, mutation safety, and defect risk. \\

Security &
Indirect dangerous behavior, unnecessary external access, and compliance with offline Pygame task constraints. \\

Efficiency &
Repeated work, algorithmic hotspots, collection growth, resource lifetime, caching, and per-frame work. \\

Conformance &
Semantic naming, readable organization, useful comments, magic values, and main control flow. \\
\bottomrule
\end{tabular}
\end{table}

Let $j_k$ denote the judge score for component $k$ on the 0--100 scale, and let $c_k^{\mathrm{judge}}$ denote the corresponding judge point cap. Its contribution is
\begin{equation}
    J_k =
    \frac{j_k}{100}
    c_k^{\mathrm{judge}}.
\end{equation}
When no special security rule is triggered, the tool and judge contributions across the five components are summed as
\begin{equation}
    D3_{\mathrm{raw}}
    =
    \sum_{k=1}^{5}
    \left(T_k+J_k\right).
\end{equation}
If the judge response is missing or invalid, the official D3 score remains unavailable rather than falling back to deterministic tools alone.

\paragraph{Critical security risk.}
A candidate is marked as having a critical security risk if any of the following conditions holds:
(1) AST analysis detects a forbidden call with one of the prefixes
\texttt{eval}, \texttt{exec}, \texttt{compile},
\texttt{os.system}, \texttt{os.popen},
\texttt{subprocess}, \texttt{socket},
\texttt{requests}, or \texttt{urllib.request};
(2) the project imports an external package other than Pygame or a Python standard-library package;
or
(3) Bandit reports at least one High-severity, High-confidence finding.
This rule is applied after deterministic and judge scores are fused, and caps the overall D3 score at 50:
\begin{equation}
    D3 =
    \begin{cases}
        \min(D3_{\mathrm{raw}},50), & \text{if a critical security risk is detected},\\
        D3_{\mathrm{raw}}, & \text{otherwise}.
    \end{cases}
\end{equation}

\subsection{D4: User-Facing Quality}
\label{app:d4}

D4 evaluates the user-facing visual quality of generated games during actual execution along three complementary components:
\textit{Visual Presentation},
\textit{Dynamic Feedback Fidelity}, and
\textit{Visual Actionability}.
All three are based on visual evidence collected from actual execution of the candidate project, but they differ in evaluation target and scoring procedure.

\paragraph{Visual Presentation.}
D4.1, formally termed \textit{Visual Presentation}, evaluates the overall visual presentation of the game. Based on a set of screenshots sampled from execution trajectories, the multimodal judge assigns an integer score of $0$, $1$, or $2$ to each of five equally weighted criteria:
\textit{Composition and Hierarchy},
\textit{Palette and Contrast},
\textit{Shape and Style Consistency},
\textit{Polish}, and
\textit{Dynamic Presentation}.
The five criteria share the three-level anchors in Table~\ref{tab:d4-visual-presentation}. The D4.1 score is the simple mean of the five criterion scores and lies in $[0,2]$. An additional overall score returned by the judge is used only for diagnostics and does not contribute to the official D4.1 score.

\begin{table}[t]
\centering
\small
\caption{Shared scoring anchors for the five Visual Presentation criteria.}
\label{tab:d4-visual-presentation}
\begin{tabular}{c p{0.82\linewidth}}
\toprule
\textbf{Score} & \textbf{Operational anchor} \\
\midrule
0 &
The visual presentation is poor or cannot be meaningfully evaluated, for example because the frame is blank, broken, clearly unfinished, or contains severe problems in composition, hierarchy, or style. \\

1 &
The visual content is recognizable and broadly coherent, but shows noticeable weaknesses in composition, color, shape and style consistency, polish, or dynamic presentation. \\

2 &
The visual presentation shows clear design intent, is coherent and well-finished overall, and contains no substantive defects in composition, palette, shape and style consistency, or dynamic presentation. \\
\bottomrule
\end{tabular}
\end{table}

\paragraph{Dynamic Feedback Fidelity.}
D4.2 evaluates whether visual feedback following user actions or game-state changes is consistent with the dynamic behavior required by the specification. The evaluation covers all D2 checkpoints marked as dynamic; static checkpoints are excluded from D4.2. Each dynamic checkpoint corresponds to one transition-level judgment, scored along three aspects:
\textit{Faithfulness}, whether the visual feedback correctly represents the required state change;
\textit{Completeness}, whether the required information is fully visible;
and \textit{Timeliness}, whether the feedback appears at the appropriate time relative to the corresponding action or state change.
All three aspects use the shared 0--5 scale in Table~\ref{tab:d4-feedback}.

\begin{table}[t]
\centering
\small
\caption{Scoring anchors for Dynamic Feedback Fidelity.}
\label{tab:d4-feedback}
\begin{tabular}{c p{0.82\linewidth}}
\toprule
\textbf{Score} & \textbf{Operational anchor} \\
\midrule
0 &
No evaluable supporting evidence is available, for example because the frame is blank or broken, the feedback is missing, the observed behavior contradicts the requirement, or the feedback is unintelligible. \\

1 &
Only minimal traces of the target feedback are present, while severe defects dominate. \\

2 &
Recognizable supporting evidence is present, but major omissions, ambiguity, inconsistency, or clear timing problems remain. \\

3 &
The main requirement is satisfied, but several noticeable weaknesses remain. \\

4 &
The feedback is overall complete and convincing, with only minor defects. \\

5 &
The feedback is clear, complete, timely, and well designed, with no substantive defects. \\
\bottomrule
\end{tabular}
\end{table}

For each dynamic checkpoint, the transition score is the simple mean of its Faithfulness, Completeness, and Timeliness scores. D4.2 then averages the transition scores equally across all dynamic checkpoints, producing a score in $[0,5]$.

\paragraph{Visual Actionability.}
D4.3 operationalizes actionability as whether the currently visible game state supports selecting the correct next action, given the game objective and the set of legal game-level operations. Before candidate evaluation, the round-scoped D4 contract fixes two eligible dynamic D2.2 source checkpoints, the game objective, the legal operations, the question count, and the evidence-selection policy. After the candidate's D4.2 traces have been collected, the evaluator deterministically selects one candidate-specific real frame for each source, subject to the frozen material-difference rule, and constructs an image-grounded binary action-choice question for that frame. Thus the source checkpoints and selection policy are fixed in advance, whereas the image bindings and question pack are candidate-specific. The multimodal judge receives the game objective, legal game-level operations, the question, two candidate actions, and the bound runtime screenshot. For a locally valid judge response, an answer matching the frozen answer key receives 5 points and the other valid answer receives 0 points. The two question scores are averaged to obtain D4.3, which lies in $[0,5]$. If candidate evidence is unavailable under the frozen protocol, D4.3 is recorded as zero. If question construction or answer validation fails because of judge or transport failure, D4.3 remains incomplete rather than being assigned zero. Judge confidence is recorded only for diagnostic purposes and does not contribute to the official score.

\paragraph{Runtime visual evidence construction.}
Evidence for D4.2 is collected through controlled execution of all dynamic checkpoints. Each case starts from an independent
\texttt{init(seed + case\_index)} state and executes a fixed finite action program compiled by the evaluator. The evaluator captures a \textit{before} frame prior to execution, \textit{transition} frames for actions marked with
\texttt{capture\_each=true}, and an \textit{after} frame after the action sequence completes. If fewer than two collected frames exhibit materially distinct visual states, additional no-input frames are collected using pre-frozen passive-time probe delays; this procedure does not adapt to candidate scores. For each transition, the judge receives at most three representative frames selected in temporal order: the first, a middle, and the last frame.

The D4.1 screenshot pool consists of the D4.2 frames described above. Using fixed values
\texttt{art\_sample\_size=8} and
\texttt{sample\_seed=41}, the evaluator performs fixed-seed random sampling to select at most eight screenshots, which are then ordered by source sequence and relative time before being passed to the judge. The same sampling rule is applied to all candidate projects, while the screenshot content itself is candidate-specific.

D4.3 does not execute separate gameplay trajectories, but reuses runtime scenes already collected for D4.2. The two source checkpoint IDs and the deterministic evidence-selection policy are fixed before candidate evaluation. After runtime collection, the evaluator selects candidate-specific real frames under that policy and constructs the associated question pack. The resulting A/B options, answer key, and image bindings are then frozen before visual answer scoring.

The three components are normalized by their respective maximum scores and then averaged equally. Let $d_1\in[0,2]$, $d_2\in[0,5]$, and $d_3\in[0,5]$ denote D4.1, D4.2, and D4.3, respectively. Then
\begin{equation}
    D4 =
    \frac{100}{3}
    \left(
        \frac{d_1}{2}
        +
        \frac{d_2}{5}
        +
        \frac{d_3}{5}
    \right).
\end{equation}
If an evaluator or judge infrastructure failure prevents a valid score from being obtained for any component, D4 remains incomplete rather than silently assigning the missing score as zero. When the frozen protocol records candidate-side visual evidence as unavailable, the affected D4 component receives a score of 0. This status records an evidence-availability outcome and does not by itself identify the source of the unavailable evidence.

\subsection{Overall Score Aggregation}
\label{app:overall-score}

To prevent D2--D4 from contributing at full weight when basic executability is insufficient, we introduce a continuous gate controlled by D1. We define
\begin{equation}
    g(D1)=
    \operatorname{clip}
    \left(
        \frac{D1-40}{20},
        0,
        1
    \right).
\end{equation}
Thus, $g(D1)=0$ when $D1\leq40$; the gate increases linearly when $40<D1<60$; and $g(D1)=1$ when $D1\geq60$.

The final overall score is defined as
\begin{equation}
    \mathrm{Overall}
    =
    0.20D1
    +
    g(D1)
    \left(
        0.50D2
        +
        0.15D3
        +
        0.15D4
    \right).
\end{equation}
D1 therefore plays two roles in the overall score: it contributes directly with weight $0.20$, and it also controls the joint contribution of D2--D4 through $g(D1)$. The final score lies in $[0,100]$; when $g(D1)<1$, lower executability correspondingly reduces the maximum contribution of D2--D4 to the overall score. This gate affects only the official overall score. When valid evaluation evidence is available, D2, D3, and D4 remain available and may be reported separately as diagnostic scores.

\section{Main Benchmark: Experimental Configuration and Additional Results}
\label{app:main_benchmark}

\subsection{Experimental Configuration}
\label{app:main_benchmark_config}

The generation temperature is 0.2 for every candidate model except Kimi K2.6, whose official non-reasoning configuration requires a temperature of 0.6. Reasoning is disabled for all formal candidate calls. For \texttt{state.py}, \texttt{game.py}, \texttt{rendering.py}, and \texttt{main.py}, the maximum output-token budgets are respectively 4,000, 7,000, 5,000, and 2,000 for Easy and Medium tasks. Hard tasks use larger budgets of 5,000, 14,000, 10,000, and 2,000. Each stage allows at most two API attempts separated by two seconds. Truncation checks are enabled, semantic retries are disabled, and candidate models receive no execution feedback.

\subsection{Core-15 Detailed Results}
\label{app:core15_detailed}




Table~\ref{tab:core15_detailed_results} summarizes the complete Core-15 results for all 14 candidate models. For each model, scores are first averaged over the three independent generations within each game family and then averaged equally across the 15 families. The Overall score is computed using the official aggregation rule defined in the main text.

\begin{table}[htbp]
\centering
\small
\setlength{\tabcolsep}{4.5pt}
\renewcommand{\arraystretch}{1.08}
\begin{tabular}{lrrrrr}
\toprule
\textbf{Model} & \textbf{D1} & \textbf{D2} & \textbf{D3} & \textbf{D4} & \textbf{Overall} \\
\midrule
Claude Sonnet 5      & 96.30 & 69.95 & 84.42 & 60.01 & 77.27 \\
Claude Haiku 4.5     & 97.04 & 69.74 & 81.42 & 56.10 & 74.91 \\
DeepSeek V4 Pro-0813 & 96.30 & 67.90 & 82.22 & 55.50 & 73.60 \\
Grok 4.3             & 95.56 & 66.81 & 76.45 & 55.10 & 73.37 \\
Qwen3.8 Max-0902     & 73.33 & 72.85 & 81.30 & 57.73 & 71.95 \\
GPT-5.6 Luna         & 89.63 & 64.53 & 86.39 & 54.04 & 71.35 \\
GPT-5.6 Terra        & 95.93 & 58.53 & 87.40 & 44.32 & 69.26 \\
GPT-5.6 Sol          & 99.26 & 58.43 & 77.81 & 47.87 & 67.92 \\
Qwen3.8 Flash        & 71.11 & 60.41 & 69.44 & 53.81 & 59.68 \\
DeepSeek V4.1 Flash  & 77.04 & 58.77 & 80.60 & 46.95 & 59.12 \\
Kimi K2.6            & 80.00 & 59.55 & 75.48 & 42.39 & 57.68 \\
GLM-5.2              & 65.93 & 52.89 & 78.46 & 38.43 & 57.26 \\
Qwen3.7 Plus         & 69.63 & 54.45 & 66.56 & 39.11 & 55.80 \\
Mistral Small 2603   & 74.07 & 25.63 & 71.60 & 16.87 & 39.11 \\
\bottomrule
\end{tabular}
\caption{Family-balanced D1--D4 and Overall scores of the 14 candidate models on Core-15.}
\label{tab:core15_detailed_results}
\end{table}



\subsection{Executability Diagnostics}
\label{app:d1_diagnostics}



\begin{figure*}[htbp]
    \centering
    \includegraphics[width=0.95\textwidth]{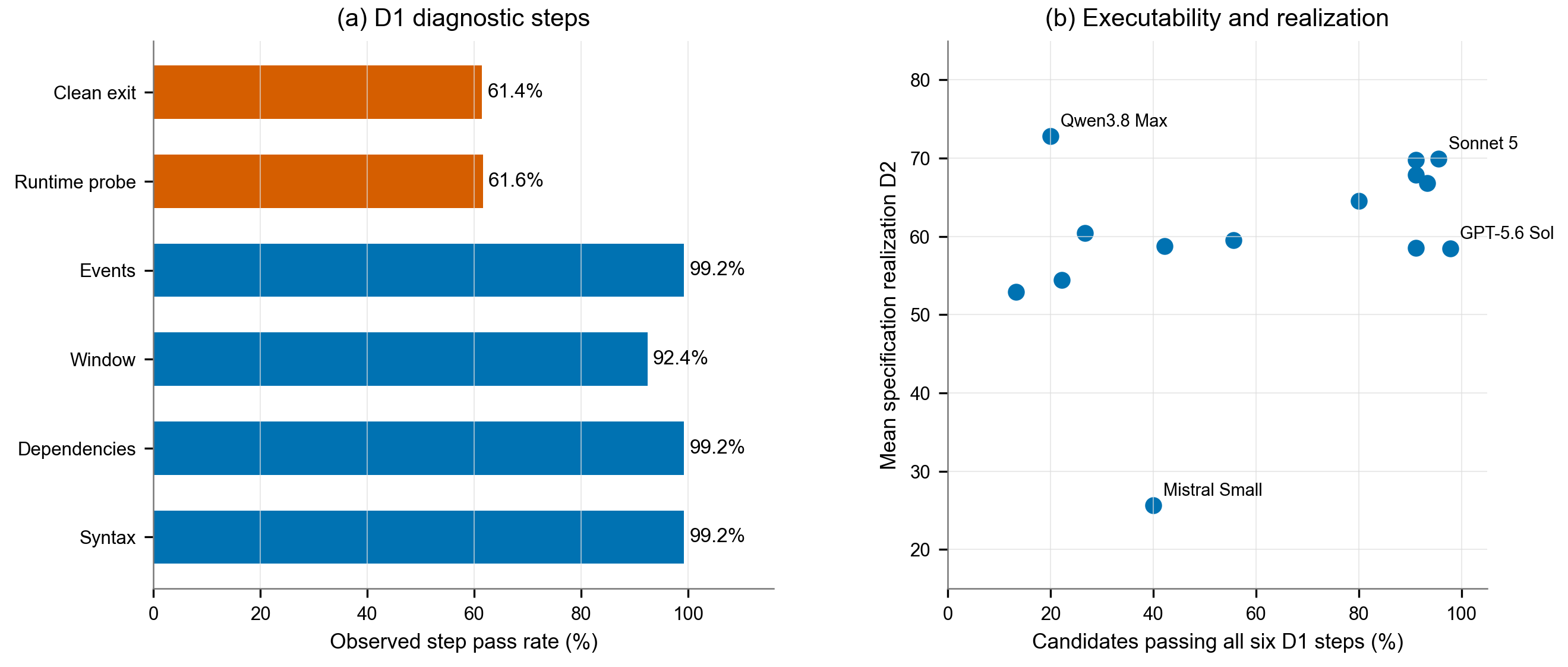}
    \caption{Executability diagnostics on Core-15. Left: observed pass rates of the six D1 checks. Right: the proportion of candidates passing all six D1 checks and the mean D2 score for each model.}
    \label{fig:d1_diagnostics}
\end{figure*}

Figure~\ref{fig:d1_diagnostics} reports the observed pass rates of the six D1 checks on Core-15 together with the model-level Executability--Specification Realization results. Among the 630 Core-15 candidate projects, Syntax Validity, Dependency Readiness, and Event Handling each achieve an observed pass rate of 99.2\%, while Window Creation reaches 92.4\%. In contrast, Short-Horizon Runtime Stability and Controlled Termination achieve 61.6\% and 61.4\%, respectively. The substantially lower observed pass rates of these two runtime-related checks indicate greater difficulty in sustaining runtime execution and terminating cleanly.

The right panel of Fig.~\ref{fig:d1_diagnostics} further shows the model-level all-six-check pass rate and mean D2 score. Models with similar all-six-check pass rates can still exhibit different D2 scores, while a lower all-six-check pass rate does not necessarily correspond to a lower D2 score.

\subsection{Component-Level Results}
\label{app:component_results}

\begin{figure*}[htbp]
    \centering
    \includegraphics[width=0.95\textwidth]{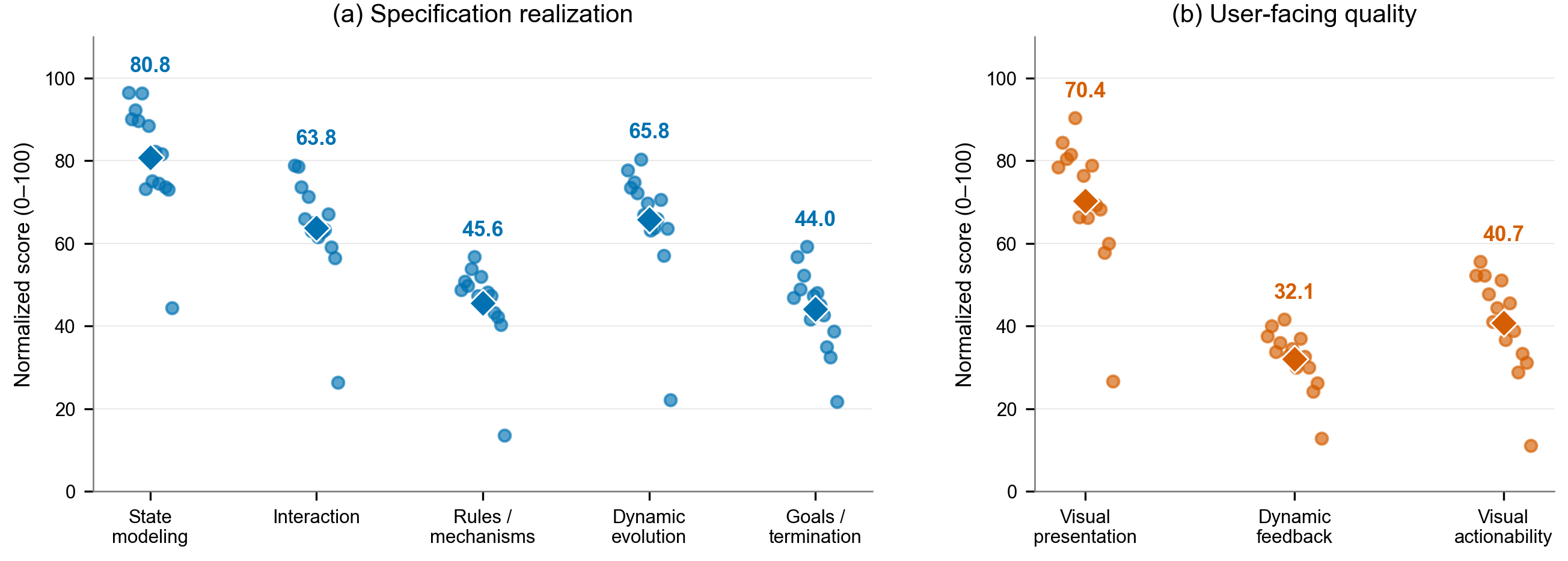}
    \caption{Component-level average scores for D2 and D4 on Core-15.}
    \label{fig:capability_components}
\end{figure*}

Figure~\ref{fig:capability_components} presents the component-level average scores for D2 and D4 on Core-15. Scores vary across the five D2 components: Game Element Modeling achieves the highest average score, at 80.8; Interaction Modeling and Dynamic Evolution score 63.8 and 65.8, respectively; and Rule and Mechanism Modeling and Goal and Termination Modeling score 45.6 and 44.0, respectively. Overall, Game Element Modeling scores higher than the other components, while Rule and Mechanism Modeling and Goal and Termination Modeling receive the lowest scores.

For D4, Visual Presentation achieves the highest average score, at 70.4, while Dynamic Feedback Fidelity and Visual Actionability score 32.1 and 40.7, respectively.

\begin{figure*}[htbp]
    \centering
    \includegraphics[width=0.95\textwidth]{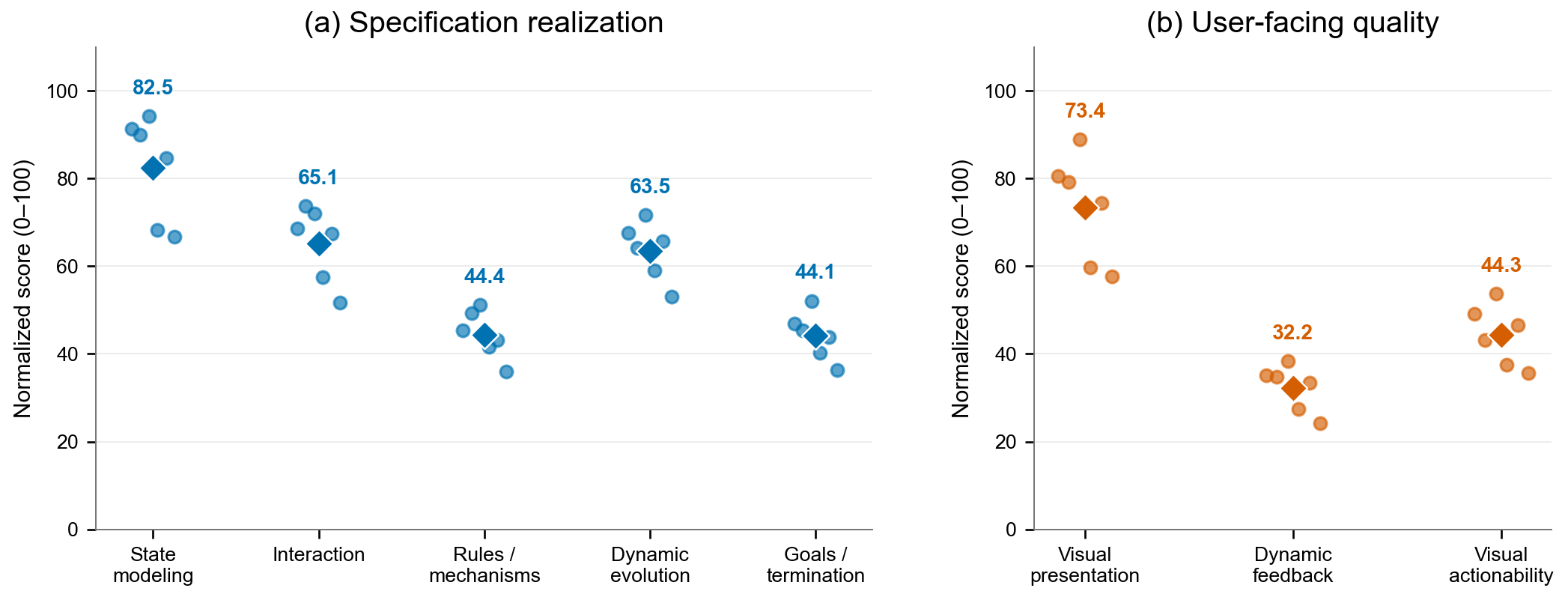}
    \caption{Component-level average scores for D2 and D4 on Full-150.}
    \label{fig:full150_capability_profile}
\end{figure*}

The Full-150 subset exhibits a component-level pattern similar to that observed on Core-15 (Fig.~\ref{fig:full150_capability_profile}). For D2, the average scores are 82.5 for Game Element Modeling, 65.1 for Interaction Modeling, 44.4 for Rule and Mechanism Modeling, 63.5 for Dynamic Evolution, and 44.1 for Goal and Termination Modeling. For D4, Visual Presentation, Dynamic Feedback Fidelity, and Visual Actionability achieve average scores of 73.4, 32.2, and 44.3, respectively.

\subsection{Performance across Implementation Complexity}
\label{app:complexity_results}





\begin{figure*}[htbp]
    \centering
    \includegraphics[width=0.9\textwidth]{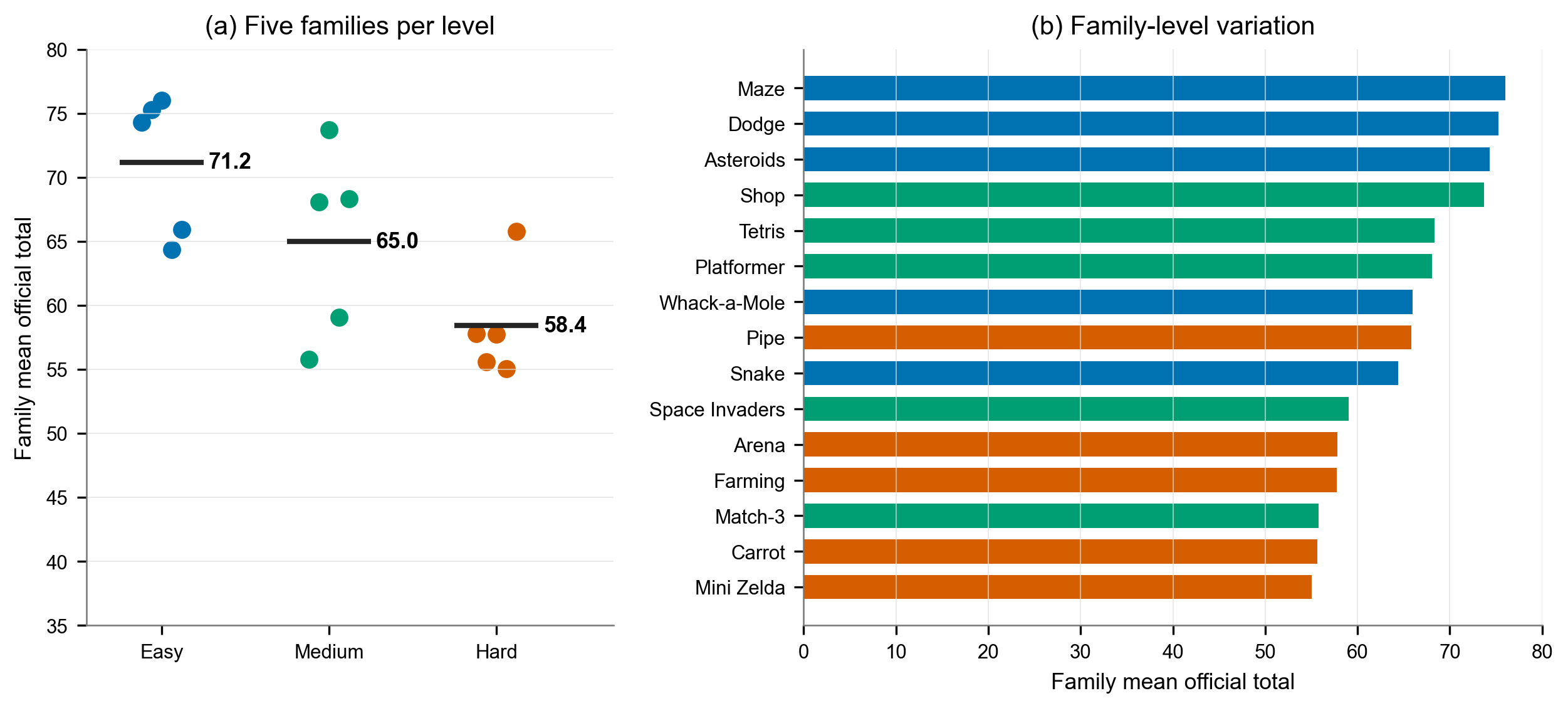}
    \caption{Performance on Core-15 by assigned implementation complexity. The left panel shows family-level Overall scores for the five game families within each complexity level together with the corresponding level mean; the right panel shows family-level variation across all 15 game families.}
    \label{fig:core15_complexity}
\end{figure*}

On Core-15, the family-balanced Overall scores for the Easy, Medium, and Hard implementation complexity levels are 71.19, 65.00, and 58.42, respectively (Fig.~\ref{fig:core15_complexity}). Although the group averages decrease from Easy to Hard, substantial family-level variation remains within each complexity level, with some overlap between adjacent levels.

\begin{figure*}[htbp]
    \centering
    \includegraphics[width=0.9\textwidth]{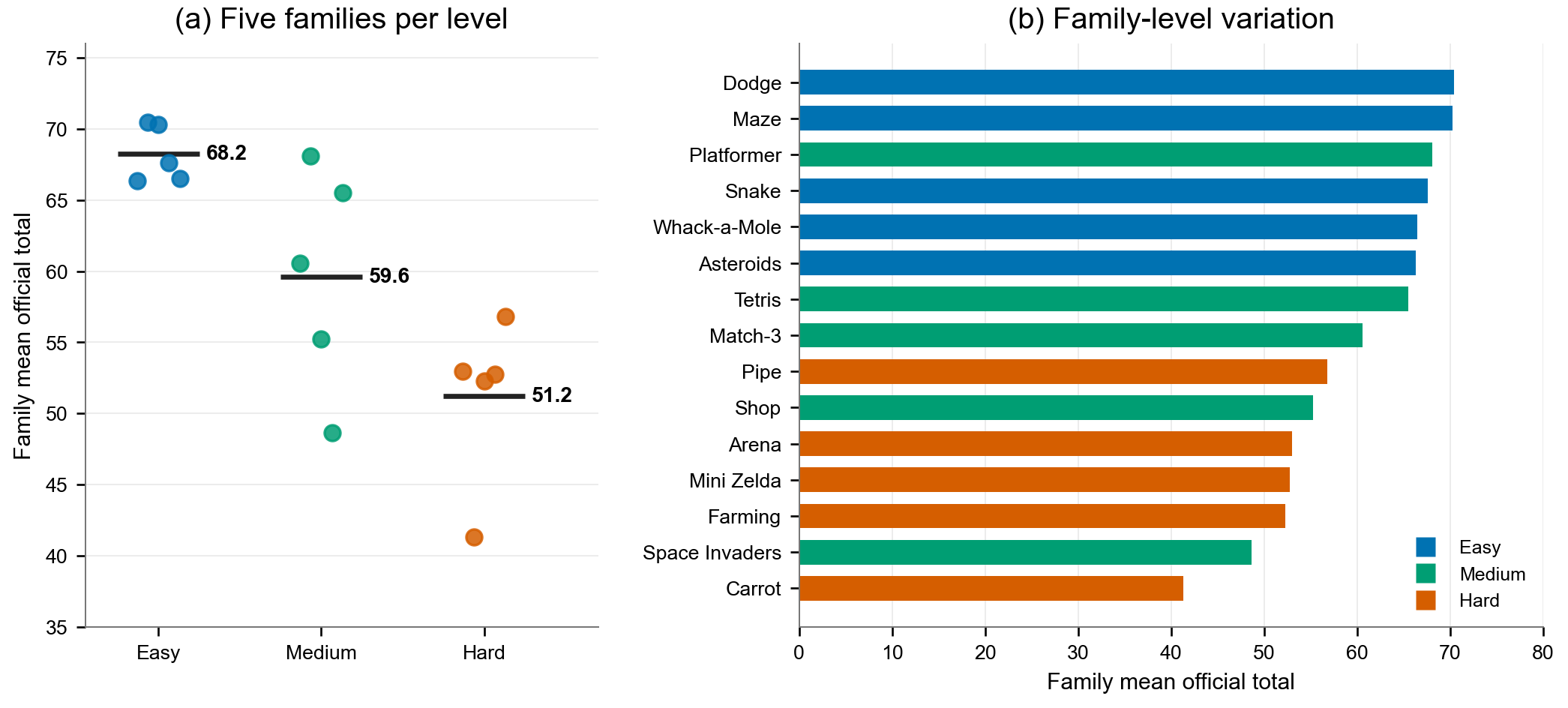}
    \caption{Performance on Full-150 by assigned implementation complexity. The left panel shows family-level Overall scores for the five game families within each complexity level together with the corresponding level mean; the right panel shows family-level variation across all 15 game families.}
    \label{fig:full150_complexity}
\end{figure*}

The same group-level ordering is observed on the six-model Full-150 subset, where the average Overall scores for Easy, Medium, and Hard are 68.24, 59.61, and 51.23, respectively (Fig.~\ref{fig:full150_complexity}). Family-level variation and cross-level score overlap are also present.

\subsection{Controlled-Variant Results}
\label{app:variant_results}

\begin{figure*}[htbp]
    \centering
    \includegraphics[width=0.9\textwidth]{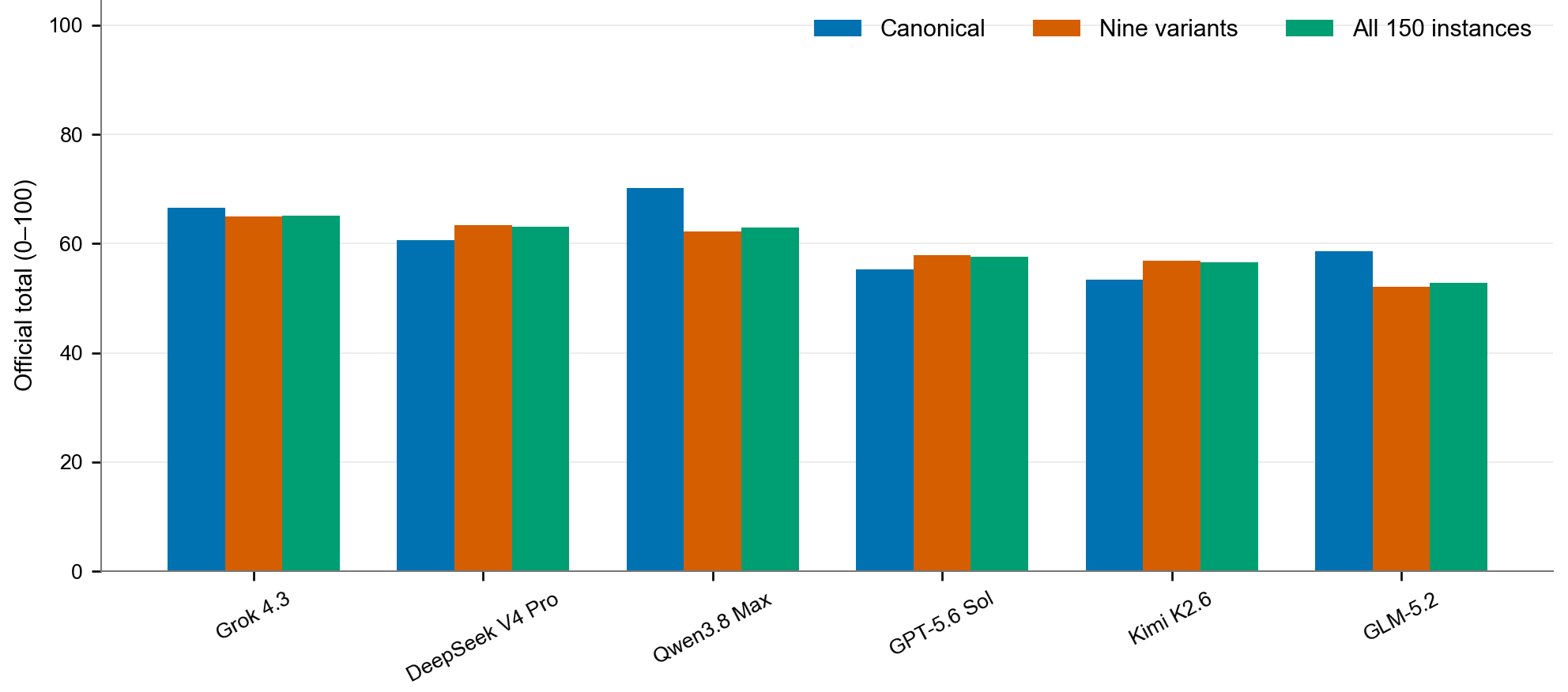}
    \caption{Overall scores of the six Full-150 models on canonical tasks, controlled variants, and all 150 task instances.}
    \label{fig:full150_variant_scores}
\end{figure*}

Figure~\ref{fig:full150_variant_scores} compares the aggregate performance of the six Full-150 models on the 15 canonical tasks, the corresponding controlled variants, and all 150 task instances. The aggregate score over the controlled variants is lower than the corresponding canonical aggregate score for Grok 4.3, Qwen3.8 Max, and GLM-5.2, whereas the opposite holds for DeepSeek V4 Pro, GPT-5.6 Sol, and Kimi K2.6. The canonical--variant score relationship therefore differs across models.

\begin{figure*}[htbp]
    \centering
    \includegraphics[width=0.9\textwidth]{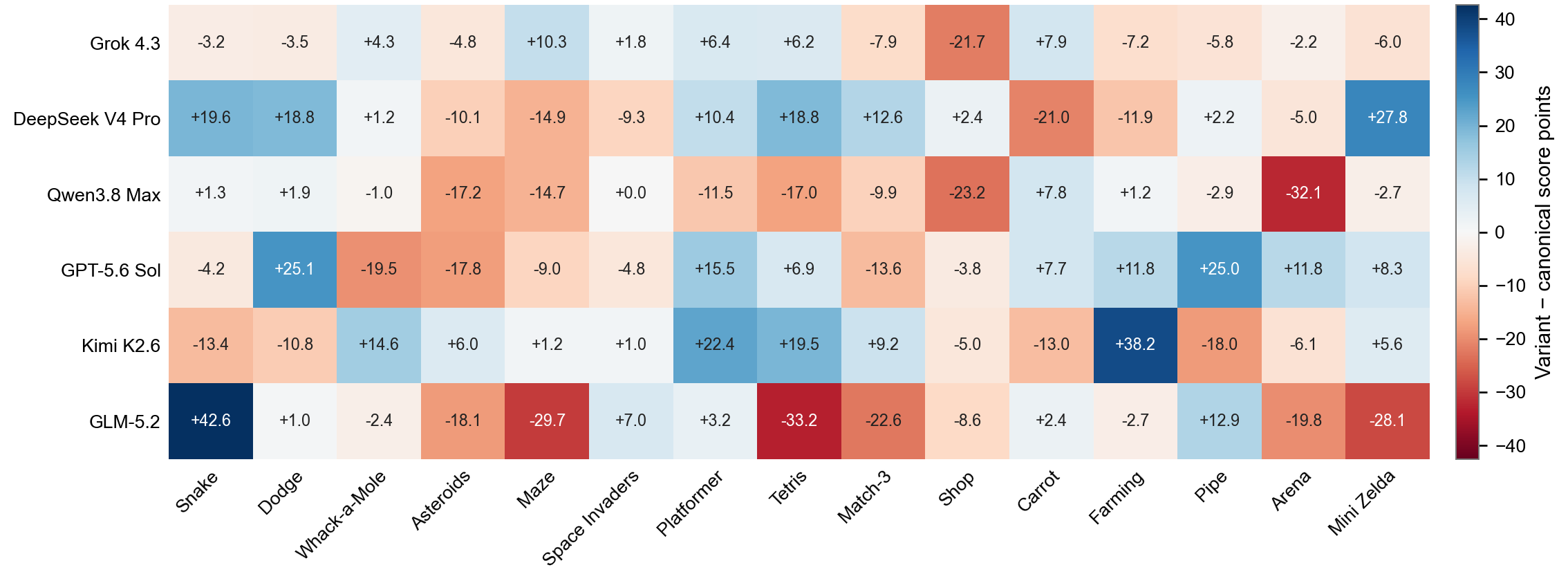}
    \caption{Family-level Overall score changes of controlled variants relative to canonical tasks in Full-150. Each cell reports the difference between the mean Overall score of the nine variants and the Overall score of the corresponding canonical task for a given model--family pair.}
    \label{fig:full150_family_variant_heatmap}
\end{figure*}

Figure~\ref{fig:full150_family_variant_heatmap} further shows variation at the model--family level. Each cell reports the difference between the mean Overall score of the nine variants and the Overall score of the corresponding canonical task for a given model--family pair. All six models exhibit both positive and negative differences, and the direction and magnitude of the changes also vary across models for the same game family, showing that variant--canonical score differences vary across both models and game families.

\subsection{Generation Variability}
\label{app:generation_variability}

\begin{figure*}[htbp]
    \centering
    \includegraphics[width=0.9\textwidth]{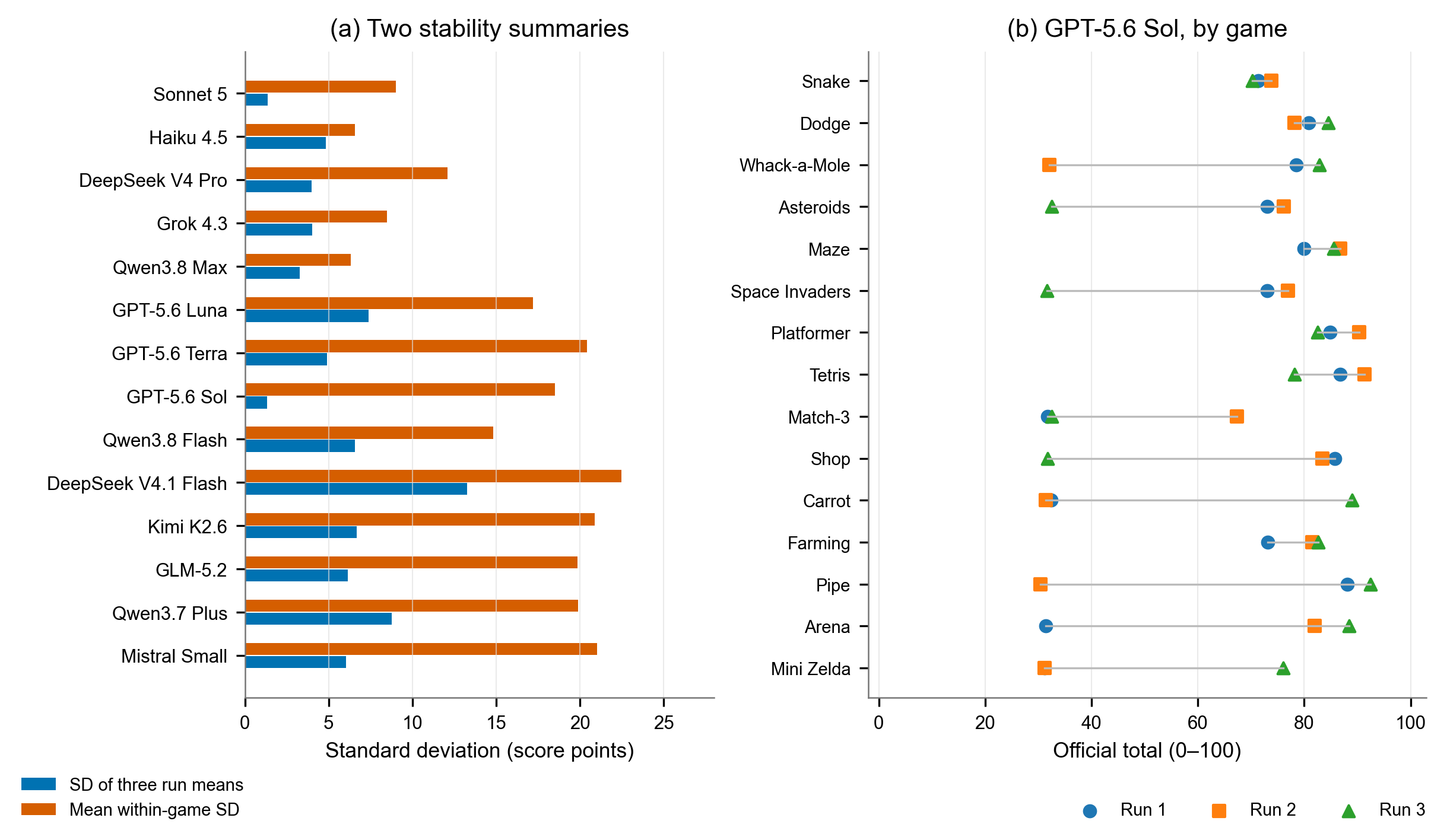}
    \caption{Overall score variability across three independent generations on Core-15. The left panel compares, for each model, the standard deviation of the three generation-index means with the mean within-task standard deviation across the 15 canonical tasks. The right panel uses GPT-5.6 Sol as an example and shows the Overall scores of its three generations for each canonical task.}
    \label{fig:generation_variability}
\end{figure*}

Figure~\ref{fig:generation_variability} compares two variability summaries on Core-15 based on three independent generations. For each model, we first average the first, second, and third generations separately across the 15 canonical tasks and compute the standard deviation of these three generation-index means. We also compute the standard deviation of the three generation scores for each canonical task and then average these values across the 15 tasks to obtain the mean within-task SD. For all 14 models, the mean within-task SD is higher than the corresponding generation-index mean SD, indicating that aggregate results across tasks can exhibit relatively small variation even when substantial variation can remain across generations of the same task.

The right panel further illustrates task-level variability using GPT-5.6 Sol as an example. For some canonical tasks, the three generation scores are close, whereas for others they span a much wider range. Thus, low aggregate variability across tasks does not imply that a model produces similar scores for every individual task; generation variability can differ across tasks.

\section{Ablation Studies: Configuration and Additional Results}

\subsection{Ablation Study I}
\label{app:ablation_study_i}

\subsubsection{Experimental Configuration}
\label{app:ablation_1_configuration}

\paragraph{Input Conditions.}

We compare three specification-input conditions on the 15 canonical
tasks in Core-15. M (Minimal) provides the game name or type, basic
actions, primary objectives, and termination/restart loop. D
(Detailed) provides the full specification content allocated to the
relevant generation stages. B augments D with public behavioral
requirements derived from functional checkpoints.

All three conditions share the same engineering constraints on
Pygame usage, file responsibilities, module dependencies, interfaces,
and output format. D and B use the same detailed specifications.
The additional checklist in B may restate or clarify requirements
and expose them to a generation stage that did not receive them in
D. Accordingly, $B-D$ measures the incremental effect of this
behavioral guidance, including its stage-specific allocation.
Checkpoint identifiers, detection methods, scoring rubrics, scores,
and weights are withheld from the generation model.

All conditions are evaluated against the same full target
specification. Because M omits some target requirements, $D-M$
measures the benefit of providing additional target information for
specification realization. It does not measure a change in the
model's intrinsic programming ability. The detailed specifications
cover seven sections: game overview; scene and interface; objects
and states; player interaction; core rules; flow and progression;
and goals, feedback, and termination.
Table~\ref{tab:ablation_1_stage_inputs} summarizes the inputs for
D and B. M receives its minimal description and the shared
engineering constraints at each stage.

\begin{table}[t]
    \centering
    \caption{Stage-specific inputs for D and B. B adds the public
    requirements in the final column to the corresponding D inputs.}
    \label{tab:ablation_1_stage_inputs}
    \small
    \setlength{\tabcolsep}{4pt}
    \renewcommand{\arraystretch}{1.10}
    \begin{tabularx}{\linewidth}{@{}lXX@{}}
        \toprule
        File & Specification and prior context & Additional inputs in B \\
        \midrule
        \texttt{state.py}
        & Overview; objects and states; flow and progression.
        & D2.1 state-modeling requirements. \\
        \texttt{game.py}
        & Overview; objects; interaction; rules; progression; goals,
          feedback, and termination; locked \texttt{state.py}.
        & D2.2--D2.5 behavioral requirements. \\
        \texttt{rendering.py}
        & Overview; scene and interface; goals, feedback, and termination;
          locked \texttt{state.py}.
        & None. \\
        \texttt{main.py}
        & Player interaction; public API summaries of preceding modules.
        & None. \\
        \bottomrule
    \end{tabularx}
\end{table}

Each candidate is generated in the fixed order
\texttt{state.py}, \texttt{game.py}, \texttt{rendering.py}, and
\texttt{main.py}. The framework supplies
\texttt{\_\_init\_\_.py}. Completed modules remain locked and
are provided to subsequent stages according to the same context
allocation as in the main benchmark. Candidate source code is not
reused across input conditions or independent repetitions.

\paragraph{Models and Generation Settings.}

We evaluate GPT-5.6 Terra, Qwen3.7 Plus, and Grok 4.3 using the
same model configurations and access routes as in the main
benchmark: Terra and Grok through OpenRouter, and Qwen through
the official Alibaba DashScope API. Generation temperature is 0.2,
and reasoning is disabled for all three models.

The per-attempt maximum output-token budgets for
\texttt{state.py}, \texttt{game.py}, \texttt{rendering.py}, and
\texttt{main.py} are 4,000, 7,000, 5,000, and 2,000 for Easy
and Medium tasks, and 5,000, 14,000, 10,000, and 2,000 for
Hard tasks. For each game, these limits are identical across
models and input conditions. The same technical retry policy is
used throughout: each stage permits one initial request and at
most one technical retry for a request failure or an incomplete
output. Retries use the unchanged prompt without compiler-error,
runtime, or evaluator feedback. Candidate programs are not
executed during generation, and no semantic repair is performed.

\paragraph{Candidate Inclusion.}

The reported results cover three models, 15 canonical games, and
three input conditions, giving 135 model--game--condition
combinations. We report one final score for each combination.
Candidate-code failures are distinguished from evaluation
infrastructure failures, and evaluation outcomes are not used to
select or replace candidates.


\paragraph{Evaluation and Score Aggregation.}
\label{app:ablation_1_evaluation}

All candidate projects are reevaluated on D1--D4 using the same
frozen evaluator version and configuration as the main benchmark.
The evaluator implementation, evidence-construction procedures,
scoring rubrics, judge configurations, and aggregation rules are
held fixed across input conditions. The text judge is
Qwen3-235B-A22B-2507 and the vision judge is
Qwen3-VL-235B-A22B-Instruct, served through Amazon Bedrock
with temperature 0 and reasoning disabled. All results reported
below use the final rerun scores.

For each candidate, the Overall score uses the continuous D1
gate defined in Appendix~\ref{app:overall-score}:
\begin{equation}
    g(D1)=\operatorname{clip}\left(\frac{D1-40}{20},0,1\right).
    \label{eq:ablation_1_gate}
\end{equation}
On the 0--100 dimension scales,
\begin{equation}
    \mathrm{Overall}
    =0.20D1+g(D1)\bigl(0.50D2+0.15D3+0.15D4\bigr).
    \label{eq:ablation_1_total}
\end{equation}
The gate is zero for $D1\leq40$, increases linearly for
$40<D1<60$, and equals one for $D1\geq60$. Passing all six
executability checks is recorded separately from this gate. For
example, passing the first four checks gives $D1=100(4/6)$ and
$g(D1)=1$, so D2--D4 contribute at full weight even if a later
executability check fails. Dimension scores may also be reported
separately as diagnostic measures. Evaluation infrastructure
failures are distinguished from candidate-code failures.

The Overall formula is applied at the candidate level.
Model-level condition summaries weight the 15 games equally,
and the overall condition summaries weight all 45
model--game combinations equally.

\subsubsection{Additional Results and Analysis}
\label{app:ablation_1_details}

\paragraph{Complete Per-Game Results.}
\label{app:ablation_1_complete_results}

Table~\ref{tab:ablation_1_complete_scores} reports the 135
model--game--condition scores from the final rerun. Means
and paired differences are calculated from unrounded scores and
reported to two decimal places; recalculation from rounded table
entries may therefore differ in the last displayed digit. Across
the 45 model--game combinations, mean Overall scores are 50.59
for M, 68.62 for D, and 68.42 for B.

\begin{table*}[t]
    \centering
    \caption{Complete Core-15 ablation results (0--100) from the
    final evaluation rerun. The table contains 135 model--game--condition scores.
    Overall scores use the continuous D1 gate in
    Equation~\ref{eq:ablation_1_total}.}
    \label{tab:ablation_1_complete_scores}
    \footnotesize
    \setlength{\tabcolsep}{3pt}
    \renewcommand{\arraystretch}{1.05}
    \begin{tabular}{@{}lrrrrrrrrr@{}}
        \toprule
        & \multicolumn{3}{c}{GPT-5.6 Terra}
        & \multicolumn{3}{c}{Qwen3.7 Plus}
        & \multicolumn{3}{c}{Grok 4.3} \\
        \cmidrule(lr){2-4}\cmidrule(lr){5-7}\cmidrule(lr){8-10}
        Game & M & D & B & M & D & B & M & D & B \\
        \midrule
        Dodge Blocks & 44.12 & 82.30 & 79.24 & 53.57 & 78.77 & 76.49 & 59.21 & 75.00 & 74.60 \\
        Farming-lite & 42.78 & 72.17 & 71.16 & 22.46 & 59.80 & 49.53 & 40.78 & 66.51 & 55.43 \\
        Carrot Defense-like & 60.89 & 68.35 & 73.58 & 48.84 & 49.15 & 61.87 & 43.78 & 45.42 & 58.93 \\
        Snake & 59.77 & 71.20 & 66.67 & 59.28 & 67.86 & 73.55 & 60.47 & 70.81 & 73.77 \\
        Tetris & 67.07 & 79.15 & 74.51 & 37.51 & 72.27 & 75.11 & 72.93 & 75.31 & 72.97 \\
        Top-down Arena Shooter & 43.04 & 70.20 & 67.02 & 35.65 & 41.17 & 46.76 & 24.07 & 66.84 & 82.52 \\
        Pipe Connection Puzzle & 58.08 & 84.08 & 89.54 & 40.58 & 73.48 & 6.67 & 47.61 & 78.86 & 84.74 \\
        Match-3 & 66.89 & 73.71 & 73.42 & 63.20 & 61.48 & 73.89 & 57.20 & 70.89 & 74.57 \\
        Maze Runner & 74.41 & 75.96 & 87.87 & 46.08 & 79.52 & 84.70 & 55.67 & 69.35 & 88.41 \\
        Simple Shop Management & 45.18 & 83.83 & 83.33 & 40.23 & 65.37 & 84.15 & 38.95 & 47.78 & 43.01 \\
        Mini Zelda-like Room Adventure & 57.75 & 59.94 & 63.33 & 28.86 & 59.23 & 6.67 & 43.99 & 62.67 & 58.25 \\
        Whack-a-Mole & 54.76 & 80.92 & 77.93 & 42.90 & 71.60 & 44.56 & 56.76 & 75.06 & 76.20 \\
        Asteroids & 40.94 & 87.15 & 74.44 & 67.11 & 74.02 & 71.87 & 44.55 & 56.44 & 53.22 \\
        Space Invaders & 65.35 & 58.39 & 70.69 & 69.27 & 78.36 & 61.48 & 66.76 & 75.13 & 74.41 \\
        Simple Platformer & 46.68 & 66.89 & 78.19 & 31.28 & 31.25 & 80.34 & 49.15 & 74.25 & 79.23 \\
        \midrule
        Mean & 55.18 & 74.28 & 75.40 & 45.79 & 64.22 & 59.84 & 50.79 & 67.35 & 70.02 \\
        \bottomrule
    \end{tabular}
\end{table*}

\paragraph{Paired Score Changes.}

Table~\ref{tab:ablation_1_per_game} reports score differences between
input conditions for the same model and game.
Positive $D-M$ values indicate higher scores under detailed
specifications, whereas positive $B-D$ values indicate higher
scores after adding the behavioral checklist.

D scores higher than M in 42 of the 45 model--game pairs. The
mean $D-M$ gains are 19.10 points for GPT-5.6 Terra, 18.43
for Qwen3.7 Plus, and 16.56 for Grok 4.3, with an overall mean
gain of 18.03 points. Thus, providing detailed target information
improves mean Overall scores for all three
models in this experiment.

The additional checklist produces smaller and model-dependent
changes. Mean $B-D$ differences are $+1.11$, $-4.38$, and
$+2.66$ points for Terra, Qwen, and Grok, respectively. B scores
higher than D in 22 pairs and lower in 23, with no ties; the
overall mean difference is $-0.20$ points. The checklist therefore
does not provide a consistent additional gain across the reported
comparisons.

\begin{table*}[t]
    \centering
    \caption{Paired differences in Overall score for the same
    model and game. Differences are computed before rounding.}
    \label{tab:ablation_1_per_game}
    \small
    \setlength{\tabcolsep}{4pt}
    \renewcommand{\arraystretch}{1.05}
    \begin{tabular}{@{}lrrrrrr@{}}
        \toprule
        & \multicolumn{2}{c}{GPT-5.6 Terra}
        & \multicolumn{2}{c}{Qwen3.7 Plus}
        & \multicolumn{2}{c}{Grok 4.3} \\
        \cmidrule(lr){2-3}\cmidrule(lr){4-5}\cmidrule(lr){6-7}
        Game & $D-M$ & $B-D$ & $D-M$ & $B-D$ & $D-M$ & $B-D$ \\
        \midrule
        Dodge Blocks & $+38.18$ & $-3.06$ & $+25.20$ & $-2.28$ & $+15.78$ & $-0.39$ \\
        Farming-lite & $+29.39$ & $-1.00$ & $+37.34$ & $-10.26$ & $+25.72$ & $-11.07$ \\
        Carrot Defense-like & $+7.46$ & $+5.23$ & $+0.31$ & $+12.72$ & $+1.64$ & $+13.51$ \\
        Snake & $+11.43$ & $-4.52$ & $+8.58$ & $+5.69$ & $+10.34$ & $+2.96$ \\
        Tetris & $+12.08$ & $-4.63$ & $+34.76$ & $+2.85$ & $+2.38$ & $-2.34$ \\
        Top-down Arena Shooter & $+27.16$ & $-3.18$ & $+5.53$ & $+5.58$ & $+42.77$ & $+15.68$ \\
        Pipe Connection Puzzle & $+26.00$ & $+5.46$ & $+32.90$ & $-66.81$ & $+31.24$ & $+5.88$ \\
        Match-3 & $+6.82$ & $-0.29$ & $-1.72$ & $+12.41$ & $+13.69$ & $+3.68$ \\
        Maze Runner & $+1.54$ & $+11.92$ & $+33.44$ & $+5.18$ & $+13.69$ & $+19.05$ \\
        Simple Shop Management & $+38.64$ & $-0.49$ & $+25.14$ & $+18.78$ & $+8.84$ & $-4.78$ \\
        Mini Zelda-like Room Adventure & $+2.19$ & $+3.39$ & $+30.37$ & $-52.56$ & $+18.68$ & $-4.42$ \\
        Whack-a-Mole & $+26.16$ & $-3.00$ & $+28.70$ & $-27.04$ & $+18.30$ & $+1.14$ \\
        Asteroids & $+46.21$ & $-12.72$ & $+6.91$ & $-2.15$ & $+11.89$ & $-3.22$ \\
        Space Invaders & $-6.96$ & $+12.31$ & $+9.10$ & $-16.89$ & $+8.37$ & $-0.72$ \\
        Simple Platformer & $+20.21$ & $+11.30$ & $-0.03$ & $+49.09$ & $+25.10$ & $+4.98$ \\
        \midrule
        Mean & $+19.10$ & $+1.11$ & $+18.43$ & $-4.38$ & $+16.56$ & $+2.66$ \\
        Improved / tied / declined & 14/0/1 & 6/0/9 & 13/0/2 & 8/0/7 & 15/0/0 & 8/0/7 \\
        \bottomrule
    \end{tabular}
\end{table*}

\paragraph{Sensitivity to Task Composition and Individual Comparisons.}
\label{app:ablation_1_sensitivity}

We retain the descriptive common-subset comparison reported for
11 games, excluding Carrot Defense-like, Pipe Connection Puzzle, Match-3,
and Mini Zelda-like Room Adventure consistently across all models and conditions.
The subset uses the same final scores as the full Core-15
comparison. Table~\ref{tab:ablation_1_common_subset} reports
its condition means.

\begin{table}[t]
    \centering
    \caption{Mean Overall scores on the common 11-game subset,
    excluding Carrot Defense-like, Pipe Connection Puzzle, Match-3, and Mini Zelda-like Room Adventure
    consistently across models and conditions. Scores use the same
    continuous D1 gate as the full Core-15 results.}
    \label{tab:ablation_1_common_subset}
    \small
    \setlength{\tabcolsep}{3pt}
    \renewcommand{\arraystretch}{1.05}
    \begin{tabular}{@{}lrrrrr@{}}
        \toprule
        Model & M & D & B & $D-M$ & $B-D$ \\
        \midrule
        GPT-5.6 Terra & 53.10 & 75.29 & 75.55 & $+22.19$ & $+0.27$ \\
        Qwen3.7 Plus & 45.94 & 65.45 & 68.05 & $+19.51$ & $+2.60$ \\
        Grok 4.3 & 51.75 & 68.41 & 70.34 & $+16.65$ & $+1.93$ \\
        \bottomrule
    \end{tabular}
\end{table}

The mean $D-M$ difference remains positive for all three models.
The mean $B-D$ differences on this subset are $+0.27$ for Terra,
$+2.60$ for Qwen, and $+1.93$ for Grok. Qwen's mean $B-D$
difference changes from $-4.38$ on the full task set to $+2.60$
on the common subset. This comparison shows sensitivity to task
composition; it does not establish a causal effect of generation
budget or replace the full-set results.

We also retain the analysis of Terra's Space Invaders pair. Its
$B-D$ difference is $+12.31$ points, compared with a mean of
$+1.11$ across all 15 games. Excluding this pair for descriptive
analysis gives $+0.31$ across the other 14 games. This comparison
quantifies the influence of one pair on the aggregate difference
without removing it from the main results.

These analyses concern task composition and individual comparisons.
They do not test sensitivity to dimension weights or the
specification-deviation penalty, and they do not establish
statistical significance for the M/D/B comparisons.

\paragraph{Interpretation.}

The main average improvement comes from supplying detailed
target requirements. Once those specifications are available,
additional checkpoint-derived guidance has a smaller effect whose
direction varies across models and tasks. The $D-M$ comparison
concerns the availability of target information, while $B-D$
concerns the incremental effect of behavioral guidance, including
clarification and stage-specific exposure of requirements. The
reported differences are descriptive; no claim of statistical
significance or equivalence between D and B is made.

\subsection{Ablation Study II}
\label{app:ablation_study_ii}







\subsubsection{Experimental Setup and Pairing}

This experiment examines whether the Qwen text judge tends to assign higher scores to candidates generated by Qwen-family models, using paired rankings to assess this potential preference. To control for task and replica differences, we pair candidates generated by Qwen3.8 Max-0902 and GPT-5.6 Terra for the same game and replica index, and introduce an independent GPT-5.6 Sol text judge as a reference. The two judge conditions use the same frozen candidate projects, game specifications, behavioral contracts, and visual evidence, so only the text judge route is changed. No candidate projects are regenerated for this experiment.

The experiment includes three replicas across nine games. Each \texttt{replica} $\times$ \texttt{game} combination forms one pair, yielding 27 paired comparisons and 54 candidate projects. Each candidate is evaluated under both text judge conditions, producing 108 judge results in total. Pairing is always performed using the same replica index and game.

\begin{table}[htbp]
    \centering
    \caption{Nine games included in Ablation Study II.}
    \label{tab:judge_bias_games}
    \small
    \setlength{\tabcolsep}{5pt}
    \renewcommand{\arraystretch}{1.05}
    \begin{tabular}{@{}rll@{}}
        \toprule
        \# & Game & Experiment ID \\
        \midrule
        1 & Snake & \texttt{snake} \\
        2 & Dodge Blocks & \texttt{dodge\_blocks} \\
        3 & Asteroids & \texttt{asteroids} \\
        4 & Tetris & \texttt{tetris} \\
        5 & Simple Platformer & \texttt{simple\_platformer} \\
        6 & Match-3 & \texttt{match3} \\
        7 & Farming-lite & \texttt{farming\_lite} \\
        8 & Top-down Arena Shooter & \texttt{arena\_shooter} \\
        9 & Mini Zelda-like Room Adventure & \texttt{mini\_zelda} \\
        \bottomrule
    \end{tabular}
\end{table}

The nine games and their internal experiment identifiers match the original experiment records.

\subsubsection{Judge Configuration and Scoring Protocol}

The main configurations of the two text judge conditions are summarized in Table~\ref{tab:judge_bias_configuration}. The Qwen condition uses Qwen3-235B-A22B-2507, while the GPT-5.6 Sol condition replaces only the text-scoring model. Both judges score D2 functionality, D2 specification deviation, and D3 code quality.

\begin{table*}[t]
    \centering
    \caption{Configurations of the two text judge conditions in Ablation Study II.}
    \label{tab:judge_bias_configuration}
    \small
    \setlength{\tabcolsep}{4pt}
    \renewcommand{\arraystretch}{1.05}
    \begin{tabularx}{\linewidth}{@{}lXX@{}}
        \toprule
        Configuration & Qwen text judge condition & GPT-5.6 Sol text judge condition \\
        \midrule
        Text judge
        & Qwen3-235B-A22B-2507
        & GPT-5.6 Sol \\
        Model ID
        & \texttt{qwen.qwen3-235b-a22b-2507}
        & \texttt{openai/gpt-5.6-sol} \\
        Access route
        & AWS Bedrock Mantle
        & OpenRouter \\
        Scored components
        & D2 functionality, D2 specification deviation, D3 code quality
        & D2 functionality, D2 specification deviation, D3 code quality \\
        Temperature
        & 0.0
        & 0.0 \\
        Reasoning
        & disabled (\texttt{enable\_thinking=false})
        & disabled (\texttt{reasoning.effort=none}) \\
        \bottomrule
    \end{tabularx}
\end{table*}

Apart from the text judge, the two conditions share the remaining evaluation configuration. Both use Qwen3-VL-235B-A22B-Instruct (\texttt{qwen.qwen3-vl-235b-a22b-instruct}) as the vision judge through AWS Bedrock Mantle, while action compilation and contract generation use Qwen3-235B-A22B-2507. Calls follow the Chat Completions format and return JSON objects. Each call allows at most two technical attempts with a two-second retry interval. Text checkpoints are processed in batches of at most eight items, while visual evidence is limited to at most 12 images and four transitions per batch. The scoring repetition is 1, and calls are stateless.

Thus, the GPT-5.6 Sol condition replaces only the text-scoring roles for functionality, specification deviation, and code quality; visual scoring, action compilation, contract generation, sampling, and the candidate projects remain unchanged. The experiment therefore compares the effect of text judge choice on paired candidate rankings rather than assessing potential preference in the vision judge.

\subsubsection{Official Scoring and Statistical Tests}

All candidate projects in this experiment achieve $D1=100$ in the official evaluation. This $D1=100$ is the executability score actually obtained by the candidates, rather than a post-hoc value introduced for analysis.

Spec2Game defines the Overall score as
\begin{equation}
    g(D1)=\operatorname{clip}\left(\frac{D1-40}{20},0,1\right),
\end{equation}
and
\begin{equation}
    \mathrm{Overall}
    =0.20D1+g(D1)\left(0.50D2+0.15D3+0.15D4\right).
\end{equation}

Because all candidates in this experiment satisfy $D1=100$, we have $g(D1)=1$, and the official Overall score reduces to
\begin{equation}
    \mathrm{Overall}
    =20+0.50D2+0.15D3+0.15D4.
\end{equation}

Thus, D1 provides the same executability baseline for all candidates, and paired rankings are determined by differences in their official D2--D4 scores. We use the official Overall score directly and do not construct an alternative score for this analysis.

For each \texttt{replica} $\times$ \texttt{game} pair, we compare the official Overall scores of the Qwen3.8 Max-0902 and GPT-5.6 Terra candidates. A pair is counted as a Qwen win if the Qwen candidate receives the higher score, as a Terra win if the Terra candidate receives the higher score, and as a tie if the two scores are equal.

We use a two-sided exact binomial test to determine whether the paired wins deviate significantly from a 1:1 split, and an exact McNemar test to compare the ranking directions produced by the two text Judges. Both tests use the 27 \texttt{replica} $\times$ \texttt{game} pairs as the units of analysis; different replicas or games are not treated as additional independent samples.








\subsubsection{Results and Scope}

The paired ranking results under the two text judge conditions are shown in Table~\ref{tab:judge_bias_appendix_results}.

\begin{table}[t]
    \centering
    \caption{Paired ranking results under the two text judge conditions.}
    \label{tab:judge_bias_appendix_results}

    \scriptsize
    \setlength{\tabcolsep}{2.2pt}
    \renewcommand{\arraystretch}{1.08}

    \begin{tabular}{
        @{}
        >{\raggedright\arraybackslash}p{0.30\columnwidth}
        >{\centering\arraybackslash}p{0.11\columnwidth}
        >{\centering\arraybackslash}p{0.11\columnwidth}
        >{\centering\arraybackslash}p{0.07\columnwidth}
        >{\centering\arraybackslash}p{0.12\columnwidth}
        >{\centering\arraybackslash}p{0.14\columnwidth}
        @{}
    }
        \toprule
        \textbf{Judge condition}
        & \shortstack{\textbf{Qwen}\\\textbf{wins}}
        & \shortstack{\textbf{Terra}\\\textbf{wins}}
        & \textbf{Ties}
        & \shortstack{\textbf{Binomial}\\$p$}
        & \shortstack{\textbf{Qwen}\\\textbf{win rate}} \\
        \midrule

        Qwen3-235B-A22B-2507
        & 13 & 14 & 0 & 1.000 & 48.1\% \\

        GPT-5.6 Sol
        & 15 & 12 & 0 & 0.701 & 55.6\% \\

        \bottomrule
    \end{tabular}
\end{table}

The Qwen text judge ranks the Qwen candidate higher in 13 of the 27 pairs, while the GPT-5.6 Sol judge does so in 15 pairs. Under neither condition do the paired outcomes deviate significantly from an even split, with two-sided exact binomial test results of $p=1.000$ and $p=0.701$, respectively.

The two Judges agree on the ranking direction in 21 of the 27 pairs. The exact McNemar test yields $p=0.688$, indicating no statistically significant difference in their ranking tendencies.

Across these paired comparisons, we do not observe the Qwen text judge systematically ranking Qwen3.8 Max-0902 candidates above GPT-5.6 Terra candidates; nor do the rankings produced by the Qwen Judge differ significantly from those produced by the independent GPT-5.6 Sol Judge. This provides some evidence that the current text judge setting does not exhibit a clear ranking preference toward Qwen-generated candidates in this experiment.

This conclusion applies to the fixed candidate set used in this experiment, covering nine games and three replicas, with the same Qwen-VL vision judge used in both conditions. Because the vision judge is held fixed across the two conditions, this ablation reflects the stability of rankings under the text judge comparison and does not directly assess potential preference in the vision judge.





\section{Human Assessment Protocol and Alignment Analysis}
\label{app:human_assessment}

\subsection{Coverage, Assignment, and Questionnaire}

\paragraph{Coverage and assignment.}
The questionnaire collection contains 585 generated projects from Core-15, covering all 15 canonical game families with 39 projects per family and 195 projects in each complexity tier. It does not include the additional controlled variants in Full-150. Each of the five annotation assignments contains 111 exclusive projects and two shared sets of 15 projects each, for 141 projects per assignment. CN-Q01 is designated for training, and INT-Q01 for response-agreement analysis. The primary assignments cover 555 distinct projects. Claude Sonnet 5 has automatic Core-15 scores but no candidates in the returned questionnaires and is therefore absent from the paired alignment analysis. Coverage counts establish balance across game families and complexity tiers, not a model-balanced or randomly sampled design.

\paragraph{Questionnaire and evidence.}
Each project has ten assessment items: five D2 functional checkpoints, one D4.1 overall visual-quality item, two D4.2 dynamic-feedback items, one D4.3 binary action-choice item, and one auxiliary question-validity item. The D2 anchors represent the five specification-realization components and retain the selected checkpoints' requirement-specific 0--2 rubrics and associated evidence. D4.1 and D4.2 use holistic 0--5 ratings of visual presentation and dynamic feedback. D4.3 records an A/B action choice for an image-grounded question. Question validity records whether the supplied visual evidence supports an unambiguous answer and is analyzed separately from game quality. The questionnaire covers selected evidence and requirements rather than all benchmark checkpoints or all automatic D4 criteria.

\paragraph{Response availability.}
The five returned files contain 7,050 item-level records. Excluding training leaves 6,300 records for 570 projects: 555 projects with one returned assessment each and 15 shared projects with five assessments each. Of these records, 3,973 contain available responses and 2,327 are marked \texttt{cannot\_rate}, each with a recorded reason. Unrateable entries are treated as missing judgments, not zeros. Candidate-level alignment uses the primary assignments, which contain 5,550 records: 3,505 available responses and 2,045 unrateable entries. The shared training and agreement sets are not pooled into this comparison. Thus the sample counts in the alignment tables refer to paired candidates with the required reported aggregate, not to annotators, questionnaire rows, or the full questionnaire collection.

\subsection{Candidate-Level Aggregation and Comparison}

\paragraph{Unit of analysis.}
Questionnaire responses are aggregated within candidate and assessment component before matching to the automatic result for the same candidate, run, and game. Where repeated ratings are available, they are averaged within candidate rather than counted as separate paired observations. In the returned primary assignments, each candidate has one assessment, so aggregation primarily combines the relevant questionnaire items. The paired set is constructed separately for each comparison; the reported $N$ therefore varies with response availability and the availability of the corresponding automatic score.

\paragraph{D2 alignment.}
For each candidate, the selected D2 checkpoint ratings are grouped by component to form five candidate-level means on the native 0--2 scale. These are compared with the corresponding automatic D2.1--D2.5 scores. The human-side D2 overall aggregate averages the five component means and is compared with the official automatic D2 aggregate after mapping both to 0--2. The dimension-level comparison contains 243 paired candidates. Because the questionnaire samples checkpoints and does not separately score specification-deviation penalties, this comparison measures correspondence with the official D2 result rather than reconstruction of its full computation.

\paragraph{D4 alignment.}
The human D4.1 rating and the mean of the available D4.2 ratings are mapped from 0--5 to 0--100. For D4.3, the recorded action choice is evaluated against the deterministic answer key to obtain candidate-level accuracy on a 0--100 scale. The human-side overall D4 aggregate averages the three normalized component values and is compared with official automatic D4; the reported dimension-level comparison contains 295 paired candidates. Normalization makes the numeric ranges comparable while preserving the distinction between holistic questionnaire ratings and the automatic criterion-level rubric. These aggregates are supplementary analytical measures and do not replace official dimension or leaderboard scores.

\paragraph{Action choice and question coverage.}
Two D4.3 comparisons answer different questions. The accuracy comparison relates human correctness against the frozen answer key to official automatic D4.3, which aggregates two questions. The same-question A/B comparison measures whether the questionnaire response and the LLM select the same action, regardless of whether that action is correct. In the question-aligned D4 sensitivity analysis, the automatic D4.3 contribution is replaced by the LLM's accuracy on the questionnaire's corresponding question, while the D4.1 and D4.2 contributions remain unchanged. The official-score and sensitivity comparisons use the same 295 candidates.

\paragraph{Statistics.}
We report Spearman's $\rho$ and Kendall's $\tau$ for rank correspondence, Pearson's $r$ for linear association, and mean absolute error (MAE) for score differences. MAE is expressed on 0--2 for D2 and 0--100 for D4, so its magnitude is interpreted within each scale. Exact A/B agreement is reported separately for action choices. These are pooled candidate-level comparisons, not correlations between model-level leaderboard ranks. Table~\ref{tab:human_candidate_alignment_detail} reports all component and dimension comparisons on their respective paired sets.

\begin{table}[t]
\centering
\small
\setlength{\tabcolsep}{4pt}
\begin{tabular}{lrrrrr}
\toprule
Comparison & $N$ & $\rho$ & $\tau$ & $r$ & MAE \\
\midrule
D2.1 Element modeling & 438 & 0.049 & 0.048 & 0.266 & 0.12 \\
D2.2 Interaction modeling & 384 & 0.466 & 0.418 & 0.596 & 0.37 \\
D2.3 Rules and mechanisms & 314 & 0.551 & 0.457 & 0.539 & 0.50 \\
D2.4 Dynamic evolution & 391 & 0.521 & 0.481 & 0.697 & 0.25 \\
D2.5 Goal and Termination Modeling & 277 & 0.805 & 0.699 & 0.799 & 0.33 \\
D2 overall & 243 & 0.779 & 0.655 & 0.791 & 0.13 \\
\midrule
D4.1 Visual presentation & 430 & 0.617 & 0.569 & 0.612 & 15.40 \\
D4.2 Dynamic feedback & 369 & 0.482 & 0.364 & 0.501 & 17.76 \\
D4.3 Action correctness & 313 & 0.540 & 0.519 & 0.548 & 27.00 \\
D4 overall & 295 & 0.564 & 0.416 & 0.525 & 12.10 \\
D4 question-aligned sensitivity & 295 & 0.692 & 0.549 & 0.674 & 9.50 \\
\bottomrule
\end{tabular}
\caption{Candidate-level questionnaire--LLM alignment on primary assignments. $N$ denotes paired candidates for each comparison. D2 errors use a 0--2 scale, and D4 errors a 0--100 scale. D4.3 compares questionnaire accuracy on one question with the official two-question automatic score. Overall comparisons use the reported candidate-level aggregates and are not averages of the correlations above.}
\label{tab:human_candidate_alignment_detail}
\end{table}

\paragraph{Alignment results.}
D2 overall exhibits positive rank correspondence ($\rho=0.779$, $\tau=0.655$), with the largest component-level Spearman correlation for Goal and Termination Modeling ($\rho=0.805$). D2.1 combines a small MAE of 0.12 with a rank correlation of 0.049, illustrating that score proximity and ranking correspondence capture different properties. For D4, the official-score comparison yields $\rho=0.564$ and $\tau=0.416$. Aligning the visual question raises these values to 0.692 and 0.549 and reduces MAE from 12.10 to 9.50. On the same-question A/B task, exact agreement is 0.885 across 313 paired candidates. The alignment results concern the available paired subsets and the questionnaire's assessment coverage.

\subsection{Shared-Set Agreement and Interpretation}

\paragraph{Response agreement.}
We assess agreement on the 15 shared projects in INT-Q01, separately from the candidate-level alignment analysis. Responses are matched by pack, project, and assessment-item identifier. Krippendorff's $\alpha$ uses ordinal distances for D2, D4.1, and D4.2 ratings and nominal distances for action choices and question-validity categories. Available-case estimates require at least two usable responses per item; complete-case estimates require all five. Training responses are excluded from both analyses.

\begin{table}[t]
\centering
\small
\setlength{\tabcolsep}{4pt}
\begin{tabular}{lrrr}
\toprule
Assessment category & Pairable / total & Available $\alpha$ & Complete $\alpha$ \\
\midrule
D2 functional checkpoints & 60/75 & 0.687 & 0.352 \\
D4.1 overall visual quality & 10/15 & 0.049 & 0.049 \\
D4.2 dynamic feedback & 20/30 & 0.753 & 0.349 \\
D4.3 action choice & 9/15 & 0.544 & 0.544 \\
Question validity (auxiliary) & 9/15 & 0.049 & 0.049 \\
\bottomrule
\end{tabular}
\caption{Descriptive agreement among shared-set responses. Counts refer to assessment items rather than distinct projects. Complete-case counts are 37, 10, 7, 9, and 9, respectively. These estimates do not measure questionnaire--LLM alignment and do not establish annotation independence.}
\label{tab:human_audit_agreement}
\end{table}

\end{document}